\pdfoutput=1

\documentclass{article}

\usepackage[table,xcdraw]{xcolor}
\PassOptionsToPackage{numbers, compress}{natbib}
\usepackage[preprint]{neurips_2025}

\usepackage[utf8]{inputenc} 
\usepackage[T1]{fontenc}    
\usepackage{hyperref}       
\usepackage{url}            
\usepackage{booktabs}       
\usepackage{amsfonts}       
\usepackage{nicefrac}       
\usepackage{microtype}      
\usepackage{microtype}
\usepackage{graphicx}
\usepackage{subfigure}
\usepackage{booktabs} 
\usepackage{hyperref}
\usepackage{multirow}
\usepackage{multicol}
\usepackage{float}

\usepackage{algorithm}
\usepackage{algorithmic}
\usepackage[inkscapelatex=false]{svg}
\usepackage{adjustbox}
\usepackage{amsmath}
\usepackage{amssymb}
\usepackage{mathtools}
\usepackage{amsthm}
\usepackage{import}
\usepackage{pifont}
\theoremstyle{plain}

\theoremstyle{definition}

\theoremstyle{remark}

\newcommand{\cmark}{\ding{51}}%
\newcommand{\xmark}{\ding{55}}%
\usepackage{cleveref}
\usepackage{wrapfig}
\usepackage{bookmark}

\title{\Large How to Unlock Time Series Editing? \\ {Diffusion-Driven Approach with Multi-Grained Control}}

\author{%
  Hao Yu$^*$ \\
  McGill University\\
  \texttt{hao.yu2@mail.mcgill.ca} \\
  \And
  Chu Xin Cheng\thanks{Equal contribution} \\
  California Institute of Technology\\
  \texttt{ccheng2@caltech.edu} \\
  \And
  Runlong Yu \\
  University of Pittsburgh\\
  \texttt{ruy59@pitt.edu} \\
  \And
  Yuyang Ye \\
  Rutgers University\\
  \texttt{yuyang.ye@rutgers.edu} \\
  \And
  Shiwei Tong\thanks{Corresponding author: shiweitong@tencent.com} \\
  IEGG, Tencent\\
  \texttt{shiweitong@tencent.com} \\
  \And
  Zhaofeng Liu \\
  IEGG, Tencent\\
  \texttt{zhaofengliu@tencent.com} \\
  \And
  Defu Lian \\
  University of Science and Technology of China\\
  \texttt{liandefu@ustc.edu.cn} \\
}
\begin{document}

\newcommand{\method}{\textsc{CocktailEdit}}

\maketitle

\begin{abstract}

Recent advances in time series generation have shown promise, yet controlling properties in generated sequences remains challenging. Time Series Editing (TSE) --- making precise modifications while preserving temporal coherence --- current methods struggle to consider multi-grain controls, including both point-level constraints and segment-level controls. We introduce the \method{} framework to enable simultaneous, flexible control across different types of constraints. 
This framework combines two key mechanisms: a confidence-weighted control for point-wise constraints and a classifier-based control for managing statistical properties such as sums and averages over segments.
Our methods achieve precise local control during the denoising inference stage while maintaining temporal coherence and integrating seamlessly, with any conditionally trained diffusion-based time series models. Extensive experiments across diverse datasets and models demonstrate its effectiveness. Our work bridges the gap between pure generative modeling and real-world time series editing needs, offering a flexible solution for human-in-the-loop time series generation and editing. The demonstration \footnote{Time Series Editor is available here: \href{https://huggingface.co/spaces/TSAnonymousDemo/TSEditor}{Huggingface Space}} is provided for experiencing Time series editing.
\end{abstract}

\begin{figure*}[hb]
    \vspace{-0.5em}
    \centering
    \includegraphics[width=1\textwidth]{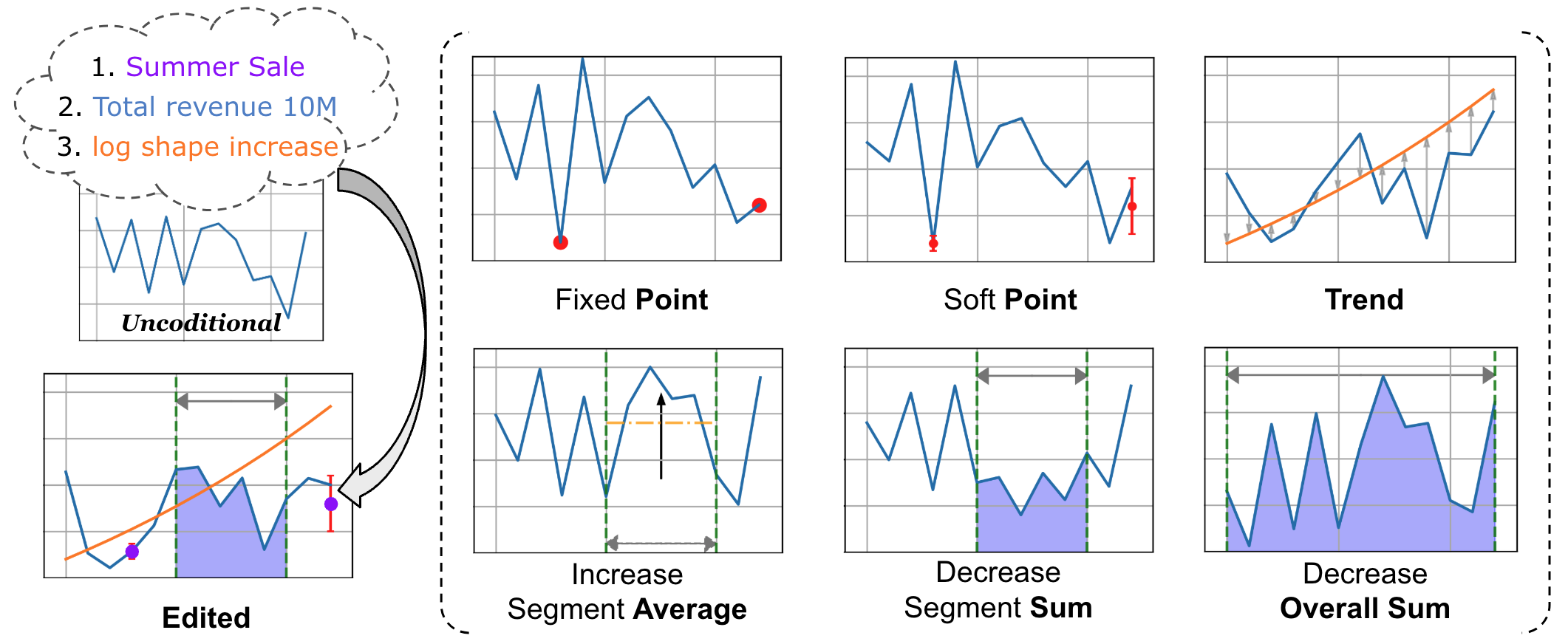}
     \caption{\method{} for Time Series Editing (TSE): From unconditional generation to fine-grained controlled time series generation, including (a) \textit{point-level control} using fixed points and soft points with uncertainty, (b) \textit{segment-level control} including trend and statistics metrics.} 
    \label{fig:example}
\end{figure*}

\vspace{-1.5em}

\section{Introduction}
Time series data is highly prevalent in our daily lives. From financial markets to healthcare systems, time series generation (TSG) is increasingly vital for analysis and prediction \cite{Wang2024}. Current approaches excel in unconditionally generating time series data, such as VAEs \cite{TimeVQVAE,CTSG,Sommers2024}, GANs \cite{Goodfellow2016,Wiese2020,GANTS}, and diffusion models \cite{Diffwave,Rasul2021,Li2022,Alcaraz2022}. However, this unconditionally generated time series may be impractical in many applications. For example, in the retail industry, companies hope to forecast the revenue of an unreleased product. Unconditionally generated data may fail to satisfy desired properties that seem obvious to producers, i.e. sales should peak during holiday seasons or after coupons are mailed. Therefore, we hope to guide the generation process with certain prior knowledge. A simple case of prior knowledge would be data points that the generated curve must pass through. An intuitive method would be to manually replace the point on the generated curve. However, this method does not consider the possibility that the point could be correlated with the rest of the time series curve. Therefore, this post-editing method risks generating out-of-distribution time series and does not leverage any information provided to us from partially observed data points or known trends. Thus, we hope to develop a principled way for generating time series while obeying certain user-specified rules, i.e., achieving controlled TSG.

We developed \method{}, a unifying framework named for its ability to mix and blend different types of constraints, to create precisely tailored time series outputs. Our framework embodies a wide range of various constraint types illustrated in Figure \ref{fig:example}.
We categorize these constraints by their range of focus. For example, we may have user-provided point-wise control, such as a single measurement on a fixed date with its confidence interval. Furthermore, we can also focus on global properties such as overall trend, periodicity, or specific data statistics (average, sum, optimizers, etc.). We regard these problems using an umbrella term called the Time Series Editing (TSE) challenge \cite{TETS}; tackling this challenge means producing time series data that can simultaneously satisfy user-defined constraints while maintaining temporal coherence and distributional fidelity.

Methods for tackling strict subsets of the TSE challenge exist. For instance, control using provided trends in data can be achieved using models GANs \cite{TimeGAN,CGAN}, VAEs \cite{TimeVQVAE,Sommers2024}, Diffusion Models \cite{Ang2023,CSDI,Microsoft,TimeDiT,Tian2024,Diff-MTS,ECG} and editing skills in Table~\ref{tab:relative-wrork}.  Diffusion models have also been applied to achieving point-wise control \cite{ConstrainedTSG}. 

\vspace{-0.5em}
\begin{table}[h]
\centering
\caption{Previous works for Time Series Editing. The RCGAN$^*$ and TimeGAN$^*$ adopt extra controls.}
\vspace{-0.75em}
\resizebox{0.85\textwidth}{!}{%
\begin{tabular}{l|ccccccc}
\toprule
\multirow{2}{*}{\textbf{Features}} & \multicolumn{3}{c}{\textbf{GAN Models}} & \multicolumn{4}{c}{\textbf{Diffusion Models}} \\ 
\cmidrule(lr){2-4} \cmidrule(lr){5-8}
 & RCGAN$^*$ & TimeGAN$^*$ & WGAN-COP & DiffTime & TimeWeaver & TEdit & \textbf{CocktailEdit} \\
 & \cite{RCGAN} & \cite{TimeGAN} & \cite{ConstrainedTSG} & \cite{ConstrainedTSG} & \cite{TimeWeaver} & \cite{TETS} & \textbf{(Our)} \\ 
\midrule
Fixed Point & \cmark & \cmark & \cmark & \cmark & \xmark & \xmark & \cmark \\
Soft Point & \xmark & \xmark & \xmark & \xmark & \xmark & \xmark & \textbf{\cmark} \\
Trend & \cmark & \cmark & \cmark & \cmark & \cmark & \cmark & \cmark \\
Statistics & \cmark & \cmark & \cmark & \cmark & \xmark & \xmark & \cmark \\
Train with Condition & \cmark & \cmark & \cmark & \cmark & \cmark & \cmark & \textbf{\xmark} \\
\bottomrule
\end{tabular}%
}
\label{tab:relative-wrork}
\end{table}
\vspace{-0.8em}

However, none of these methods demonstrates satisfactory performance while being sufficiently fast and easy to use. They often incorporate prior knowledge by means of altering the model or training data, limiting flexibility for real-time human-centered modifications.
Some methods choose to set a hard constraint on the learned distribution by embedding signals into the model architecture, e.g. time signals are injected into attention layers through additional embeddings, which limit the scope of this method and fail to consider possibilities of different kinds of human feedback/expected control signal \cite{Microsoft,TimeDiT,Tian2024,Diff-MTS,ScoreCDM,ECG}. In light of these considerations, we propose a novel method that overcomes these difficulties and can be used off-the-shelf with no additional training process. Our method utilizes diffusion models because they offer a mathematically rigorous framework that surpasses GANs in perceptual quality while avoiding adversarial training difficulties \cite{ScoreBased,DDIM}.

Through extensive experiments, we demonstrate that our method achieves precise control while maintaining temporal coherence and distributional fidelity. By 
bridging the gap between pure generative modelling and practical time series editing needs, our work offers a flexible solution for human-in-the-loop applications like revenue forecasting with expert knowledge integration and scenario analysis.

Our key contributions include:
\vspace{-0.5em}
\begin{itemize}\setlength{\itemsep}{0em}\setlength{\parskip}{0em}
    \item We propose the first unified framework for interactive TSE. Our algorithm combines multifaceted constraints across three axes: time (point-wise and segment-wise constraints), type (exact value, trend, or more complicated functional statistics), and confidence level.
    \item We conduct extensive experiments and demonstrate the effectiveness of our methods across four diverse datasets while incorporating standard and novel experimental metrics suited towards our task. 
    \item Our framework incorporates the constraints and operates only during the sampling process.
\end{itemize}

\section{Preliminaries}
\label{sec:preliminaries}

\subsection{Time Series Editing (TSE)}

TSE is an umbrella term embodying difficulties from traditional time series tasks: prediction, imputation, and generation.
However, while these classical tasks operate under specific, disjoint conditions -- prediction uses historical data up to time $t$, imputation relies on partially observed segments, and unconditional generation requires no prior data -- TSE must handle hybrid scenarios with multiple constraints.

To manage this complexity, TSE challenges are often decomposed along three dimensions: [1] \textit{temporal granularity} (point-wise, segment-wise, and whole-series operations), [2] \textit{conditioning type} (observed values, user-defined constraints, and learned patterns), and [3] \textit{transformation scope} ($\mathcal{X}$ for temporal range and $\mathcal{Y}$ for value constraints). This decomposition allows for systematic editing across scales while maintaining consistency and supporting extensions from local to global modifications.

\subsection{Problem Formulation}
\label{sec:problem_definition}
The objective of TSE is to generate time series data that replicates the statistical properties of real-world sequences while adhering to specific user-defined controls. Given a set of time series $S = \{\mathbf{x}_i\}_{i=1}^N$, where each $\mathbf{x}_i \in \mathbb{R}^{L \times D}$ ($L$ is the sequence length, $D$ is the number of features), we aim to develop a model $f_\theta$ that learned from $S$ and parameterized by $\theta$ such that:
$\mathbf{x} = f_\theta(\mathbf{C})$, $\mathbf{C}$ represents the prior conditions. The model should be able to generate time series $\mathbf{x}$ that satisfy both the learned distribution and the provided constraints.

\subsection{Diffusion Models: DDPM and Conditional}

Diffusion models, particularly Denoising Diffusion Probabilistic Models (DDPMs) \cite{DDPM,VAE,Goodfellow2016}, offer a robust backbone for addressing these TSE challenges. They tackle the controlled generation of time series, even with complex temporal dependencies and multi-scale patterns, by iteratively denoising data. 
This approach avoids common pitfalls such as the training instabilities of GANs or the reconstruction limitations of VAEs. The iterative nature of DDPMs, combined with guidance mechanisms like classifier-free and classifier-based methods \cite{ScoreBased}, facilitates precise control at each step. This ensures distributional consistency and the preservation of essential local and global time series properties, effectively supporting editing constraints across diverse temporal granularities.

\label{sec:ddpm-prelimiary}
DDPM approximate a data distribution $q(\mathbf{x})$ by gradually adding noise to data samples and then learning to reverse this noising process. The forward process is a Markov chain:
\vspace{-0.5em}
\small
\begin{equation}
q(\mathbf{x}_{1:K} \mid \mathbf{x}_0) 
= \prod_{k=1}^{K} q(\mathbf{x}_k \mid \mathbf{x}_{k-1})
\vspace{-0.5em}
\end{equation}
\small
\begin{equation}
q(\mathbf{x}_k \mid \mathbf{x}_{k-1}) 
= \mathcal{N}\!\Bigl(\mathbf{x}_k; \sqrt{1 - \beta_k} \,\mathbf{x}_{k-1}, \beta_k \mathbf{I}\Bigr),
\end{equation}
where $\beta_k \in [0,1]$ controls the noise variance at each step. The reverse process is modeled by a parameterized Gaussian \cite{DDPM,TSDiffusionSurvey}:
\begin{equation}
p_\theta(\mathbf{x}_{k-1} \mid \mathbf{x}_k) 
= \mathcal{N}\!\Bigl(\mathbf{x}_{k-1}; \boldsymbol{\mu}_\theta(\mathbf{x}_k, k), \sigma_k^2 \mathbf{I}\Bigr).
\end{equation}

To generate samples conditioned on extra inputs $\mathbf{C}$ (e.g., labels or prompts), Denoising Diffusion models augment the reverse process:
\vspace{-1em}
\small
\begin{equation}
p_\theta(\mathbf{x}\mid \mathbf{C}) = \prod_{t=1}^T p_\theta(\mathbf{x}_{t-1}\mid \mathbf{x}_t, \mathbf{C}), \text{where}
\vspace{-1em}
\end{equation}
\vspace{-1em}

\small
\begin{equation}
p_\theta(\mathbf{x}_{t-1}\mid\mathbf{x}_t,\mathbf{C}) 
= \mathcal{N}\!\bigl(\mathbf{x}_{t-1};\,\mu_\theta(\mathbf{x}_t,t,\mathbf{C}),\,\Sigma_\theta(\mathbf{x}_t,t)\bigr).
\end{equation}

\section{\method{}}

\label{sec:methodology}
To tackle the multi-grained TSE task, we propose \method{} that integrates point-wise and segment-wise controls, which includes two control mechanisms:
1) We view constraints as ``masks'' and use a teacher-forcing approach to guide the sampling process to enable confidence-weighted control.
2) Incorporate classifier-based control for managing segment-level aggregated statistical properties.
These mechanisms are implemented during the reverse diffusion process and can be used simultaneously, enforcing constraints across different granularities. A complete table of notations is included in Appendix~\ref{app:notation}.
\label{sec:point_wise_control}

\subsection{Problem Setup}
We begin by considering point-wise control as the fundamental problem. Given a time series $\mathbf{x}$ with both observed values $\mathbf{x}_{\text{ob}}$ at known indices $\Omega(\mathbf{x})$ and target values $\mathbf{x}_{\text{ta}}$ which is the final target output,
we aim to synthesize the target values using a diffusion model $p_\theta(\mathbf{x})$ trained on complete data while respecting the constraints. Formally, we define point-wise control constraints as:
$\mathbf{C}_{\text{point}} = \{(\mathbf{t}_i, \mathbf{v}_{i}, \mathbf{c}_{i}, \mathbf{w}_{i})\}_{i=1}^N$, where $\mathbf{t}_i \in \{1,\ldots,L\}$ specifies the time index,
$\mathbf{v}_{i} \in \mathbb{R}$ defines the expected value,
$\mathbf{c}_i \in \{1,\ldots,D\}$ means feature index, and
$\mathbf{w}_{i} \in \mathbb{R}$ represents the confidence level.

\subsection{Point-Wise Control}

\paragraph{Replace-Based Masking}
In Diffusion-TS and Diffwave \cite{Diffusion-TS, Diffwave}, conditional sampling is achieved by replace-based value-infilling during the sampling process of the diffusion model. That is, at the given time indices in the condition, the intermediate noisy samples are replaced with observed values. However, since the observed values belong to the data distribution rather than the intermediate noisy distributions, given the fixed point values $\mathbf{x}^{\text{ob}}_0$, we inject the same level of noise to it as in the forward process. 
\small
\begin{equation}
q\bigl(\mathbf{x}^{\text{ob}}_t \mid \mathbf{x}^{\text{ob}}_0\bigr) = \mathcal{N}\left(\mathbf{x}^{\text{ob}}_t; \sqrt{\overline{\alpha}_t} \, \mathbf{x}^{\text{ob}}_0, (1 - \overline{\alpha}_t)\mathbf{I}\right),
\label{eq:forward_process}
\end{equation}
\begin{equation}
\mathbf{x}^{\text{ta}}_t = \mathbf{x}^{\text{ta}}_{t+1} - \sqrt{\beta_t} \, \nabla_{\mathbf{x}^{\text{ta}}_{t+1}} \log p_{\theta}\bigl(\mathbf{x}^{\text{ta}}_{t+1} \mid \mathbf{x}^{\text{ta}}_{t+1}, \mathbf{x}^{\text{ob}}_t\bigr),
\end{equation}
where $\overline{\alpha}_t = \prod_{i=1}^{t}(1 - \beta_i)$. 

After obtaining $\mathbf{x}_{\text{ob}}^t$, we can then combine it with the intermediate noisy samples, similar to a teacher-forcing approach usin discrete masks $\mathbf{m} \in \{0, 1\}^{L}$ for the indices relative to the given conditions.
\begin{equation}
\mathbf{x}_t = \mathbf{m} \odot \mathbf{x}^{\text{ob}}_t + (1 - \mathbf{m}) \odot \mathbf{x}^{\text{ta}}_t.
\end{equation}

\textbf{Confidence-Based Masking}
A natural extension is to consider continuous masks $\mathbf{m} \in [0,1]^{L \times D}$ for enabling multivariate constraints and better generalization instead of $\mathbf{m} \in \{0,1\}^{L}$. 
The continuous mask allows for interpolating between observed data and generated samples, which enables viewing masks as confidence weighting. We can thus interpret $\mathbf{C}_{\text{point}}$ as the following: $\mathbf{x}_i, \mathbf{y}_{i}, \mathbf{c}_{i}$ are grouped as hard constraints $\mathbf{x}^{\text{ob}}_0$, and the combined $\mathbf{w}_{i} $ is the float mask $\mathbf{m}$, which is restricted to $\mathbf{w}_{i} \in [0,1]$ and $\mathbf{m} \in [0,1]^{L \times D}$. When $\mathbf{w}_{i} = 1$, we rely entirely on the observed data; when $\mathbf{w}_{i} = 0$, the model-generated sample is used. Appendix \ref{app:float-mask-proof} provides the formal proof.

\paragraph{Time-Dependent Weighted Guidance}
To emphasize stronger control in the later stages of the denoising process (i.e., as $t \rightarrow 0 $), we introduce the time-dependent weight:
\vspace{-0.5em}
\begin{equation}
\label{eq:time-dep-weight}
\omega_t = \exp(-\gamma \,\frac{t}{\mathrm{num\_timesteps}}),
\end{equation}
and rewrite the linear combination between constraints and target values as
\begin{equation}
\mathbf{x}_{t} \gets \omega_t \mathbf{m}\odot \mathbf{x}^{\text{ob}}_{t} + (1 - \omega_t\mathbf{m}) \odot \mathbf{x}^{\text{ta}}_t. 
\end{equation}

The visualized plot shows that the rate against the timestamp is in Appendix~\ref{app:time_dependent}.

\paragraph{Dynamic Error Adjustment on Mask}
Instead of a static mask, $\mathbf{m}$ can be dynamically adjusted during the reverse diffusion steps based on intermediate predictions or error metrics. This allows the model to adaptively allocate more attention to regions with higher uncertainty or discrepancy from desired constraints:
$\mathbf{m}_{t-1} = \mathbf{m}_t + \Delta \mathbf{m}_t$,
where $\Delta \mathbf{m}_t$ is a function of the current estimation error $\|\mathbf{x}^{\text{ta}}_{t} - \mathbf{x}^{\text{ta} \text{target}}\|$ and with less inference steps.

\subsection{Segment-wise Extension: Trend Control}

\label{subsubsec:trend_control}

Having a framework for confidence-weighted point-wise constraints allows natural extension to trend constraints, e.g. we may hope to generate a time-series that follows a seasonable periodic trend. The idea would be to compute $\mathbf{x}_{\text{ob}}^t$ using the given trend, which are represented as time-dependent functions. Let $\mathbf{L}$ encode the relationship between time and the expected function value over any segment. If $\mathbf{L}$ is continuous, we can interpolate it to a discrete reference trend $\mathbf{l}_t \in \mathbb{R}$ for each time and corresponding values:
\begin{equation}
\mathbf{l}_t = \mathbf{L}_{t_s} + \frac{t - t_s}{t_e - t_s} (\mathbf{L}_{t_e} - \mathbf{L}_{t_s}), \quad t \in \{t_s, \ldots, t_e\}.
\end{equation}
By having multiple groups of $(\mathbf{t}_{i}, \mathbf{v}_{i}, \mathbf{c}_\text{trend}, \mathbf{w}_{i})$ for $\mathbf{C}_\text{point}$ with $\mathbf{t}_{i} \in \{t_s, \ldots, t_e\}$, $\mathbf{v}_{i} = \mathbf{l}_t$, we thus achieve trend control. Similarly, $\mathbf{m}_{i}$ can be adjusted according to the user's confidence level.

\subsection{Combining Point-wise and Segment-wise Constraints}

To handle different temporal scale constraints on $\mathbf{C}_\text{point}$, we aim to work with multiple masks at varying granularities (e.g., local, segment, global). In order to control the diffusion process at different temporal resolutions, we obtain the final masks $\mathbf{m}$ using a reweighting scheme.
\begin{equation}
\mathbf{m} = \frac{\lambda_1 \mathbf{m}_{\text{local}} + \lambda_2 \mathbf{m}_{\text{segment}} + \lambda_3 \mathbf{m}_{\text{global}}}{\lambda_1 + \lambda_2 + \lambda_3},
\end{equation}
where $\lambda_i$ are weighting coefficients ensuring that the combined mask maintains values within $[0,1]$.

\subsection{Statistics Control}
\label{subsec:segment_wise_control}
\label{subsec:statistics_control}
We extend our notation set to formally define statistics control, which can be viewed as a variant of segment-wise control: $\mathbf{C}_{\text{segment}} = \{(\mathbf{s}_j, \mathbf{e}_j, \mathbf{c}_j, \alpha_{j}, \mathbf{w}_{j})\}_{j=1}^M$, where $\mathbf{s}_j, \mathbf{e}_j \in \{1, 2, \ldots, L\}$ are the start and end indices of the segment, $\alpha_{j} \in \mathbb{R}$ the parameters for aggregated functions, and $\mathbf{w}_{j}$ the confidence levels. 

\paragraph{Additional Loss Term}  
Following controllable diffusion models and \citep{ConstrainedTSG}, we inject a penalty term $\mathcal{L}_{\mathrm{pen}}(\mathbf{x}_t)$ inspired by classifier guidance to enforce alignment with the given condition $\mathbf{C}$:
\begin{equation}
\mathcal{L}_{\mathrm{pen}} \;=\; -\beta\,\log p_\phi(\mathbf{C} \mid \mathbf{x}_t),
\vspace{-0.2em}
\end{equation}
with a hyperparameter $\beta$.
Then, minimizing $\mathcal{L}_{\mathrm{pen}}$ effectively adds  
$-\beta \,\nabla_{\mathbf{x}_t} \log p_\phi(\mathbf{C} \mid \mathbf{x}_t)$
into the gradient flow \cite{TETS}.

\textit{\textbf{Example} of aggregated function -- Sum}
\vspace{-1em}
\vspace{-0.2em}
\small{
\begin{equation}
\mathcal{L}_{\mathrm{sum} [s_j:e_j]} = \left( \sum_{i=s_j}^{e_j} \mathbf{x}_{t,i} - S_{\text{target} [s_j:e_j]} \right)^2,
\vspace{-0.2em}
\end{equation}}
where $s_j$ and $e_j$ denote the start and end indices of the segment. The final loss term is:
\vspace{-0.2em}
\small
\begin{equation}
\mathcal{L}_{\mathrm{sum}} = \omega_t \sum \beta_{\mathrm{sum} [s_j:e_j]} \mathcal{L}_{\mathrm{sum} [s_j:e_j]},
\end{equation}
where $\beta_{\mathrm{sum} [s_j:e_j]}$ is the weight for the sum control term, and $\omega_t$ is the timestep-dependent weight that yields stronger guidance in the later stages of the denoising process defined in Eq.~\ref{eq:time-dep-weight}.

\subsection{Combined Multi-Grained Control}

Algorithm \ref{alg:sampling} in appendix presents our complete denoising control framework. Throughout the denoising process, we interleave point-wise floating mask control with segment-wise statistical control to gradually guide the denoising trajectory, requiring no model retraining or fine-tuning. 

\section{Experiments}

\subsection{Datasets}

Our evaluation employs four diverse datasets across real-world and simulated scenarios. Table \ref{tab:datasets} summarizes the dataset specifications.
The datasets include three real-world sources: ETTh \footnote{https://github.com/zhouhaoyi/ETDataset} for 
electricity transformer measurements, fMRI \footnote{https://www.fmrib.ox.ac.uk/datasets/netsim} for blood-oxygen-level-dependent time series and the Revenue dataset. The private Revenue dataset is sourced internally, containing revenue and two other features of hundreds of released video games.
The simulated Sines dataset \cite{TimeGAN} is generated with varying frequencies, amplitudes, and phases.

\begin{table}[H]
\centering
\caption{Dataset specifications summarizing the data sources, sequence lengths, and dimensionality of four datasets used.}
\resizebox{0.75\columnwidth}{!}{ 
\begin{tabular}{lcccc}
\toprule
\small
Dataset & Type & Source & Length & Features\\
\midrule
ETTh & Real-world & Electricity transformer & 24 & 28\\
fMRI & Real-world & BOLD time series & 24 & 50\\
Revenue & Real-world & Game sales & 365 & 3\\
Sines & Simulated & Synthetic waves & 24 & 5\\
\bottomrule
\end{tabular}
}
\label{tab:datasets}
\end{table} 

The only difference is the fMRI dataset, which has a sequence length of 48 compared to the original with 24 timesteps.

\subsection{Experimental Setup}
\label{sec:experimental_setup}
Following the infilling training setup of the Diffusion-TS framework \cite{Diffusion-TS}, we adopt the CSDI \cite{CSDI} and Diffusion-TS \cite{Diffusion-TS} as baseline fundamental models.
The training stage uses the entire dataset in a multivariate time series aligned with the community. For all experiments, control signals were applied exclusively to the first channel ($\textbf{c} = 0$) to observe inter-channel influences while enabling multi-channel extensibility. All figures and results are based on the first channel unless stated otherwise. 
Training parameters and inference settings are expressed in Appendix \ref{app:train}, such as choosing controlling points and confidence scores.

\subsection{Evaluation Metrics}
We employ multiple metrics across two key aspects: control accuracy and distribution quality. For \textit{control accuracy}, we measure: (i) \textbf{Mean Absolute Difference (MAD)} between generated and target values at controlling points, defined as $\text{MAD} = \frac{1}{N}\frac{1}{A} \sum_{i=1}^{N} \sum_{j=1}^{A} \left| x_{\text{cond gen} (i, j)} - x_{\text{target} (i, j)} \right|$; and (ii) \textbf{Statistic Control Result} to evaluate adherence to target statistic constraints. For example, the control target function is the sum of series, we directly observe the actual sum value change to validate the controllability.
For \textit{distribution alignment} check, followed by many works \cite{Diffwave,Diffusion-TS,DiffusionBridge,Diff-MTS}, we utilize: 
(i) \textbf{Discriminative \& Predictive scores} to assess how well-generated sequences replicate real data patterns, where discriminative score is $\mid$accuracy $-$ 0.5$\mid$ and predictive score uses mean absolute error (MAE) between predictions and ground truth; (ii) \textbf{Context-FID score} which leverages a pre-trained TS2Vec model to measure distribution-level alignment, with lower scores indicating better similarity; and (iii) \textbf{Correlational score} that quantifies feature-level covariance preservation. We also complement these quantitative metrics with \textbf{Kernel Density Function} visualizations 
to provide qualitative insights into distribution alignment.

\section{Results}

\subsection{Point-Wise Control}

We conducted experiments to examine the relationship between point confidence values and prediction accuracy across different datasets. These experiments aim to validate our hypothesis that higher confidence values lead to more precise point control and verify the behavior at extreme confidence values (0.01 and 1.0).

\begin{figure}[hb]
\centering
\begin{adjustbox}{trim={0 0 {0.59\textwidth} {0.01\textheight}},clip}
\includegraphics[width=1.02\textwidth]{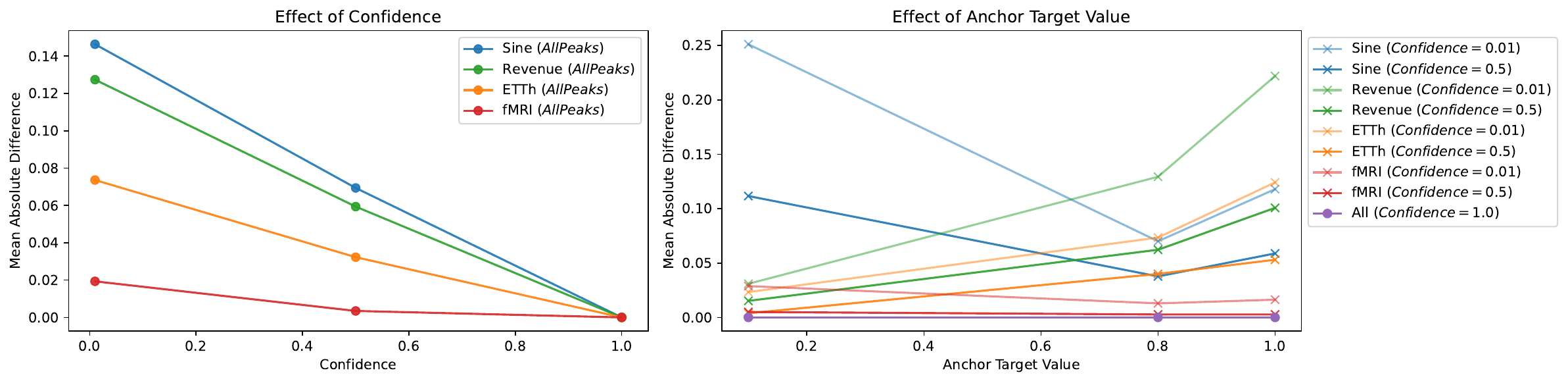}
\end{adjustbox}
\begin{adjustbox}{trim={0 0 {0.59\textwidth} {0.01\textheight}},clip}
\includegraphics[width=1.02\textwidth]{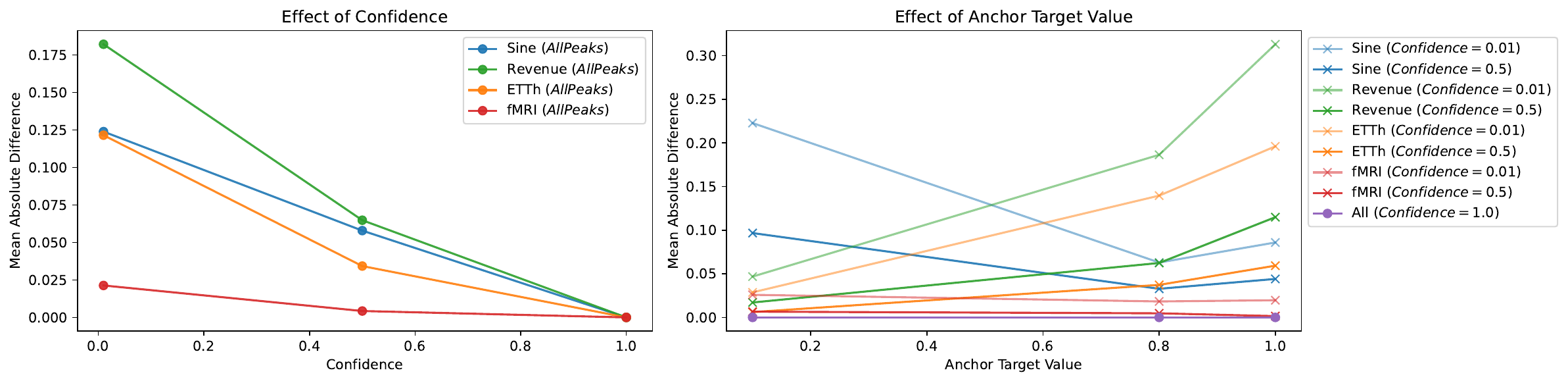}
\end{adjustbox}
\vspace{-1em}
\caption{\textbf{Point Control} The influence of confidence across datasets. The right figure is examined when enabling the dynamic error adjustment and time-dependent weight, while the left figure is not. ``AllPeaks'' means the average of all experiments of point-wise control. More details are in Appendix \ref{app:anchor_analysis_full}}.
\vspace{-2em}
\label{fig:anchor_trend_final}
\end{figure}

Table \ref{tab:anchor-summary} with the Diffusion-TS backbone demonstrates that increasing confidence values leads to a consistent MAD reduction across all datasets, ultimately converging to 0.0 at maximum confidence. This convergence at confidence = 1.0 validates our theoretical guarantee that maximum confidence forces the point to be a fixed point. The monotonic and linear decrease of MAD with increasing confidence is visualized in Figure~\ref{fig:anchor_trend_final}. 

The second finding reveals that extreme target values (0.1 or 1.0) produce larger MAD values than moderate confidence levels (e.g., 0.5) across different datasets. For instance, on the fMRI Dataset, the MAD at $a_{target} = 0.1$ is $0.113$, while at $a_{target} = 0.8$, it reduces to $0.052$. This pattern, visible in Table~\ref{tab:anchor-summary}, supports our hypothesis about the trade-off between very low and very high confidence values in controlling point selection. Comparing the MAD of the Original Dataset and Unconditional across the three target values, the point approach demonstrates a smaller MAD even with a confidence score of 0.01. This indicates that point-wise control helps generate time series data closer to the target value at a specified time. Furthermore, the pattern of largest and smallest MAD values aligns with those in the original and unconditional versions, suggesting that point-wise control preserves the original distribution and MAD for each dataset while reducing its MAD magnitude.

\begin{table*}[!tbp]
\centering
\begin{minipage}{0.49\textwidth}
\centering
\caption{\small \textbf{Point Control} MAD of the given point indices in different setups of different confidence levels and target values.}
\label{tab:anchor-summary}
\resizebox{\textwidth}{!}{%
\begin{tabular}{ccccccc}
\toprule
\multicolumn{7}{c}{\textbf{ETTh Dataset}} \\ \midrule
\multicolumn{1}{c|}{Target Value \textbackslash Confidence} & Original & 0.00 (Uncon) & 0.01 & 0.50 & 1.00 & Average \\ \midrule
\multicolumn{1}{c|}{0.1} & \cellcolor[HTML]{C4E7D7}.0661 & \cellcolor[HTML]{C8E9D9}.0614 & \cellcolor[HTML]{EBF7F1}.0233 & \cellcolor[HTML]{FCFEFD}.0038 & \cellcolor[HTML]{FFFFFF}.0000 & \cellcolor[HTML]{F7FCFA}.0090 \\
\multicolumn{1}{c|}{0.8} & \cellcolor[HTML]{6DC49A}.7201 & \cellcolor[HTML]{6BC498}.7363 & \cellcolor[HTML]{BEE5D2}.0734 & \cellcolor[HTML]{DCF1E7}.0400 & \cellcolor[HTML]{FFFFFF}.0000 & \cellcolor[HTML]{DEF2E8}.0378 \\
\multicolumn{1}{c|}{0.8} & \cellcolor[HTML]{7BCAA4}.5775 & \cellcolor[HTML]{79C9A2}.6021 & \cellcolor[HTML]{A8DCC3}.1293 & \cellcolor[HTML]{C8E9D9}.0622 & \cellcolor[HTML]{FFFFFF}.0000 & \cellcolor[HTML]{C6E8D8}.0638 \\
\multicolumn{1}{c|}{1.0} & \cellcolor[HTML]{67C296}.7775 & \cellcolor[HTML]{65C194}.8021 & \cellcolor[HTML]{9FD8BD}.2219 & \cellcolor[HTML]{ABDDC5}.1007 & \cellcolor[HTML]{FFFFFF}.0000 & \cellcolor[HTML]{AADDC5}.1075 \\
\multicolumn{1}{c|}{Average} & \cellcolor[HTML]{83CDA9}.4988 & \cellcolor[HTML]{82CDA8}.5102 & \cellcolor[HTML]{A8DCC3}.1274 & \cellcolor[HTML]{CAEADB}.0594 & \cellcolor[HTML]{FFFFFF}.0000 & \multicolumn{1}{l}{} \\ \midrule

\multicolumn{7}{c}{\textbf{fMRI Dataset}} \\ \midrule
\multicolumn{1}{c|}{0.1} & \cellcolor[HTML]{89CFAD}.4423 & \cellcolor[HTML]{89CFAD}.4409 & \cellcolor[HTML]{E6F5EE}.0288 & \cellcolor[HTML]{FBFEFC}.0050 & \cellcolor[HTML]{FFFFFF}.0000 & \cellcolor[HTML]{F5FBF8}.0113 \\
\multicolumn{1}{c|}{0.8} & \cellcolor[HTML]{9BD7BA}.2620 & \cellcolor[HTML]{9BD7BA}.2632 & \cellcolor[HTML]{F4FBF8}.0129 & \cellcolor[HTML]{FDFFFE}.0027 & \cellcolor[HTML]{FFFFFF}.0000 & \cellcolor[HTML]{FBFEFC}.0052 \\
\multicolumn{1}{c|}{1.0} & \cellcolor[HTML]{87CFAC}.4577 & \cellcolor[HTML]{87CFAC}.4591 & \cellcolor[HTML]{F1FAF5}.0163 & \cellcolor[HTML]{FDFFFE}.0027 & \cellcolor[HTML]{FFFFFF}.0000 & \cellcolor[HTML]{FAFDFC}.0063 \\
\multicolumn{1}{c|}{Average} & \cellcolor[HTML]{8ED2B1}.3873 & \cellcolor[HTML]{8ED2B1}.3877 & \cellcolor[HTML]{EEF8F4}.0193 & \cellcolor[HTML]{FCFEFD}.0035 & \cellcolor[HTML]{FFFFFF}.0000 & \multicolumn{1}{l}{} \\
\midrule

\multicolumn{7}{c}{\textbf{Sine Dataset}} \\ \midrule
\multicolumn{1}{c|}{0.1} & \cellcolor[HTML]{74C79E}.6530 & \cellcolor[HTML]{73C79E}.6617 & \cellcolor[HTML]{9CD7BA}.2511 & \cellcolor[HTML]{AADDC4}.1116 & \cellcolor[HTML]{FFFFFF}.0000 & \cellcolor[HTML]{A9DCC4}.1209 \\
\multicolumn{1}{c|}{0.8} & \cellcolor[HTML]{A6DBC2}.1466 & \cellcolor[HTML]{A6DBC2}.1455 & \cellcolor[HTML]{C1E6D4}.0700 & \cellcolor[HTML]{DEF2E8}.0377 & \cellcolor[HTML]{FFFFFF}.0000 & \cellcolor[HTML]{DFF2E9}.0359 \\
\multicolumn{1}{c|}{1.0} & \cellcolor[HTML]{9CD7BB}.2470 & \cellcolor[HTML]{9DD8BB}.2383 & \cellcolor[HTML]{A9DDC4}.1179 & \cellcolor[HTML]{CBEADB}.0588 & \cellcolor[HTML]{FFFFFF}.0000 & \cellcolor[HTML]{CBEADB}.0589 \\
\multicolumn{1}{c|}{Average} & \cellcolor[HTML]{92D3B4}.3489 & \cellcolor[HTML]{92D3B4}.3485 & \cellcolor[HTML]{A6DBC2}.1463 & \cellcolor[HTML]{C1E6D4}.0694 & \cellcolor[HTML]{FFFFFF}.0000 & \multicolumn{1}{l}{} \\ 
\bottomrule
\end{tabular}%
}
\end{minipage}\hfill
\begin{minipage}{0.49\textwidth}
\centering
\caption{\small \textbf{Sum Control} Average summation value for various weights of sum control. ``Original'': the original training set; ``Uncon'': Unconditional generated samples. The \textbf{biggest} and \underline{smallest} MAD are labelled in each dataset. (Diffusion-TS)}
\label{tab:sum-change-weight}
\resizebox{\textwidth}{!}{%
\begin{tabular}{c|cc|cccc}
\toprule
\multirow{2}{*}{\textbf{Dataset}} & \multirow{2}{*}{Original} & \multirow{2}{*}{Uncon} & \multicolumn{4}{c}{Target Value} \\
    &  &  & -100.0 & 20.0 & 50.0 & 150.0 \\ \hline
Sine & 17.881 & 18.086 & \underline{ 7.114} & 20.146 & 20.991 & \textbf{21.031} \\
Revenue & 80.194 & 76.619 & \underline{ 52.675} & 55.380 & 58.585 & \textbf{117.868} \\
ETTh & 1.924 & 1.535 & \underline{ 0.802} & 8.502 & 10.323 & \textbf{11.102} \\
fMRI & 12.990 & 12.980 & \underline{ 4.675} & 17.192 & 19.811 & \textbf{20.508} \\ 
\toprule
\multirow{2}{*}{\textbf{Dataset}} & \multirow{2}{*}{Original} & \multirow{2}{*}{Uncon} & \multicolumn{4}{c}{Weight Value} \\
    &  &  & 1 & 10 & 50 & 100 \\ \toprule
Sine & 17.881 & 18.086 & \underline{ 7.110} & 7.114 & \textbf{7.117} & 7.115 \\
Revenue & 80.194 & 76.619 & \textbf{52.767} & \underline{ 52.675} & 52.741 & 52.725 \\
ETTh & 1.924 & 1.535 & 0.798 & \textbf{0.802} & \underline{ 0.796} & 0.800 \\
fMRI & 12.990 & 12.980 & 4.678 & \underline{ 4.675} & 4.677 & \textbf{4.682} \\ \bottomrule
\end{tabular}%
}
\end{minipage}
\vspace{-1em}
\end{table*}

\subsection{Segment-Wise Statistical Control}

Our experiments investigate two key aspects of the aggregated statistic adjustment mechanism: (1) the effectiveness of different target values in steering the \textbf{segment sum} and \textbf{sequence sums}, and (2) the impact of \textbf{weight parameters} on control strength. These experiments aim to validate whether our method can reliably guide sequences toward desired sum targets while maintaining data fidelity.

\paragraph{Whole Sequence Sum}
As shown in Table~\ref{tab:sum-change-weight}, the controlled sequences consistently respond to different target values across all datasets. For the Revenue dataset, when targeting 150.0, the sequence sum increases significantly from 76.6 to 117.9, while targeting -100.0 reduces it to 52.7. Similar patterns are observed in other datasets - ETTh shows controlled variation from 1.5 to 11.1 (100) and 0.8 (-150), etc.
Moreover, Revenue shows increasing steps: +2.705 from -100.0 to 20.0, +3.206 from 20.0 to 50.0, and +59.283 from 50.0 to 150.0. 
Both patterns demonstrate fine-grained control for smaller adjustments and the ability to make substantial changes when needed. Figures in Appendix \ref{app:sum_demo} visualize more controlled time series with different targets.

\begin{figure*}[!ht]
\label{fig:sum_trends_normalized}
\centering
\includegraphics[width=\textwidth]{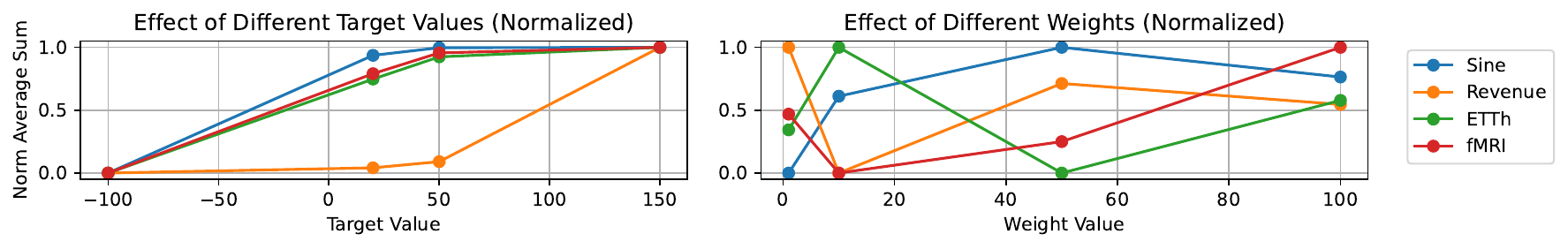}
\vspace{-2em}
\caption{Normalized comparison of sum control effectiveness across datasets. Values are scaled relative to dataset-specific ranges to enable direct comparison between different domains. Lines show progressive convergence toward target sums. (Diffusion-TS)}
\vspace{-0.5em}
\end{figure*}

The weight parameter experiments reveal that varying weights from 1 to 100 produces only minor changes in the resulting sums. Target value rather than the control weight is the primary driver of loss-based control performance.
Then, the weight parameters may be simplified in practical applications since they do not contribute substantially to control performance. The normalized trends in Figure~\ref{fig:sum_trends_normalized} further support this observation.

\vspace{-1em}

\begin{figure}[h]

    \includegraphics[width=.24\textwidth]{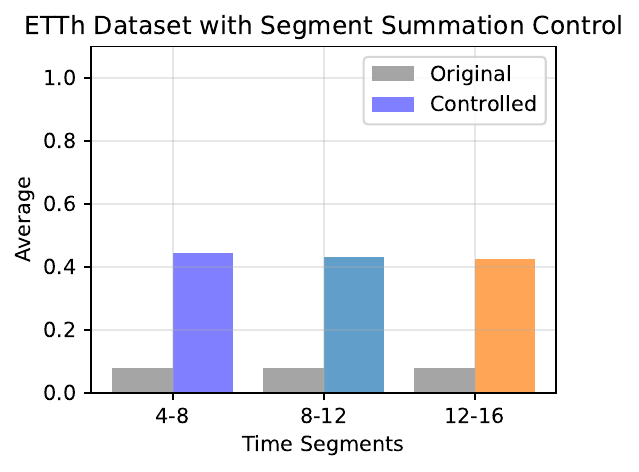}\hfill
    \includegraphics[width=.24\textwidth]{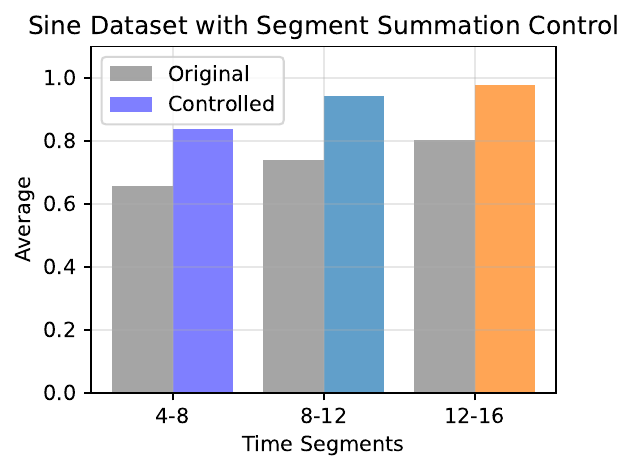}\hfill
    \includegraphics[width=.24\textwidth]{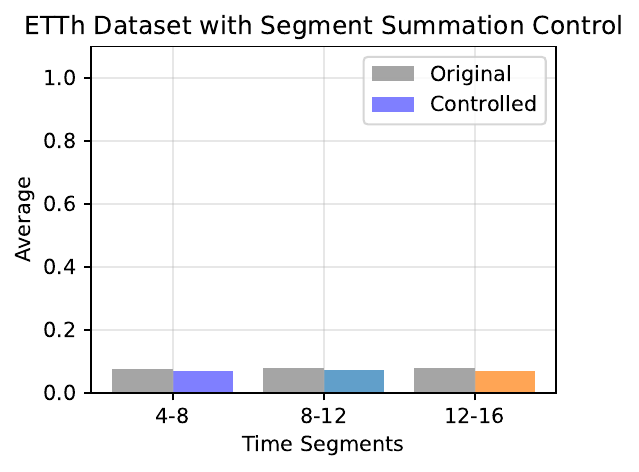}\hfill
    \includegraphics[width=.255\textwidth]{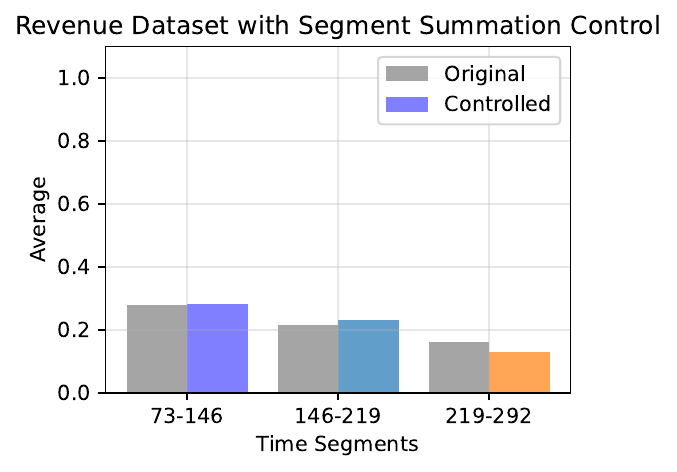}

    \caption{Segment-wise sum control results for different datasets. The shaded area represents the controlled segment, with the corresponding average sum value across different time segments. The datasets from left to right are: ETTh, Revenue, fMRI and Sine.}
    \vspace{-1em}
    \label{fig:bins}
\end{figure}

\paragraph{Segment Sum}
To evaluate segment-wise control, we tested 3 segments: $(0.2L, 0.4L)$, $(0.4L, 0.6L)$, $(0.6L, 0.8L)$ with target value 150. 
Figure~\ref{fig:bins} shows that each controlled segment demonstrates increased sum adjustments. 
The effectiveness is particularly visible in the ETTh and Revenue datasets. The controlled segments exhibit clear increases in area under the curve when targeting higher sums with Diffusion-TS, while the effect of CSDI controlling is not obvious.
\vspace{-1em}

\subsection{Distribution Alignment}

Table \ref{tab:complete-distribution} demonstrates that Diffusion-TS achieves better distribution matching than CSDI in unconditional generation across all datasets, with scores approaching 0, particularly for ETTh (0.034) and Sines (0.019). However, both Point-wise and Statistics controls lead to increased divergence, suggesting a trade-off between control and distribution preservation. The comprehensive table can be found in Appendix~\ref{app:metrics}.

\begin{table}[h]
\centering
\caption{Discriminative score for our method on CSDI and Diffusion-TS (DTS). The lower the score, the more similar the distribution of generated time series with original datasets.}
\label{tab:complete-distribution}
\resizebox{0.7\columnwidth}{!}{
\begin{tabular}{l|l|cccc}
\toprule
Model & Control & ETTh & Revenue & fMRI & Sines \\ \midrule
\multirow{3}{*}{Our - CSDI} & \textbf{Unconditional} & 0.361$\pm$0.007 & 0.245$\pm$0.164& 0.306$\pm$0.021 & 0.017$\pm$0.007 \\
    & Point-wise & 0.470$\pm$0.003 & 0.313$\pm$0.046& 0.482$\pm$0.004 & 0.430$\pm$0.038 \\
    & Statistics & 0.373$\pm$0.007 & 0.272$\pm$0.055& 0.377$\pm$0.019 & 0.034$\pm$0.007 \\ \midrule
\multirow{3}{*}{Our - DTS} & \textbf{Unconditional} & 0.034$\pm$0.026 & 0.209$\pm$0.185& 0.089$\pm$0.033 & 0.019$\pm$0.008 \\
    & Point-wise & 0.437$\pm$0.004 & 0.393$\pm$0.030& 0.495$\pm$0.001 & 0.460$\pm$0.011 \\
    & Statistics  & 0.477$\pm$0.003 & 0.426$\pm$0.032& 0.498$\pm$0.001 & 0.451$\pm$0.029 \\ \bottomrule
\end{tabular}%
}
\end{table}

Our time efficiency analysis from Table \ref{tab:time} reveals CSDI's computational advantage, processing samples 3-9x faster than Diffusion-TS across all datasets. The runtime scales primarily with sequence length, as evidenced by the Revenue dataset's higher processing times. Notably, both point-wise and statistics-based controls add minimal computational overhead, maintaining consistent performance across control configurations. Feature dimensionality has a secondary but observable impact on processing time.
\begin{table}[h]
\centering
\caption{Time efficiency analysis of our method compared on CSDI and Diffusion-TS (DTS). The results show the average time per sample for each dataset and control configuration. 500 examples per batch with an average of 5 batches.}
\resizebox{0.7\columnwidth}{!}{%
\begin{tabular}{l|cccccc}
\toprule
\multicolumn{1}{c|}{\multirow{2}{*}{Method}} & \multicolumn{3}{c|}{Dataset} & \multicolumn{3}{c}{Time Per Sample (ms)} \\
\multicolumn{1}{c|}{} & Name & Seq & \multicolumn{1}{c|}{Features} & Uncon & Point-Wise & Statistics \\ \midrule
\multirow{4}{*}{Our - CSDI} & Energy & 24 & \multicolumn{1}{c|}{28} & \textbf{0.064$\pm$0.000} & 0.064$\pm$0.004 & 0.064$\pm$0.001 \\
 & fMRI & 24 & \multicolumn{1}{c|}{50} & 0.116$\pm$0.000 & \textbf{0.115$\pm$0.001} & 0.117$\pm$0.001 \\
 & Revenue & 365 & \multicolumn{1}{c|}{3} & \textbf{1.000$\pm$0.001} & 1.016$\pm$0.022 & 1.001$\pm$0.013 \\
 & Sines & 24 & \multicolumn{1}{c|}{5} & 0.018$\pm$0.000 & 0.017$\pm$0.001 & 0.018$\pm$0.001 \\ \midrule
\multirow{4}{*}{Our - DTS} & Energy & 24 & \multicolumn{1}{c|}{28} & \textbf{0.278$\pm$0.017} & 0.304$\pm$0.002 & 0.292$\pm$0.001 \\
 & fMRI & 24 & \multicolumn{1}{c|}{50} & 0.376$\pm$0.154 & \textbf{0.232$\pm$0.883} & 0.549$\pm$0.274 \\
 & Revenue & 365 & \multicolumn{1}{c|}{3} & \textbf{9.163$\pm$0.411} & 9.580$\pm$5.126 & 9.223$\pm$0.044 \\
 & Sines & 24 & \multicolumn{1}{c|}{5} & \textbf{0.046$\pm$0.015} & 0.059$\pm$0.020 & 0.048$\pm$0.001 \\ \bottomrule
\end{tabular}%
\label{tab:time}
}
\end{table}

\section{Related Work}
Traditional generative tasks employed GANs \cite{Wiese2020,Nguyen2021,TimeGAN,CGAN}, VAEs \cite{TimeVAE,TimeVQVAE}, Diffusion Models \cite{Ang2023,CSDI,Microsoft,TimeDiT,Tian2024,Diff-MTS,ECG,TETS,DDBM}, and Flow Matching \cite{FM-TS}, establishing foundational techniques for synthetic time series generation. 
\subsection{Time Series Generation Models}

TimeGAN \cite{TimeGAN} introduced temporal-aware adversarial training, while subsequent works like C-TimeGAN \cite{CGAN} and CGAN-TS \cite{GANTS} incorporated conditional generation capabilities. Recent Transformer-based architectures \cite{Sommers2024,Wen2023} have further enhanced representation capabilities for temporal patterns.
Early control mechanisms focused on global attributes through models like TimeVAE \cite{TimeVAE}, TimeGAN \cite{TimeGAN}, and C-TimeGAN \cite{CGAN}. Recent advances include hierarchical approaches \cite{Torres2021} and attention-based mechanisms \cite{Liu2024}. Notably, CGAN-TS \cite{GANTS} and ControlTS \cite{Wang2024} introduced attribute-based and temporal feature control.

\subsection{Diffusion Models for Time Series}
The adaptation of diffusion models to time series data has seen rapid advancement through several key developments. Early works like TimeGrad \cite{Rasul2021} and CSDI \cite{CSDI} established the viability of diffusion models for temporal data, particularly in handling missing value imputation and uncertainty quantification. These foundations led to architectural innovations including TimeDiT \cite{TimeDiT}, which introduced specialized temporal attention mechanisms, and Diff-MTS \cite{Diff-MTS}, which enhanced multivariate time series generation through improved cross-channel modelling.

Recent advances have focused on both architectural improvements and control mechanisms. The Latent Diffusion Transformer \cite{Feng2024} demonstrated efficient generation through compressed latent spaces, while RATD \cite{Liu2024b} introduced robust attention mechanisms for handling temporal dependencies. Score-CDM \cite{ScoreCDM} and DiffusionBridge \cite{DiffusionBridge} advanced controlled generation through score-based approaches and bridge construction methods respectively.
These developments have enabled successful applications across diverse domains, from healthcare monitoring \cite{ECG} and industrial systems \cite{Tian2024} to financial forecasting \cite{Microsoft,Hamdouche2023}.

While existing methods show promise, they primarily focus on global trend control and distribution matching, lacking, fine-grained control at the individual timestamp level except DiffTime and Guided-DiffTime \cite{ConstrainedTSG}, working on the point-wise constrained time series generation. However, their points control needs additional training.

   \vspace{-0.5em}

\section{Conclusion}
\vspace{-0.5em}
We present the \method{} for generalized Time Series Editing that enables fine-grained control and global statistical manipulation without model retraining. It is achieved via float masking and score-based guidance for statistical properties. The interleaving nature of the two mechanisms allows them to be seamlessly combined, enabling sophisticated editing operations that respect both fine-grained constraints and coarse-grained properties. This makes our framework particularly suitable for real-world applications where domain experts need to incorporate specific knowledge while maintaining overall statistical validity. 
Figure \ref{fig:editor} demonstrates the practical interactive editing interface to enable the intuitive manipulation of temporal data like Photoshop.
The control mechanism is not perfect and the obvious distribution drift still exists. 
Our method can be further improved by incorporating more advanced control mechanisms and carefully considering the trade-off between control precision and distribution preservation.

\bibliography{custom}

\begin{thebibliography}{10}

\bibitem{ECG}
Edmonmd Adib, Amanda~S. Fernandez, Fatemeh Afghah, and John~J. Prevost.
\newblock Synthetic ecg signal generation using probabilistic diffusion models.
\newblock {\em IEEE Access}, 11:75818--75828, 2024.

\bibitem{Alcaraz2022}
Juan Miguel~Lopez Alcaraz and Nils Strodthoff.
\newblock Diffusion-based time series imputation and forecasting with structured state space models.
\newblock {\em Version published by Transactions on Machine Learning Research in 2022 (TMLR ISSN 2835-8856) https://openreview.net/forum?id=hHiIbk7ApW}, August 2022.

\bibitem{Ang2023}
Yihao Ang, Qiang Huang, Yifan Bao, Anthony K.~H. Tung, and Zhiyong Huang.
\newblock Tsgbench: Time series generation benchmark.
\newblock September 2023.

\bibitem{CGAN}
Gaby Baasch, Guillaume Rousseau, and Ralph Evins.
\newblock A conditional generative adversarial network for energy use in multiple buildings using scarce data.
\newblock {\em Energy and AI}, 5:100087, 2021.

\bibitem{CTSG}
Yifan Bao, Yihao Ang, Qiang Huang, Anthony K.~H. Tung, and Zhiyong Huang.
\newblock Towards controllable time series generation.
\newblock March 2024.

\bibitem{TimeDiT}
Defu Cao, Wen Ye, Yizhou Zhang, and Yan Liu.
\newblock Timedit: General-purpose diffusion transformers for time series foundation model.
\newblock September 2024.

\bibitem{ConstrainedTSG}
Andrea Coletta, Sriram Gopalakrishnan, Daniel Borrajo, and Svitlana Vyetrenko.
\newblock On the constrained time-series generation problem.
\newblock In A.~Oh, T.~Naumann, A.~Globerson, K.~Saenko, M.~Hardt, and S.~Levine, editors, {\em Advances in Neural Information Processing Systems}, volume~36, pages 61048--61059. Curran Associates, Inc., 2023.

\bibitem{TimeVAE}
Abhyuday Desai, Cynthia Freeman, Zuhui Wang, and Ian Beaver.
\newblock Timevae: A variational auto-encoder for multivariate time series generation.
\newblock November 2021.

\bibitem{RCGAN}
Cristóbal Esteban, Stephanie~L. Hyland, and Gunnar Rätsch.
\newblock Real-valued (medical) time series generation with recurrent conditional gans.
\newblock June 2017.

\bibitem{Feng2024}
Shibo Feng, Chunyan Miao, Zhong Zhang, and Peilin Zhao.
\newblock Latent diffusion transformer for probabilistic time series forecasting.
\newblock {\em Proceedings of the AAAI Conference on Artificial Intelligence}, 38(11):11979--11987, Mar. 2024.

\bibitem{Goodfellow2016}
Ian Goodfellow.
\newblock Nips 2016 tutorial: Generative adversarial networks.
\newblock {\em arXiv preprint arXiv:1701.00160}, 2016.

\bibitem{Hamdouche2023}
Mohamed Hamdouche, Pierre Henry-Labordere, and Huyên Pham.
\newblock Generative modeling for time series via schr{ö}dinger bridge.
\newblock April 2023.

\bibitem{DDPM}
Jonathan Ho, Ajay Jain, and Pieter Abbeel.
\newblock Denoising diffusion probabilistic models.
\newblock June 2020.

\bibitem{FM-TS}
Yang Hu, Xiao Wang, Lirong Wu, Huatian Zhang, Stan~Z. Li, Sheng Wang, and Tianlong Chen.
\newblock Fm-ts: Flow matching for time series generation.
\newblock November 2024.

\bibitem{TETS}
Baoyu Jing, Shuqi Gu, Tianyu Chen, Zhiyu Yang, Dongsheng Li, Jingrui He, and Kan Ren.
\newblock Towards editing time series.
\newblock In {\em The Thirty-eighth Annual Conference on Neural Information Processing Systems}, 2024.

\bibitem{Diffwave}
Zhifeng Kong, Wei Ping, Jiaji Huang, Kexin Zhao, and Bryan Catanzaro.
\newblock Diffwave: A versatile diffusion model for audio synthesis.
\newblock {\em arXiv preprint arXiv:2009.09761}, 2020.

\bibitem{TimeVQVAE}
Daesoo Lee, Sara Malacarne, and Erlend Aune.
\newblock Vector quantized time series generation with a bidirectional prior model.
\newblock March 2023.

\bibitem{Li2022}
Yan Li, Xinjiang Lu, Yaqing Wang, and Dejing Dou.
\newblock Generative time series forecasting with diffusion, denoise, and disentanglement.
\newblock In S.~Koyejo, S.~Mohamed, A.~Agarwal, D.~Belgrave, K.~Cho, and A.~Oh, editors, {\em Advances in Neural Information Processing Systems}, volume~35, pages 23009--23022. Curran Associates, Inc., 2022.

\bibitem{Liu2024b}
Jingwei Liu, Ling Yang, Hongyan Li, and Shenda Hong.
\newblock Retrieval-augmented diffusion models for time series forecasting.
\newblock October 2024.

\bibitem{Liu2024}
Xinhe Liu and Wenmin Wang.
\newblock Deep time series forecasting models: A comprehensive survey.
\newblock {\em Mathematics}, 12(10), 2024.

\bibitem{TSDiffusionSurvey}
Caspar Meijer and Lydia~Y. Chen.
\newblock The rise of diffusion models in time-series forecasting.
\newblock January 2024.

\bibitem{GANTS}
Xiaoye Miao, Yangyang Wu, Jun Wang, Yunjun Gao, Xudong Mao, and Jianwei Yin.
\newblock Generative semi-supervised learning for multivariate time series imputation.
\newblock In {\em Proceedings of the AAAI conference on artificial intelligence}, volume~35, pages 8983--8991, 2021.

\bibitem{TimeWeaver}
Sai~Shankar Narasimhan, Shubhankar Agarwal, Oguzhan Akcin, Sujay Sanghavi, and Sandeep Chinchali.
\newblock Time weaver: A conditional time series generation model.
\newblock March 2024.

\bibitem{Nguyen2021}
Nam Nguyen and Brian Quanz.
\newblock Temporal latent auto-encoder: A method for probabilistic multivariate time series forecasting.
\newblock In {\em Proceedings of the AAAI conference on artificial intelligence}, volume~35, pages 9117--9125, 2021.

\bibitem{DiffusionBridge}
Jinseong Park, Seungyun Lee, Woojin Jeong, Yujin Choi, and Jaewook Lee.
\newblock Leveraging priors via diffusion bridge for time series generation.
\newblock August 2024.

\bibitem{VAE}
Lucas Pinheiro~Cinelli, Matheus Ara{\'u}jo~Marins, Eduardo~Ant{\'u}nio Barros~da Silva, and S{\'e}rgio Lima~Netto.
\newblock Variational autoencoder.
\newblock In {\em Variational Methods for Machine Learning with Applications to Deep Networks}, pages 111--149. Springer, 2021.

\bibitem{Rasul2021}
Kashif Rasul, Calvin Seward, Ingmar Schuster, and Roland Vollgraf.
\newblock Autoregressive denoising diffusion models for multivariate probabilistic time series forecasting.
\newblock In Marina Meila and Tong Zhang, editors, {\em Proceedings of the 38th International Conference on Machine Learning}, volume 139 of {\em Proceedings of Machine Learning Research}, pages 8857--8868. PMLR, 18--24 Jul 2021.

\bibitem{Diff-MTS}
Lei Ren, Haiteng Wang, and Yuanjun Laili.
\newblock Diff-mts: Temporal-augmented conditional diffusion-based aigc for industrial time series toward the large model era.
\newblock {\em IEEE Transactions on Cybernetics}, 54(12):7187--7197, 2024.

\bibitem{Sommers2024}
Alexander Sommers, Logan Cummins, Sudip Mittal, Shahram Rahimi, Maria Seale, Joseph Jaboure, and Thomas Arnold.
\newblock A survey of transformer enabled time series synthesis.
\newblock June 2024.

\bibitem{DDIM}
Jiaming Song, Chenlin Meng, and Stefano Ermon.
\newblock Denoising diffusion implicit models.
\newblock October 2020.

\bibitem{ScoreBased}
Yang Song, Jascha Sohl-Dickstein, Diederik~P. Kingma, Abhishek Kumar, Stefano Ermon, and Ben Poole.
\newblock Score-based generative modeling through stochastic differential equations.
\newblock November 2020.

\bibitem{CSDI}
Yusuke Tashiro, Jiaming Song, Yang Song, and Stefano Ermon.
\newblock Csdi: Conditional score-based diffusion models for probabilistic time series imputation.
\newblock In M.~Ranzato, A.~Beygelzimer, Y.~Dauphin, P.S. Liang, and J.~Wortman Vaughan, editors, {\em Advances in Neural Information Processing Systems}, volume~34, pages 24804--24816. Curran Associates, Inc., 2021.

\bibitem{Tian2024}
Muhang Tian, Bernie Chen, Allan Guo, Shiyi Jiang, and Anru~R Zhang.
\newblock Reliable generation of privacy-preserving synthetic electronic health record time series via diffusion models.
\newblock {\em Journal of the American Medical Informatics Association}, 31(11):2529--2539, 09 2024.

\bibitem{Torres2021}
José~F. Torres, Dalil Hadjout, Abderrazak Sebaa, Francisco Martínez-Álvarez, and Alicia Troncoso.
\newblock Deep learning for time series forecasting: A survey.
\newblock {\em Big Data}, 9(1):3--21, February 2021.

\bibitem{Wang2024}
Yuxuan Wang, Haixu Wu, Jiaxiang Dong, Yong Liu, Mingsheng Long, and Jianmin Wang.
\newblock Deep time series models: A comprehensive survey and benchmark.
\newblock July 2024.

\bibitem{Wen2023}
Haomin Wen, Youfang Lin, Yutong Xia, Huaiyu Wan, Qingsong Wen, Roger Zimmermann, and Yuxuan Liang.
\newblock Diffstg: Probabilistic spatio-temporal graph forecasting with denoising diffusion models.
\newblock In {\em Proceedings of the 31st ACM International Conference on Advances in Geographic Information Systems}, pages 1--12, 2023.

\bibitem{Wiese2020}
Magnus Wiese, Robert Knobloch, Ralf Korn, and Peter Kretschmer.
\newblock Quant gans: deep generation of financial time series.
\newblock {\em Quantitative Finance}, 20(9):1419--1440, 2020.

\bibitem{Microsoft}
Fangkai Yang, Wenjie Yin, Lu~Wang, Tianci Li, Pu~Zhao, Bo~Liu, Paul Wang, Bo~Qiao, Yudong Liu, Mårten Björkman, Saravan Rajmohan, Qingwei Lin, and Dongmei Zhang.
\newblock Diffusion-based time series data imputation for cloud failure prediction at microsoft 365.
\newblock In {\em Proceedings of the 31st ACM Joint European Software Engineering Conference and Symposium on the Foundations of Software Engineering}, ESEC/FSE ’23, pages 2050--2055. ACM, November 2023.

\bibitem{TimeGAN}
Jinsung Yoon, Daniel Jarrett, and Mihaela van~der Schaar.
\newblock Time-series generative adversarial networks.
\newblock In H.~Wallach, H.~Larochelle, A.~Beygelzimer, F.~d\textquotesingle Alch\'{e}-Buc, E.~Fox, and R.~Garnett, editors, {\em Advances in Neural Information Processing Systems}, volume~32. Curran Associates, Inc., 2019.

\bibitem{Diffusion-TS}
Xinyu Yuan and Yan Qiao.
\newblock Diffusion-ts: Interpretable diffusion for general time series generation.
\newblock March 2024.

\bibitem{ScoreCDM}
S.~Zhang, S.~Wang, H.~Miao, H.~Chen, C.~Fan, and J.~Zhang.
\newblock Score-cdm: Score-weighted convolutional diffusion model for multivariate time series imputation.
\newblock May 2024.

\bibitem{DDBM}
Linqi Zhou, Aaron Lou, Samar Khanna, and Stefano Ermon.
\newblock Denoising diffusion bridge models.
\newblock September 2023.

\end{thebibliography}
\bibliographystyle{plain}

\appendix

\newpage
\section{Methodology Additional Details}
\subsection{Summary of Variants}
\label{app:notation}

\begin{table}[h]
\centering
\caption{Summary of variables and their Meanings}
\begin{tabular}{ll}
\hline
\textbf{Symbol} & \textbf{Meaning} \\
\hline
$N$ & Number of time series \\
$\mathbf{x}_i \in \mathbb{R}^{L \times D}$ & $i$-th time series, length $L$, dimension $D$ \\
$S = \{\mathbf{x}_i\}_{i=1}^N$ & Training dataset of time series \\
$\theta$ & Model parameters of $f_\theta$ \\
$\mathbf{C}$ & Prior/conditioning input \\
$T$ & Total forward/backward diffusion steps \\
$K$ & Set of number of gradient steps per diffusion step \\
$\beta_k \in [0,1]$ & Noise variance schedule \\
$\mathbf{x}_t$ & Noisy sample at diffusion step $t$ \\
$\mathbf{t}_i$ & Time index in the sample $\mathbf{x}$ \\
$\mathbf{v}_i$ & Corresponding $\mathbf{y}$ value of the time index in sample $\mathbf{x}$ \\
$\mu_\theta, \sigma_k^2$ & Mean and variance in reverse diffusion \\
$\Omega(\mathbf{x}), \overline{\Omega}(\mathbf{x})$ & Observed and missing indices \\
$\mathbf{x}_{\text{ob}}, \mathbf{x}_{\text{ta}}$ & Observed and target parts of a time series \\
$\mathbf{m} \in [0,1]^{L \times D}$ & Confidence/float mask \\
$\omega_t$ & Time-dependent weight in reverse steps \\
$\mathbf{C}_{\text{point}}$ & Point-wise control with confidence \\
$\mathbf{C}_{\text{segment}}$ & Segment-wise control constraints \\
$\mathcal{L}_{\mathrm{sum}}, \mathcal{L}_{\mathrm{pen}}, \mathcal{L}_{\mathrm{statistics}}$ & Loss terms enforcing constraints \\
$\gamma, \eta, \beta_{\mathrm{sum}[s_j:e_j]}$ & Additional scale factors for guidance \\
\hline
\end{tabular}
\end{table}

\subsection{Formal Proof of Confidence-Based Masking}
\label{app:float-mask-proof}
\begin{proof}
Let $p(x)$ be the jointly trained diffusion model over both observed and missing dimensions, and let $p(x_{ta} \mid x_{ob})$ be the target conditional distribution. Denote $p_t(x_{ob}^t, x_{ta}^t)$ as the distribution of $(x_{ob}^t, x_{ta}^t)$ at iteration $t$. We show that as $t \to 0$, $p_t(x_{ta}^t \mid x_{ob}^t)$ converges to $p(x_{ta} \mid x_{ob})$.

\textbf{I. Forward-Process Marginals.}  
By construction, $x_{ob}^t$ at each step is drawn from the forward process $q(x_{ob}^t \mid x_{ob}^0)$, which is a Gaussian transition that preserves the exact marginal of the known dimensions $x_{ob}$. Formally, for each $t$,  
\[
x_{ob}^t \sim \mathcal{N}\bigl(\sqrt{\bar{\alpha}_t}\,x_{ob}^0,\,(1-\bar{\alpha}_t)\mathbf{I}\bigr).
\]
Hence, $p_t(x_{ob}^t)$ remains consistent with the correct marginal distribution $\prod_{\Omega(x)} p(x_{ob})$.

\textbf{II. Denoising of Missing Entries.}  
The reverse step for missing entries $x_{ta}^t$ is governed by 
\[
x_{ta}^t \leftarrow x_{ta}^{t+1} - \sqrt{\beta_t}\,\nabla_{x_{ta}^{t+1}}
\log p_{\theta}(x_{ta}^{t+1}\mid x_{ta}^{t+2}, x_{ob}^t).
\]
As shown in \cite{ScoreBased}, iteratively applying the reverse diffusion steps in this score-based framework converges to sampling from $p(x_{ta}\mid x_{ob}^t)$.

\textbf{III. Replace and Float-Mask Consistency.}  
Whether we \emph{replace} $x_{ob}^t$ completely or \emph{blend} it with a float mask $\mathbf{m}$:
\[
x^t \;=\; \mathbf{m} \odot x_{ob}^t + (1 - \mathbf{m}) \odot x_{ta}^t,
\]
the observed indices remain consistent with their forward-sampled values. This ensures that at every iteration, the joint distribution respects the known-data constraints. The $0.0$ value in masks means no restriction on this point, and $1.0$ means the fixed point.

\textbf{IV. Convergence to the Conditional.}  
Consider $p_t(x_{ta}^t \mid x_{ob}^t)$. By the score-based argument (the forward-reverse chain forming a time-indexed Markov process), the mixture of denoising steps and partial resets of observed entries yields
\[
\lim_{t \to 0} p_t(x_{ta}^t \mid x_{ob}^t) \;=\; p(x_{ta}\mid x_{ob}),
\]
where the convergence follows from the fact that each reverse diffusion step corrects the noise injected in the forward pass, conditioned on the known $x_{ob}^t$.

Thus, replace-based or float-mask imputation each preserves $x_{ob}$’s marginals and iteratively refine $x_{ta}$ until the distribution of the missing entries matches $p(x_{ta}\mid x_{ob})$. 
\end{proof}

\paragraph{Implementation Details} The float mask $\mathbf{m}$ can be adjusted to control the influence of observed data on the imputed sequence. By iterative sampling from the diffusion model and applying the floating mask, we can generate time series data that respects observed values while conforming to the correct conditional distribution.

\vspace{-1em}
\begin{enumerate}
    \item \textbf{Initialize} $x^T$ from Gaussian noise.
    \item \textbf{For} $t = T$ down to $1$:
    \begin{itemize}
        \item \emph{Forward-process sampling (for $x_{ob}^t$):} $x_{ob}^t \sim q\bigl(x_{ob}^t \mid x_{ob}^0\bigr)$.
        \item \emph{Denoise missing parts (for $x_{ta}^t$):} perform gradient step using $\log p_{\theta}$.
        \item \emph{Apply float mask (optional):} fuse observed and imputed regions via $\mathbf{m}$.
    \end{itemize}
\end{enumerate}

\subsection{Time-Dependent Weight}
\label{app:time_dependent}
\begin{figure}[htb]
    \centering
    \includegraphics[width=0.5\textwidth]{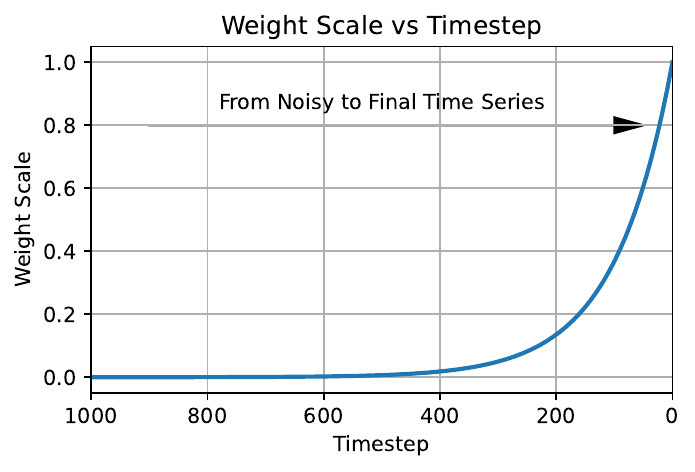}
    \vspace{-1em}
    \caption{Time-dependent weight schedule during denoising process. The exponential decay ($\gamma = 5.0$) provides stronger control signals in later timesteps while allowing smoother adjustments near completion.}
    \label{fig:weight_scale}
\end{figure}

\subsection{\method{} Algorithm}

\begin{algorithm}[H] 
\caption{\small DDPM Denoising with \method{}}
\label{alg:sampling}
\small
\begin{algorithmic}[1]
\REQUIRE Gradient scale $\eta$, trade-off coefficient $\gamma$, conditional data $\mathbf{x}_a$, time-dependent weight $\omega_t$
\STATE Initialize $\mathbf{x}_T \sim \mathcal{N}(\mathbf{0}, \mathbf{I})$
\FOR{$t = T$ to $1$}
    \STATE \textbf{// Step 1:} Predict and Refine Sample
    \STATE $[\hat{\mathbf{x}}_{a}, \hat{\mathbf{x}}_{b}] \leftarrow p_\theta(\mathbf{x}_t, t, \theta)$
    \STATE $\mathcal{L}_1 = \|\mathbf{x}_a - \hat{\mathbf{x}}_{a}\|_2^2$
    \STATE $\mathbf{x}_{t-1} \leftarrow \mathcal{N}\big(\mu(p_\theta(\mathbf{x}_t, t, \theta), \mathbf{x}_t), \Sigma\big)$
    \STATE $\mathcal{L}_2 = \|\mathbf{x}_{t-1} - \mu(p_\theta(\mathbf{x}_t, t, \theta), \mathbf{x}_t)\|_2^2 / \Sigma$
    \STATE \textbf{// Step 2:} Statistics Control
    \STATE $\mathcal{L}_{statistics} = \omega_t \sum \beta_{\mathrm{sum} [s_j:e_j]} \mathcal{L}_{\mathrm{sum} [s_j:e_j]}$
    \STATE $\tilde{\mathbf{x}}_0 = p_\theta(\mathbf{x}_t, t, \theta) + \eta \nabla_{\mathbf{x}_t}(\mathcal{L}_1 + \gamma \mathcal{L}_2 + \mathcal{L}_{statistics})$
    \STATE $\mathbf{x}_{t-1} \leftarrow \mathcal{N}\big(\mu(\tilde{\mathbf{x}}_0, \mathbf{x}_t), \Sigma\big)$
    \STATE \textbf{// Step 3:} Point-Wise Control
    \STATE $\mathbf{x}_{t-1} \gets \omega_t \mathbf{m}^{\text{ob}} \odot \mathbf{x}^{\text{ob}}_{t} + (1 - \omega_t\mathbf{m}^{\text{ob}}) \odot \mathbf{x}_{t-1}$
\ENDFOR
\end{algorithmic}
\end{algorithm}

\newpage

\section{Experiment Additional Details}

\label{app:train}
\begin{table}[h]
\centering
\caption{Training hyperparameters and settings for each dataset.}
\label{tab:expr_setup}
\resizebox{0.8\columnwidth}{!}{%
\begin{tabular}{l|ccc|ccc|c}
\toprule
\multicolumn{1}{c|}{\multirow{2}{*}{Method}} & \multicolumn{3}{c|}{Dataset} & \multicolumn{3}{c|}{Training} & \multicolumn{1}{c}{Inference} \\
\multicolumn{1}{c|}{} & Name & Seq Length & Features & \multicolumn{1}{c}{LR} & \multicolumn{1}{c}{Train Steps} & \multicolumn{1}{c|}{Batch Szie} & \multicolumn{1}{l}{DDPM Timesteps} \\ \midrule
\multirow{4}{*}{Our - CSDI} & Energy & 24 & 28 & 1.00E-03 & 25000 & 64 & 200 \\
    & fMRI & 24 & 50 & 1.00E-03 & 15000 & 64 & 200 \\
    & Revenue & 365 & 3 & 2.00E-03 & 2230 & 64 & 200 \\
    & Sines & 24 & 5 & 1.00E-03 & 12000 & 128 & 50 \\ \midrule
\multirow{4}{*}{Our - Diffusion-TS} & Energy & 24 & 28 & 1.00E-05 & 25000 & 64 & 1000 \\
    & fMRI & 24 & 50 & 1.00E-05 & 15000 & 64 & 1000 \\
    & Revenue & 365 & 3 & 2.00E-05 & 2230 & 64 & 500 \\
    & Sines & 24 & 5 & 1.00E-05 & 12000 & 128 & 500 \\ \bottomrule
\end{tabular}%
}
\end{table}

We run all training and inferencing of all experiments with NVIDIA L40S GPUs. All experiments are fixed on the random seed with 2024. Table \ref{tab:expr_setup} summarizes the training hyperparameters for each dataset and method. We use the Adam optimizer with a learning rate of $1.00 \times 10^{-3}$ for CSDI and $1.00 \times 10^{-5}$ for Diffusion-TS. The training steps are set to 25,000 for CSDI and 15,000 for Diffusion-TS. The batch size is 64 for CSDI and 128 for Diffusion-TS. The number of diffusion steps is set according to the different datasets.

For point-wise control, we placed target points (called ``anchor'' in the following) at normalized values $\mathbf{v}_{anchor} \in \{0.1, 0.8, 1.0\}$ at relative temporal positions $\mathbf{t}_{anchor} \in \{0.1L, 0.3L, 0.5L, 0.7L, 0.9L\}$. The confidence levels are set to $\mathbf{c}_{anchor} = \{0.01, 0.50, 1.0\}$ at the corresponding anchor indices, and the other positions were set to 0.0 during mask converting.

For the segment-wise control, we test on segments $\{(s, e)\} = $$\{(0.2L, 0.4L), (0.4L, 0.6L), (0.6L, 0.8L),\\ (0, L)\}$, $(0, L)$ represents the whole sequence. We choose ``sum'' as the aggregated function, then $\alpha_{sum} \in \{-100, 20, 50 , 100\}$ as the targeted aggregated statistics, weight $\beta_{sum} \in \{1, 10, 50, 100\}$.
We note that baseline sums vary significantly between datasets due to sequence length and data scaling, we will include the various sequence length results, such $L \in \{96, 192, 384\}$ in the camera-ready version.

\section{Point-Wise Control}

\subsection{Demonstrates}
\label{app:anchor_demo}
The following figures demonstrate the effectiveness of point-wise control across different datasets varying this confidence and target value with Diffusion-TS and CSDI. We can see as the confidence increases, the model respects the anchor points more strictly. 
\subsubsection{Pure Float Mask Control (Diffusion-TS)}
\begin{figure*}[!htbp]
\centering
\includegraphics[width=0.75\linewidth]{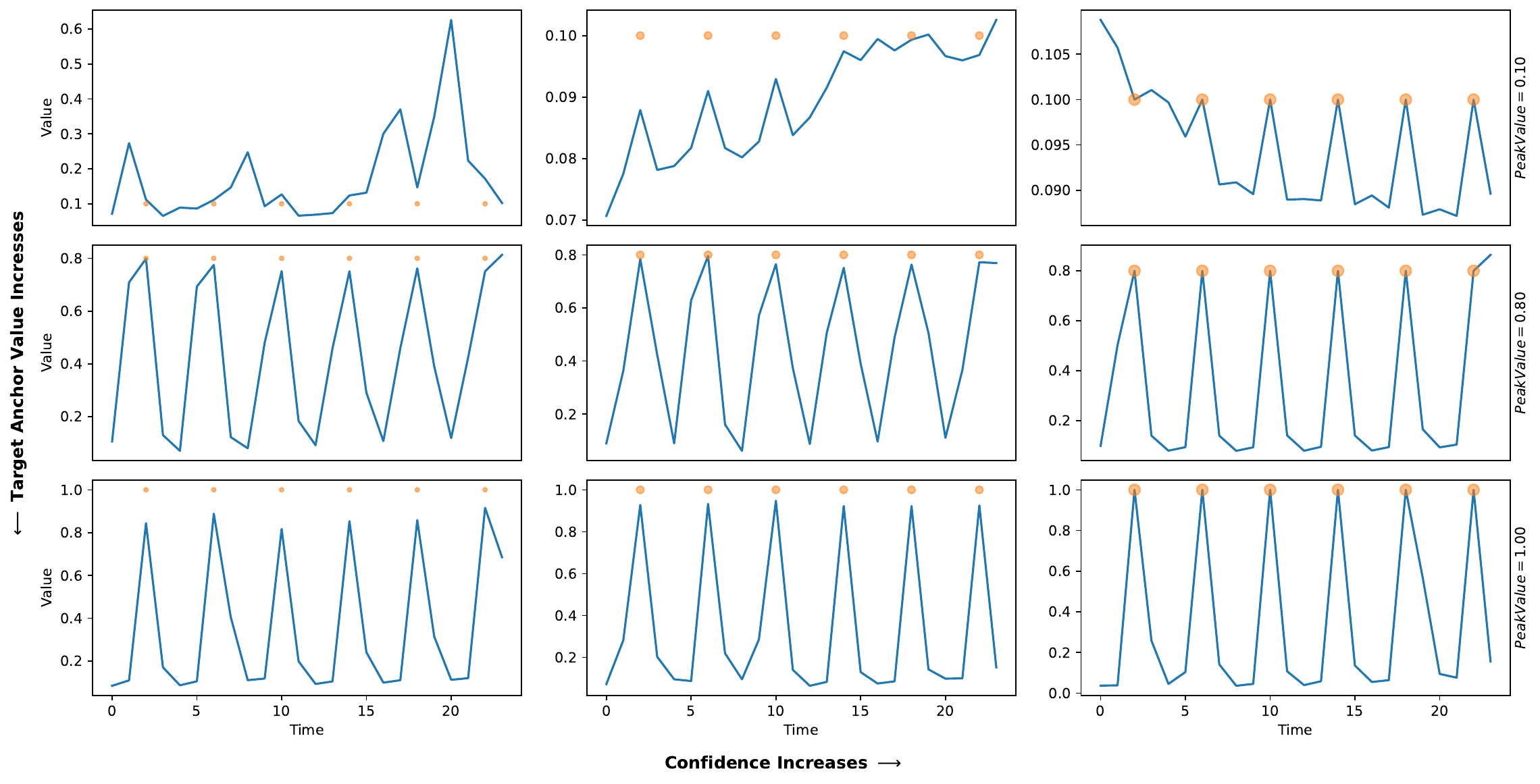}
\vspace{-0.3cm}
\caption{Demonstration of Point-Wise Control in ETTh datasets with multiple control points and confidences}
\end{figure*}
\begin{figure*}[!htbp]
\centering
\includegraphics[width=0.75\linewidth]{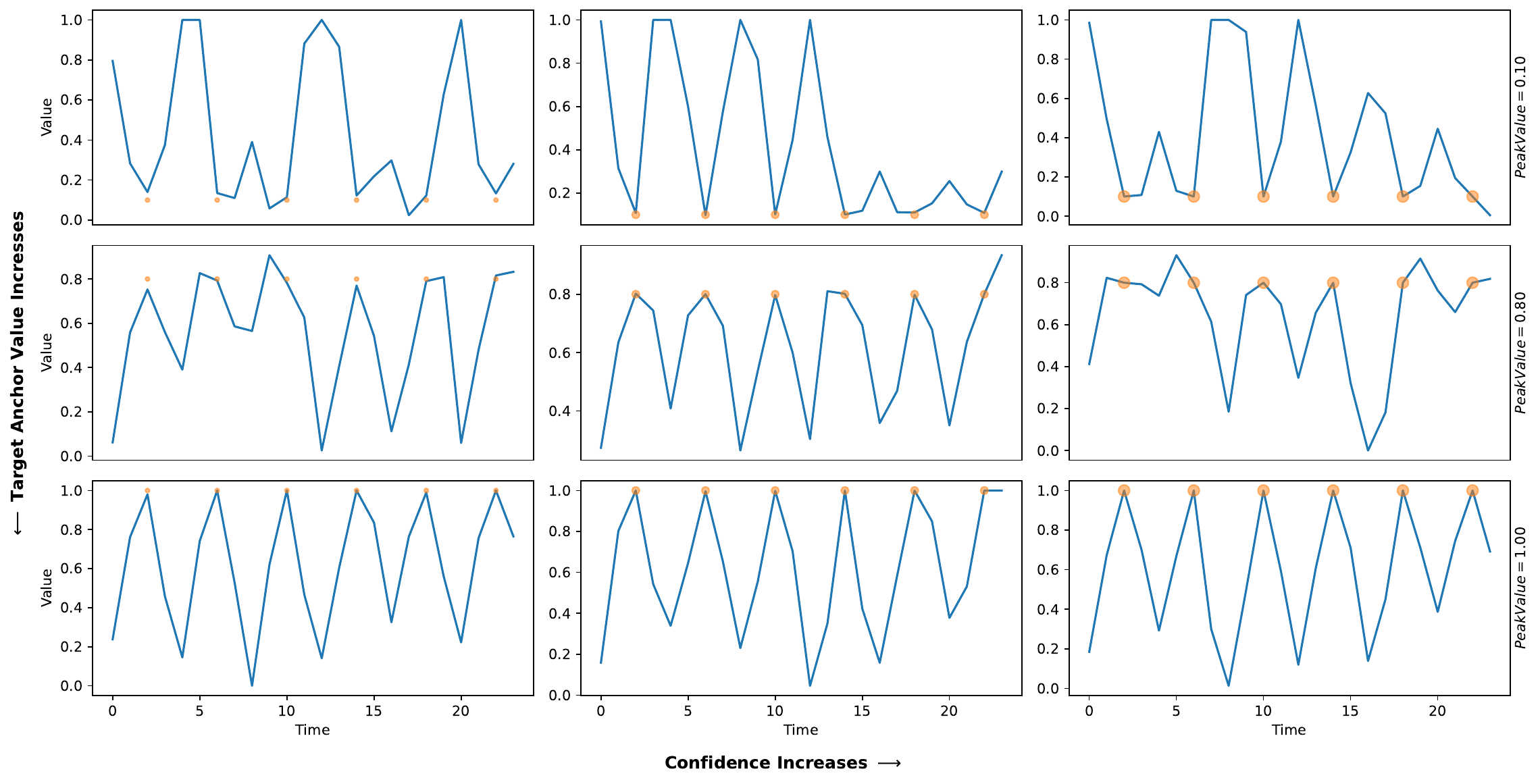}
    \vspace{-0.3cm}
    \caption{Demonstration of Point-Wise Control in fMRI datasets with multiple control points and confidences.}
\end{figure*}

\begin{figure*}[!htbp]
\centering
\includegraphics[width=0.75\linewidth]{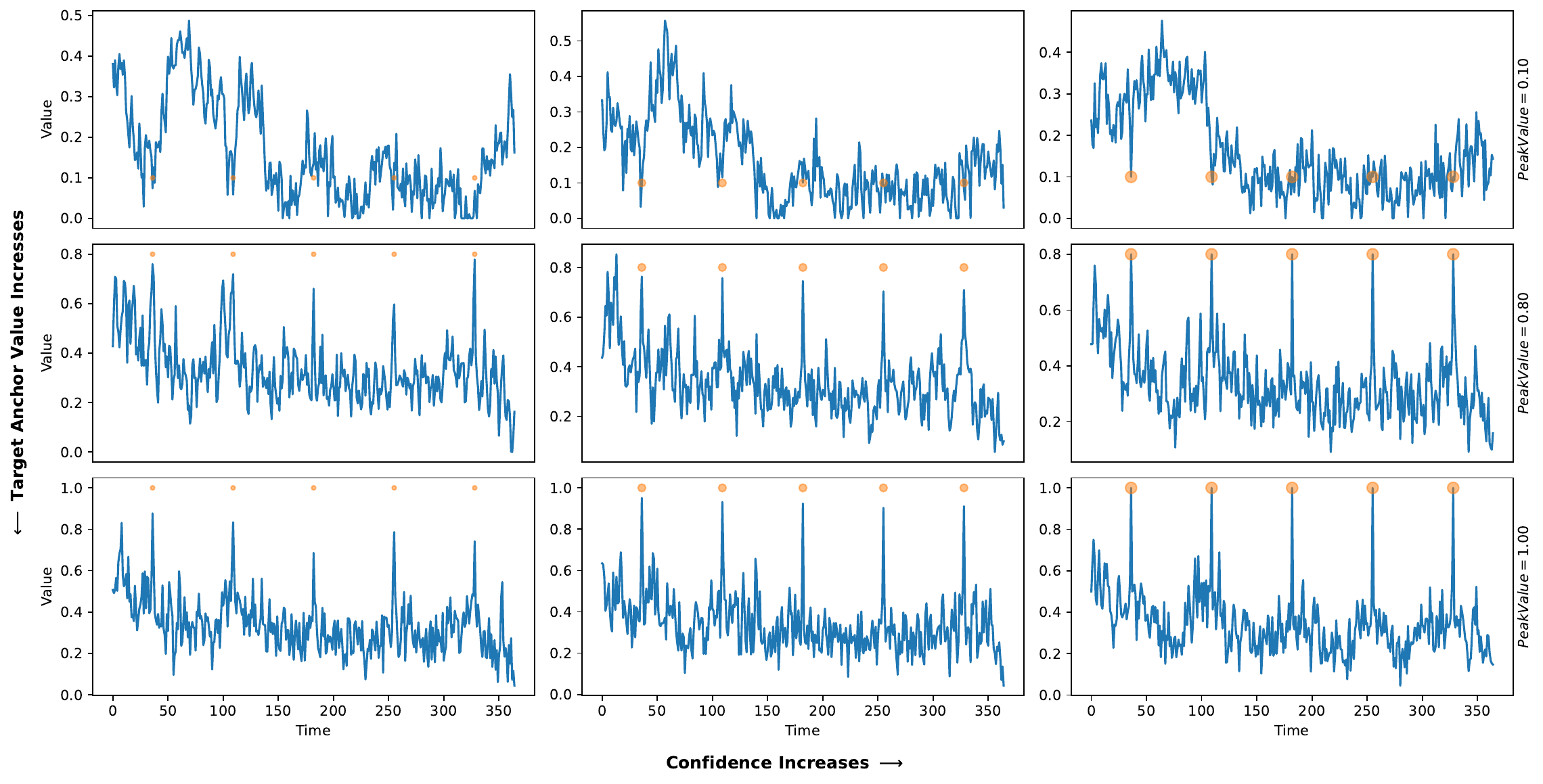}
    \vspace{-0.3cm}
    \caption{Demonstration of Point-Wise Control in Revenue datasets with multiple control points and confidences}
\end{figure*}

\begin{figure*}[!htbp]
\centering
\includegraphics[width=0.75\linewidth]{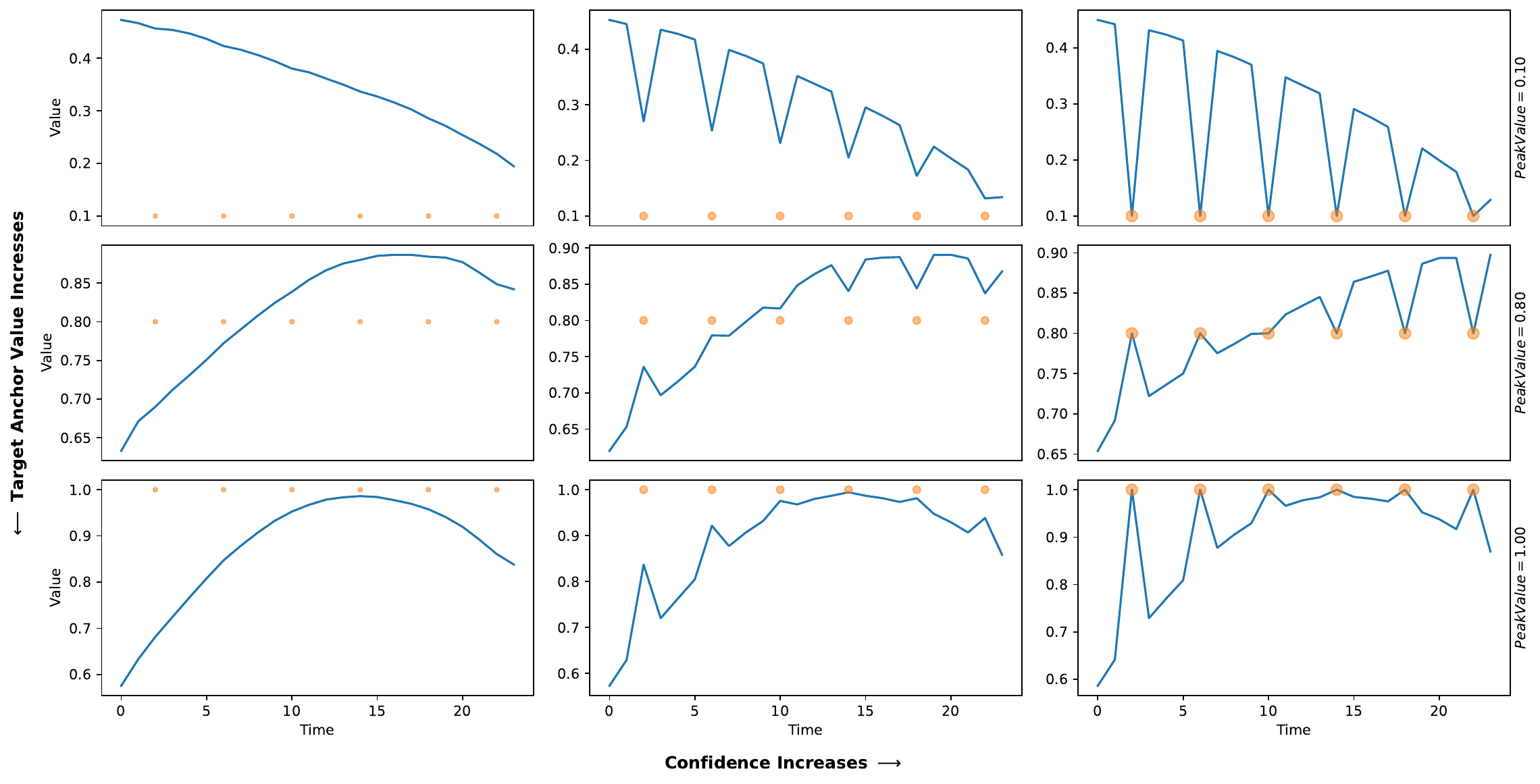}
\vspace{-0.3cm}
\caption{Demonstration of Point-Wise Control in Synthetic sine wave datasets with multiple control points and confidences}

\end{figure*}

\newpage
\subsubsection{Float Mask Control with Extensions (Diffusion-TS)}
\begin{figure*}[!htbp]
\centering
\includegraphics[width=0.75\linewidth]{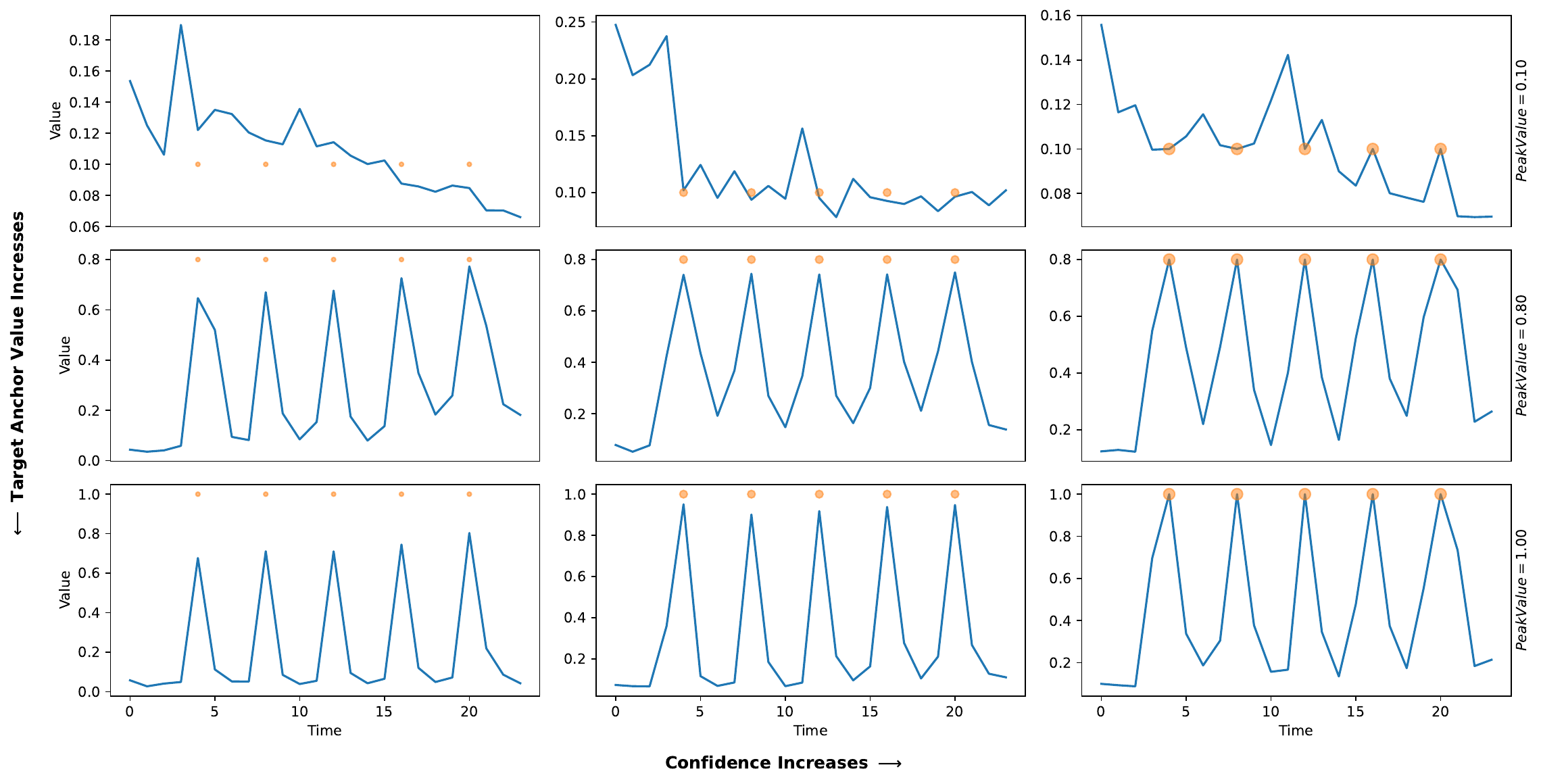}
\vspace{-0.3cm}
\caption{Demonstration of Point-Wise Control in ETTh datasets with multiple control points and confidences}
\end{figure*}
\begin{figure*}[!htbp]
\centering
\includegraphics[width=0.75\linewidth]{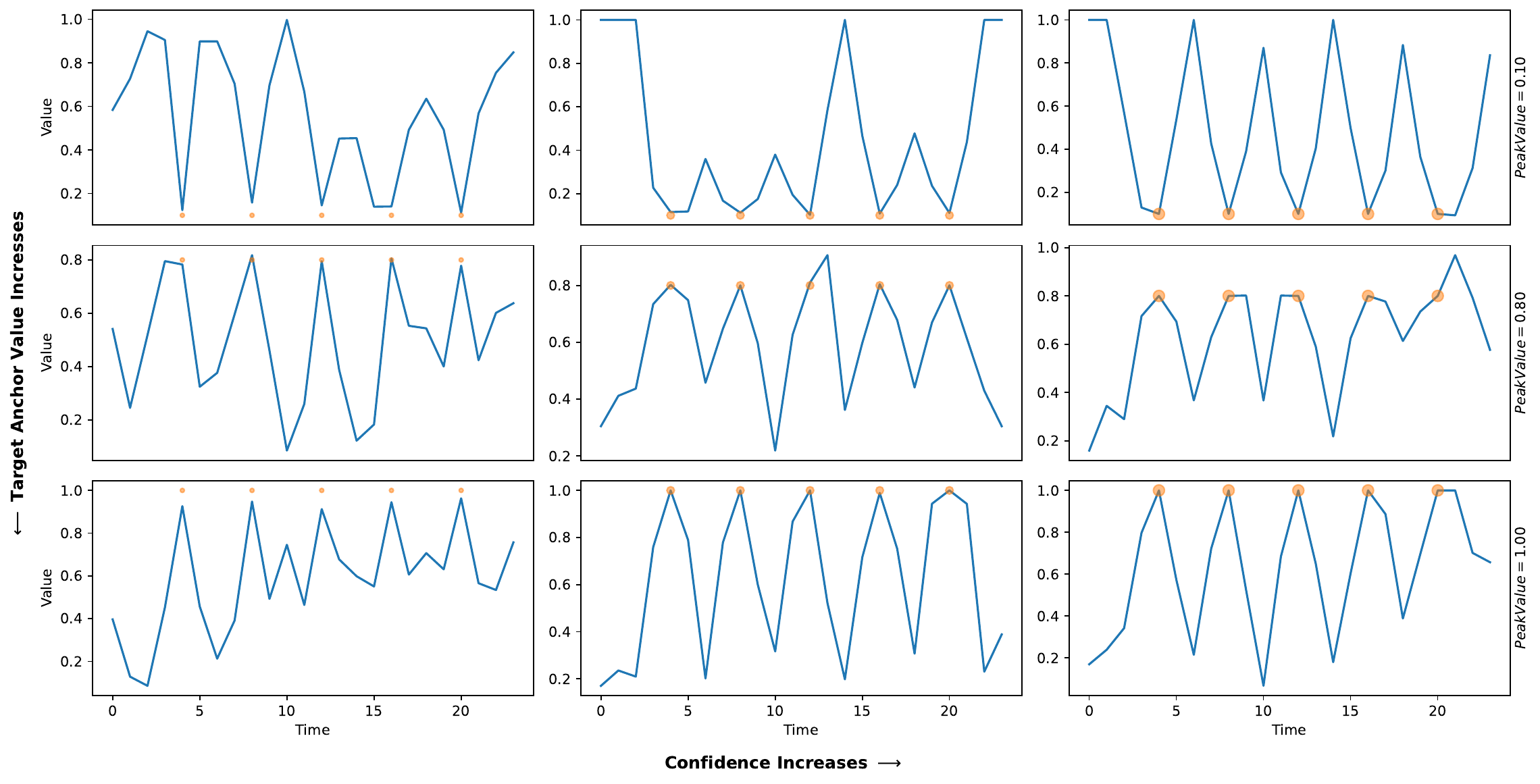}
    \vspace{-0.3cm}
    \caption{Demonstration of Point-Wise Control in fMRI datasets with multiple control points and confidences.}
\end{figure*}

\begin{figure*}[!htbp]
\centering
\includegraphics[width=0.75\linewidth]{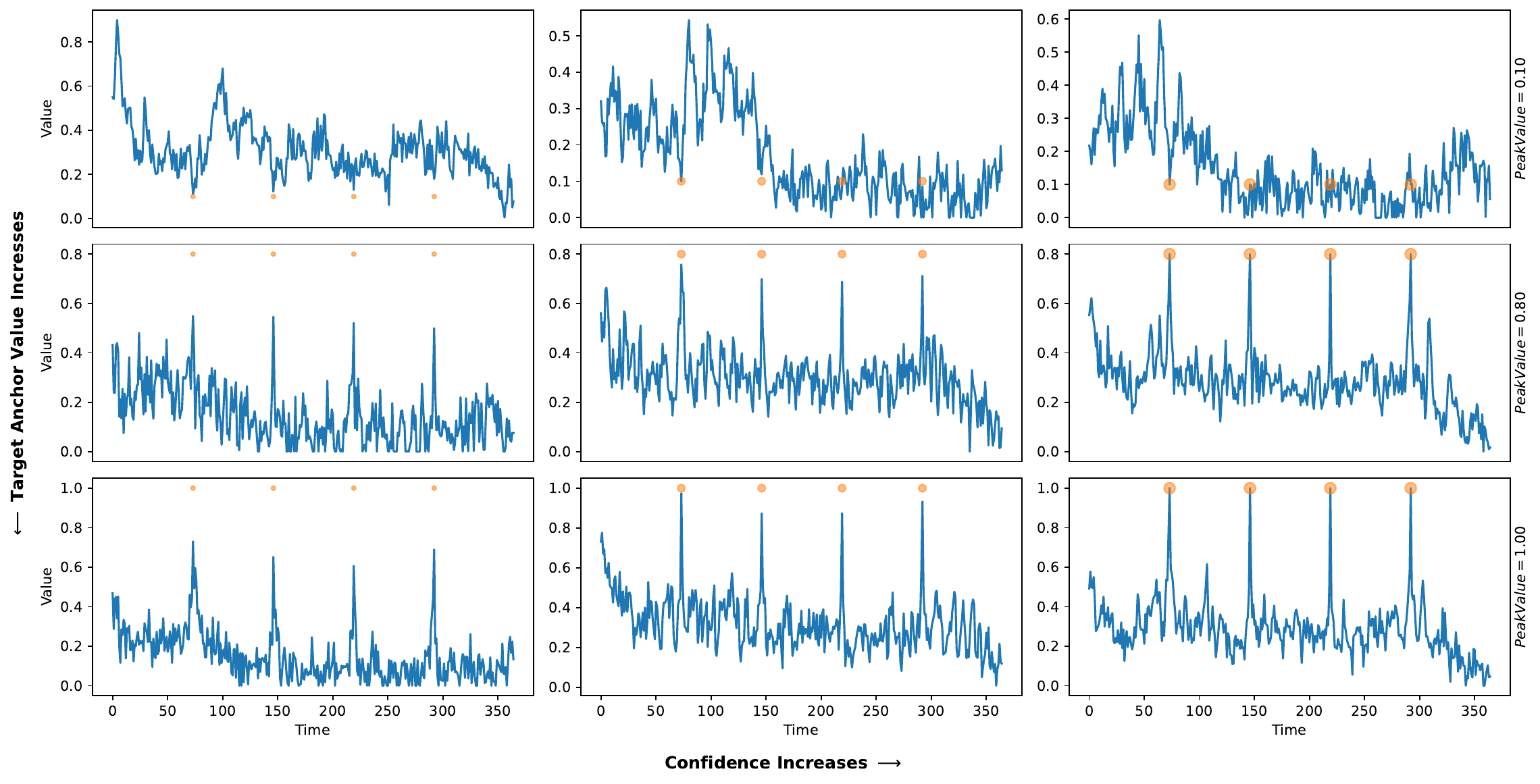}
    \vspace{-0.3cm}
    \caption{Demonstration of Point-Wise Control in Revenue datasets with multiple co'n'tr'l points and confidences}
\end{figure*}

\begin{figure*}[!htbp]
\centering
\includegraphics[width=0.75\linewidth]{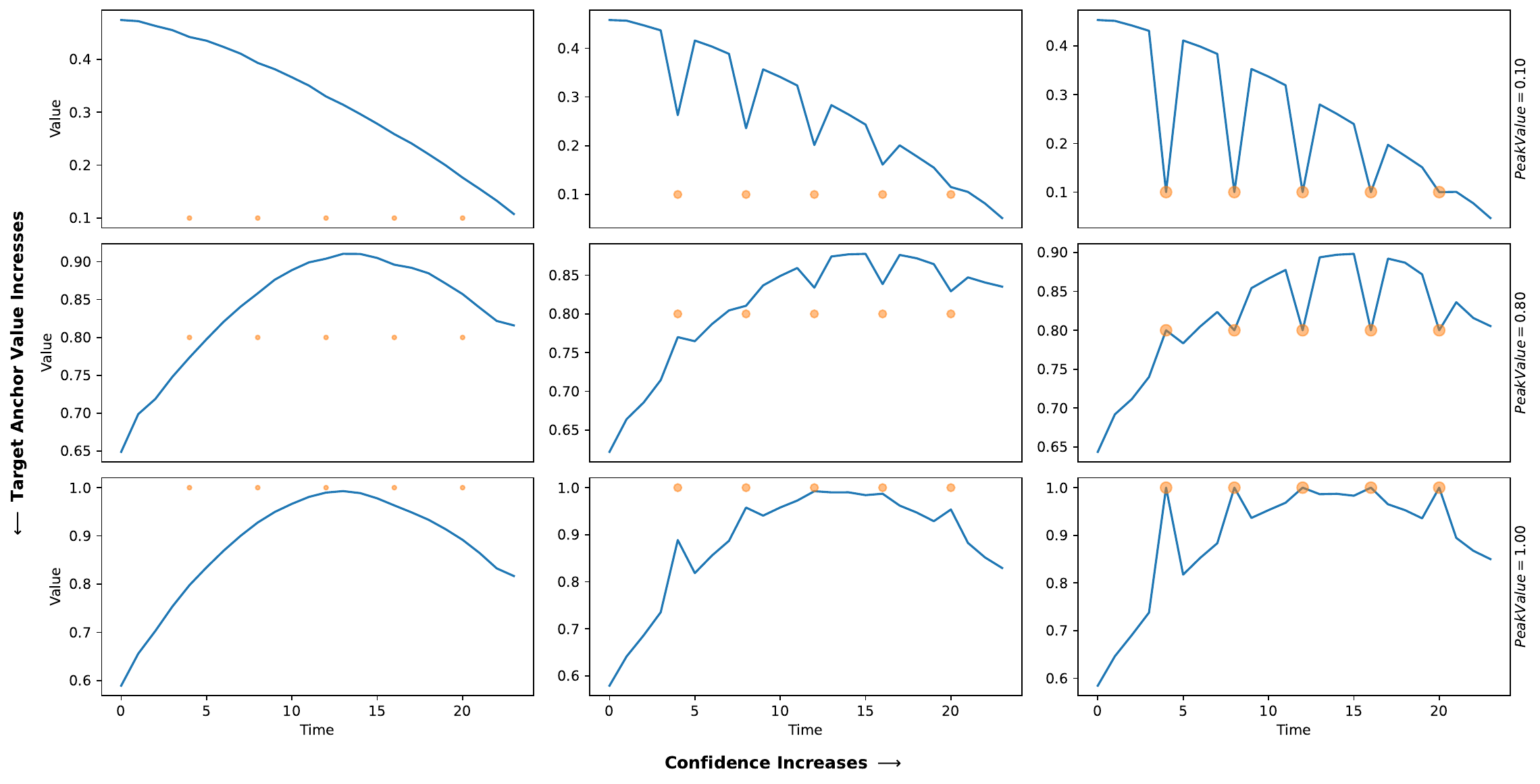}
\vspace{-0.3cm}
\caption{Demonstration of Point-Wise Control in Synthetic sine wave datasets with multiple control points and confidences}

\end{figure*}

\subsubsection{Float Mask Control with Extensions (CSDI)}
\begin{figure*}[!htbp]
\centering
\includegraphics[width=0.75\linewidth]{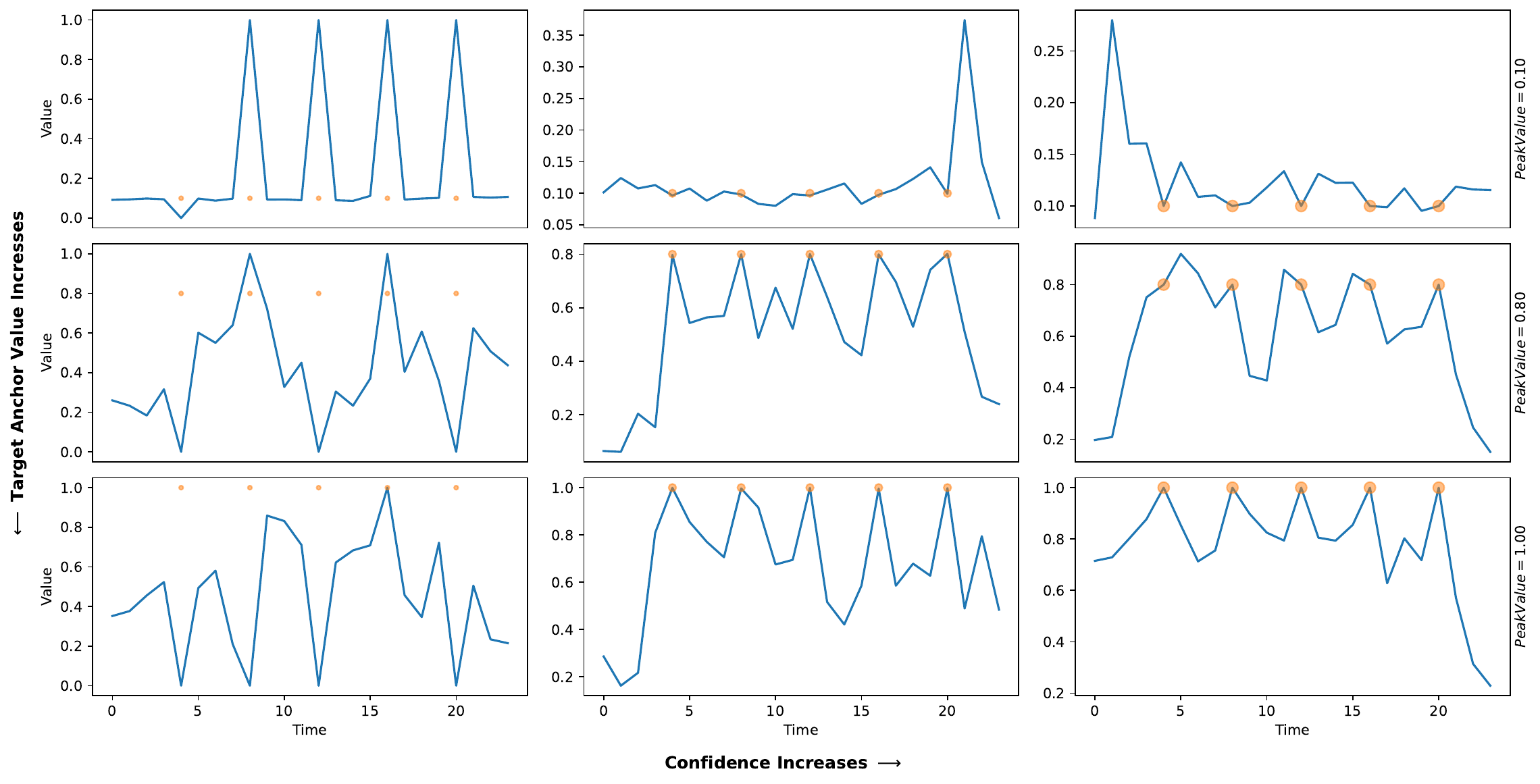}
\vspace{-0.3cm}
\caption{Demonstration of Point-Wise Control in ETTh datasets with multiple control points and confidences}
\end{figure*}
\begin{figure*}[!htbp]
\centering
\includegraphics[width=0.75\linewidth]{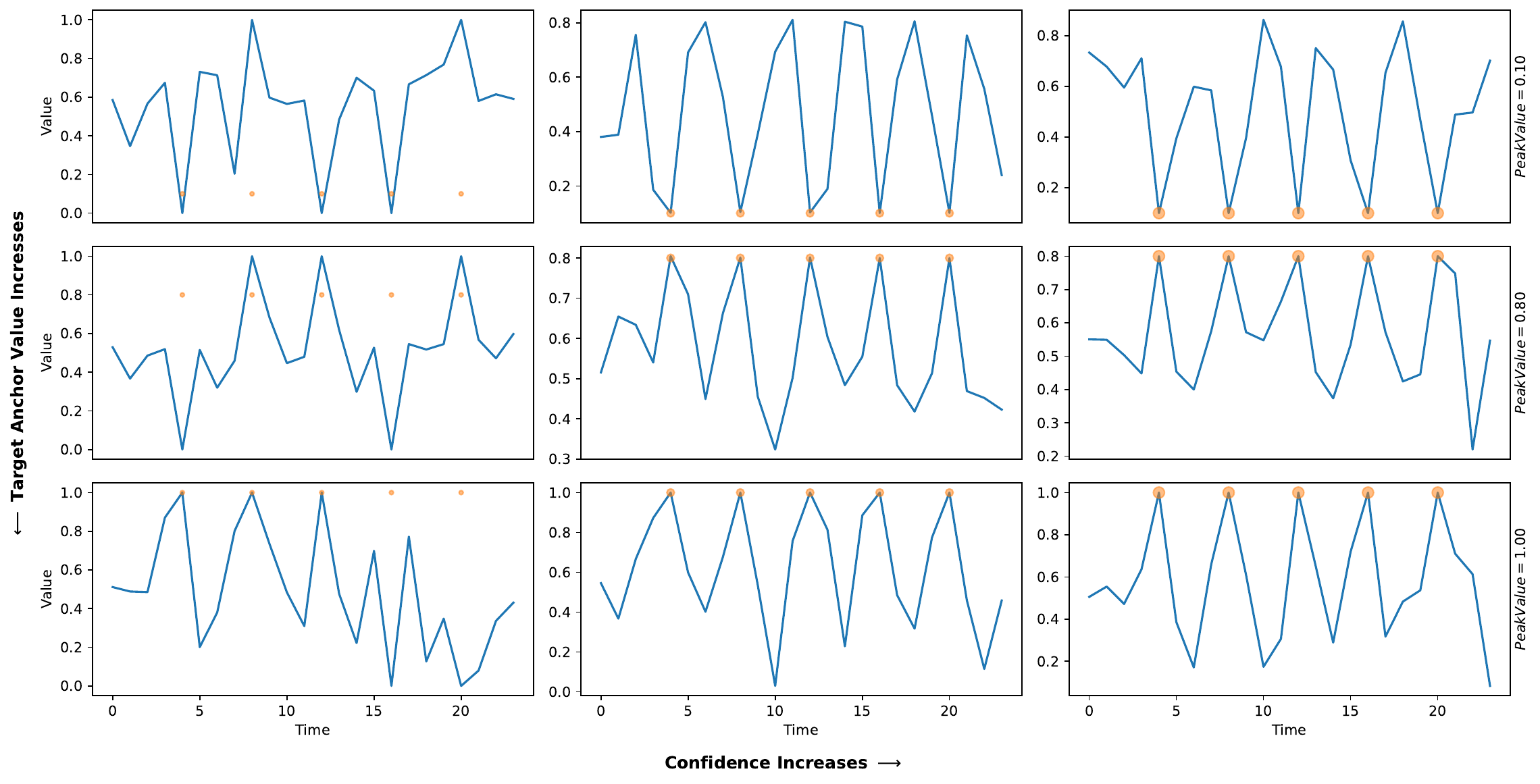}
    \vspace{-0.3cm}
    \caption{Demonstration of Point-Wise Control in fMRI datasets with multiple control points and confidences.}
\end{figure*}

\begin{figure*}[!htbp]
\centering
\includegraphics[width=0.75\linewidth]{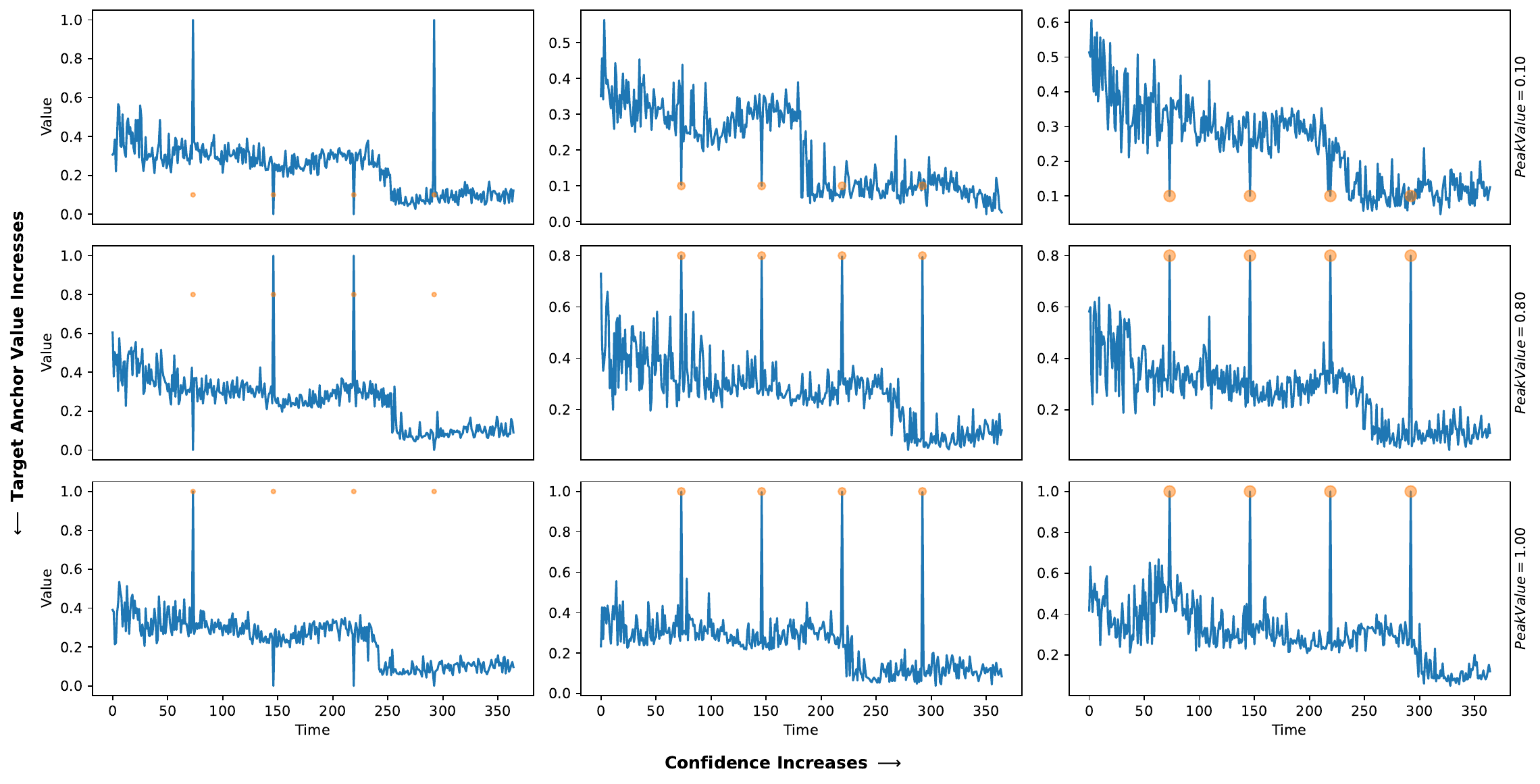}
    \vspace{-0.3cm}
    \caption{Demonstration of Point-Wise Control in Revenue datasets with multiple control points and confidences}
\end{figure*}

\begin{figure*}[!htbp]
\centering
\includegraphics[width=0.75\linewidth]{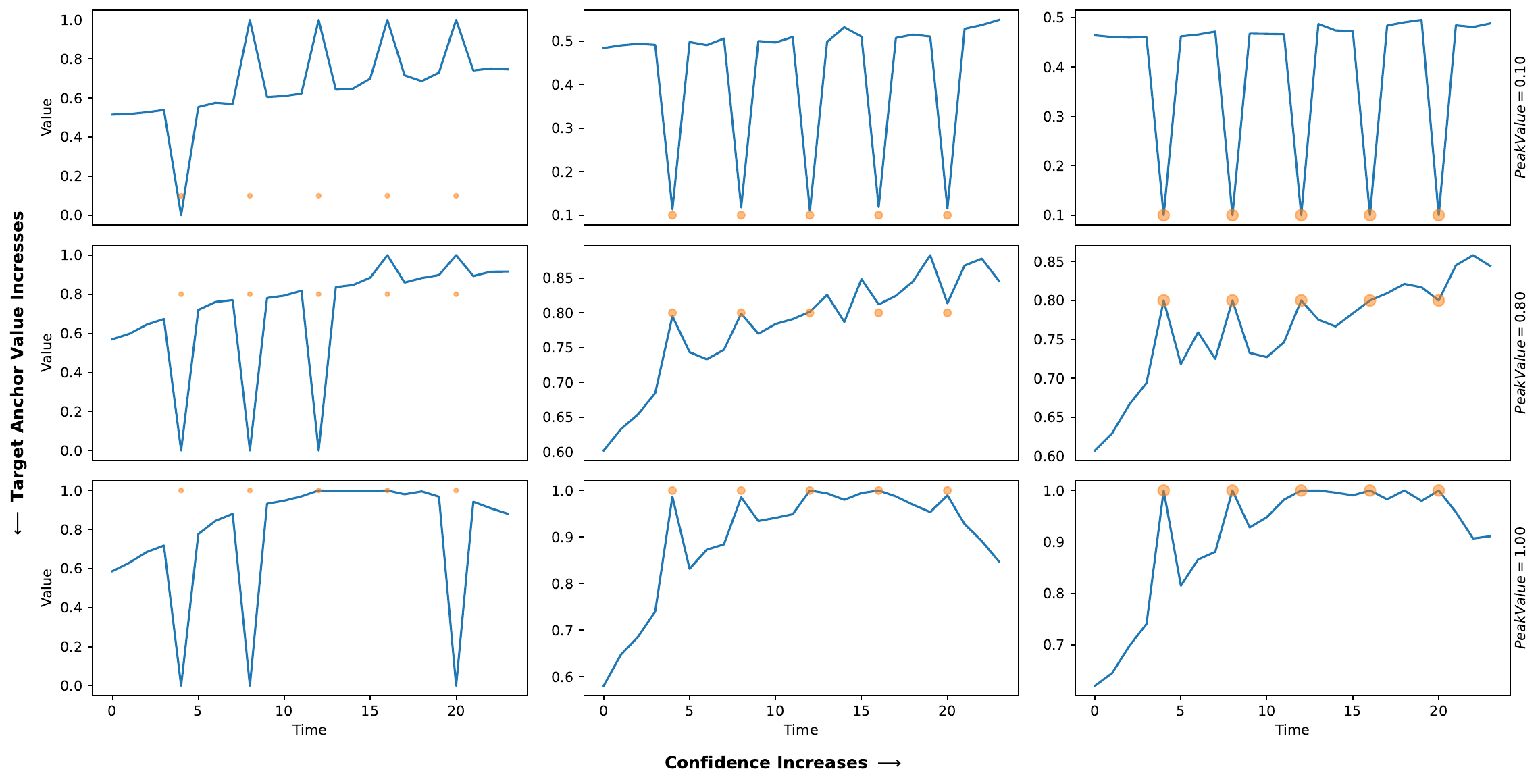}
\vspace{-0.3cm}
\caption{Demonstration of Point-Wise Control in Synthetic sine wave datasets with multiple control points and confidences}

\end{figure*}

\newpage
\subsection{KED of Point-Wise Control}
\label{app:ked-anchor}
\begin{figure*}[!htbp]
\centering
    \label{fig:etth_kde_anchor}
    \includegraphics[width=0.9\linewidth]{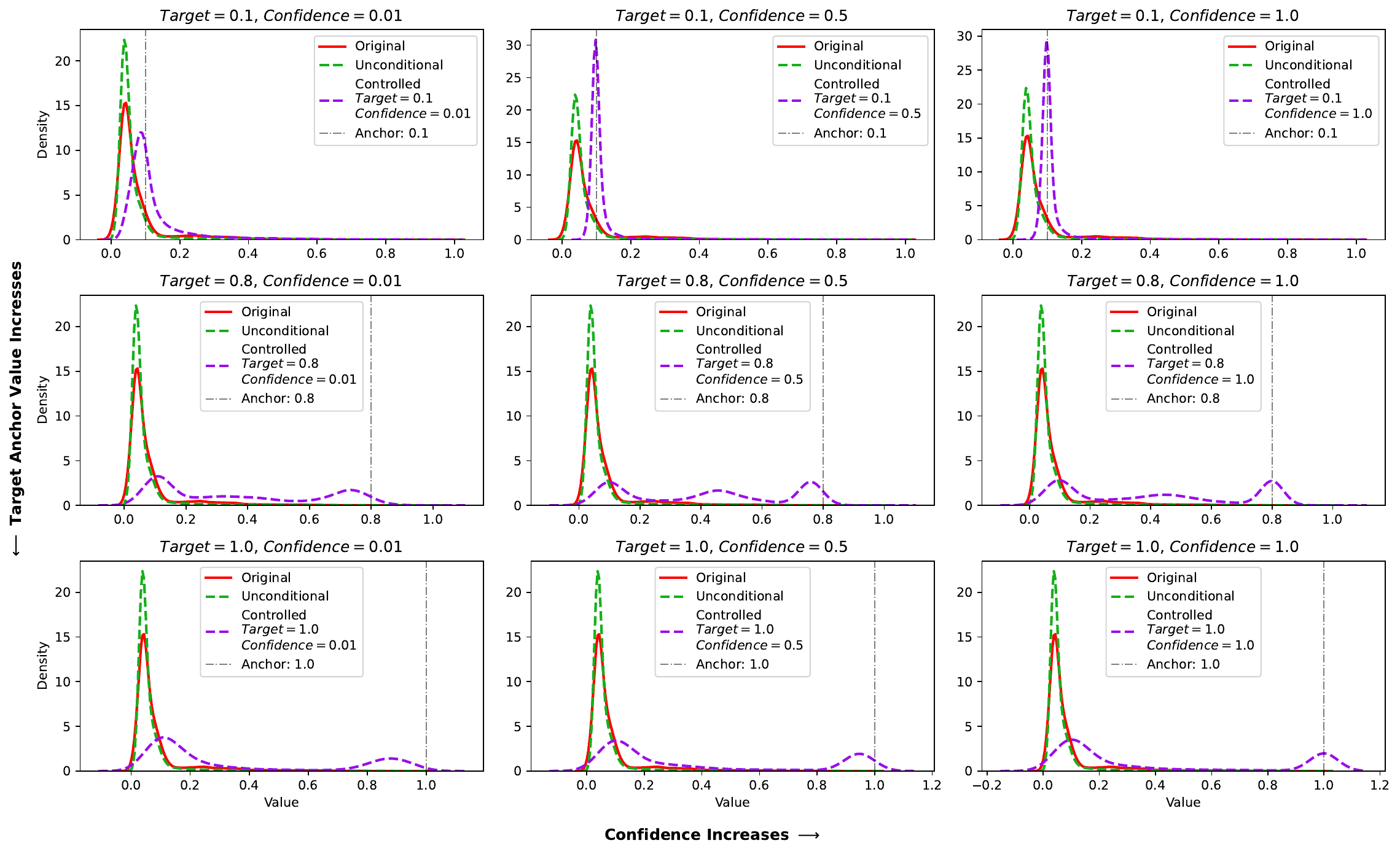}
    \caption{KDE analysis of ETTh dataset generation.}
\end{figure*}

The following plots of Kernel Density Estimation (KDE) clearly demonstrate how the distribution peaks (purple dash line) shift towards the anchor points as confidence increases.
For instance, in the ETTh dataset, the most pronounced shift occurs at anchor points under the highest confidence level. The model generates sequences that accurately respect anchor points while preserving the dataset's inherent distributional characteristics. In Figure \ref{fig:etth_kde_anchor}'s middle row, where the target value is 0.8, increasing confidence levels cause the peaks (y-value density) of controlled results (purple line) to intensify and converge toward the target value. This pattern is consistently observed across all datasets.

\subsubsection{Pure Float Mask Control (Diffusion-TS)}
\begin{figure*}[!htbp]
\centering
\includegraphics[width=0.9\linewidth]{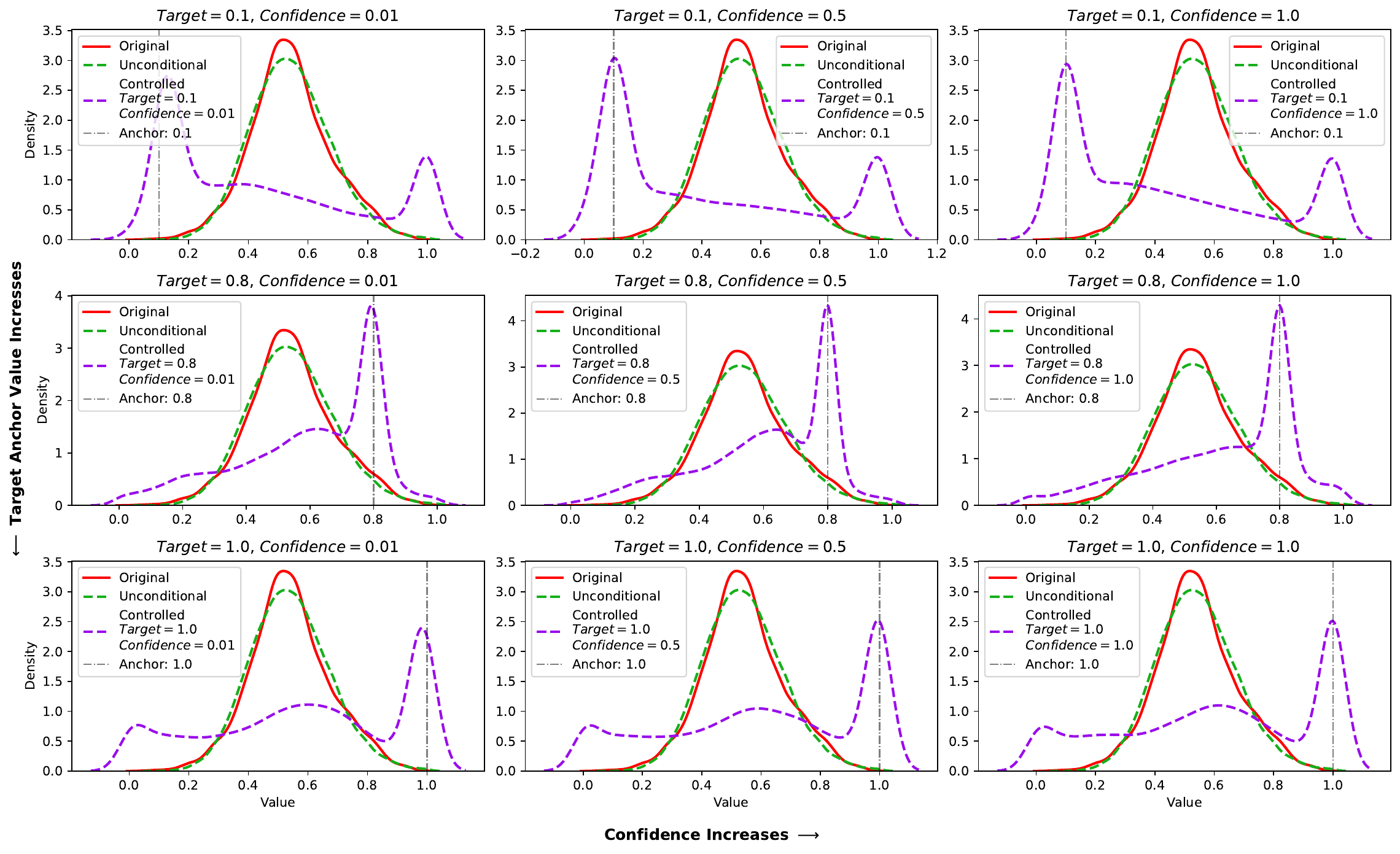}
    \caption{KDE analysis of fMRI dataset generation.}
\end{figure*}
\begin{figure*}[!htbp]
\centering
\includegraphics[width=0.9\linewidth]{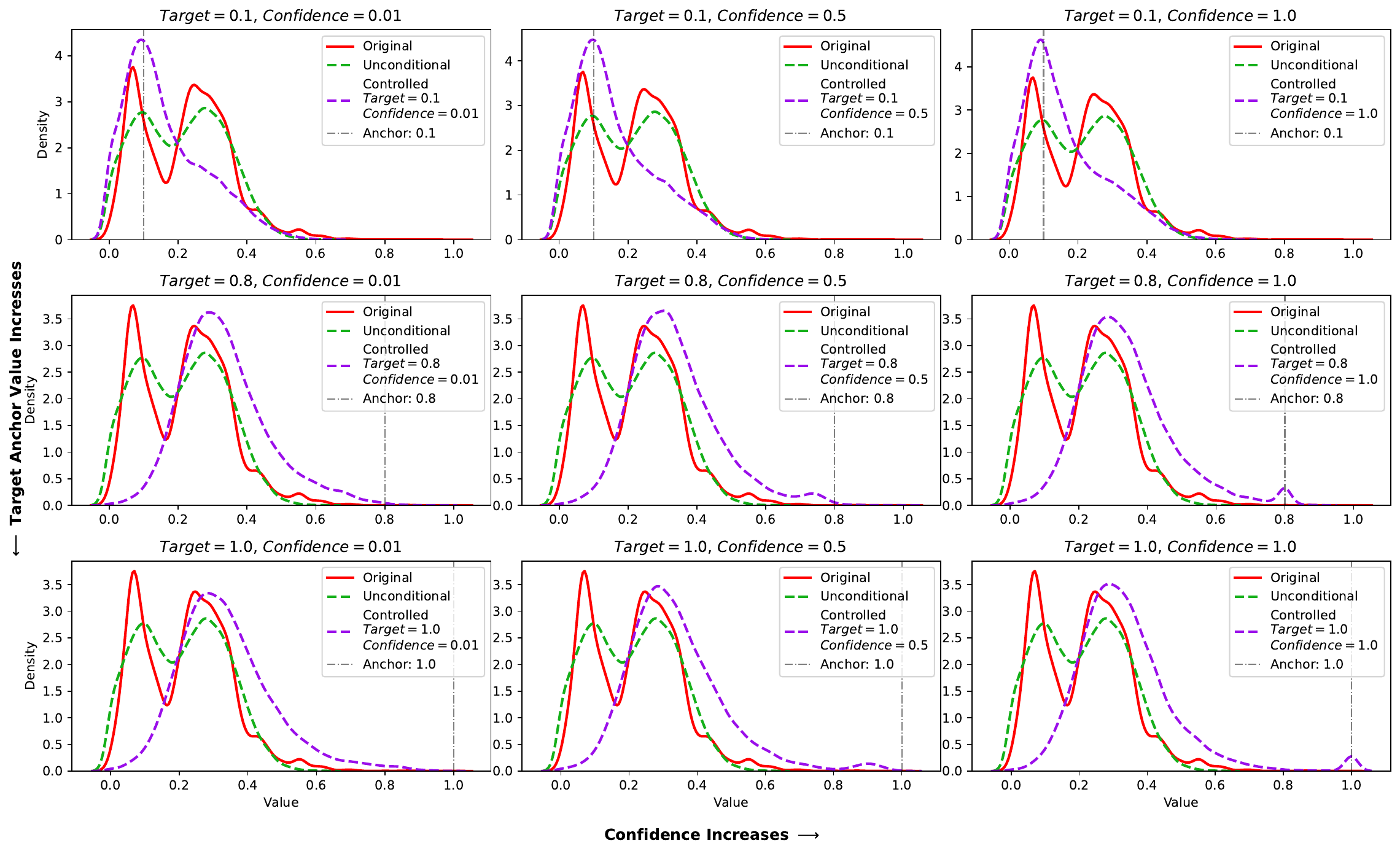}
    \caption{KDE analysis of Revenue dataset generation.}
\end{figure*}
\begin{figure*}[!htbp]
\centering
    \includegraphics[width=0.9\linewidth]{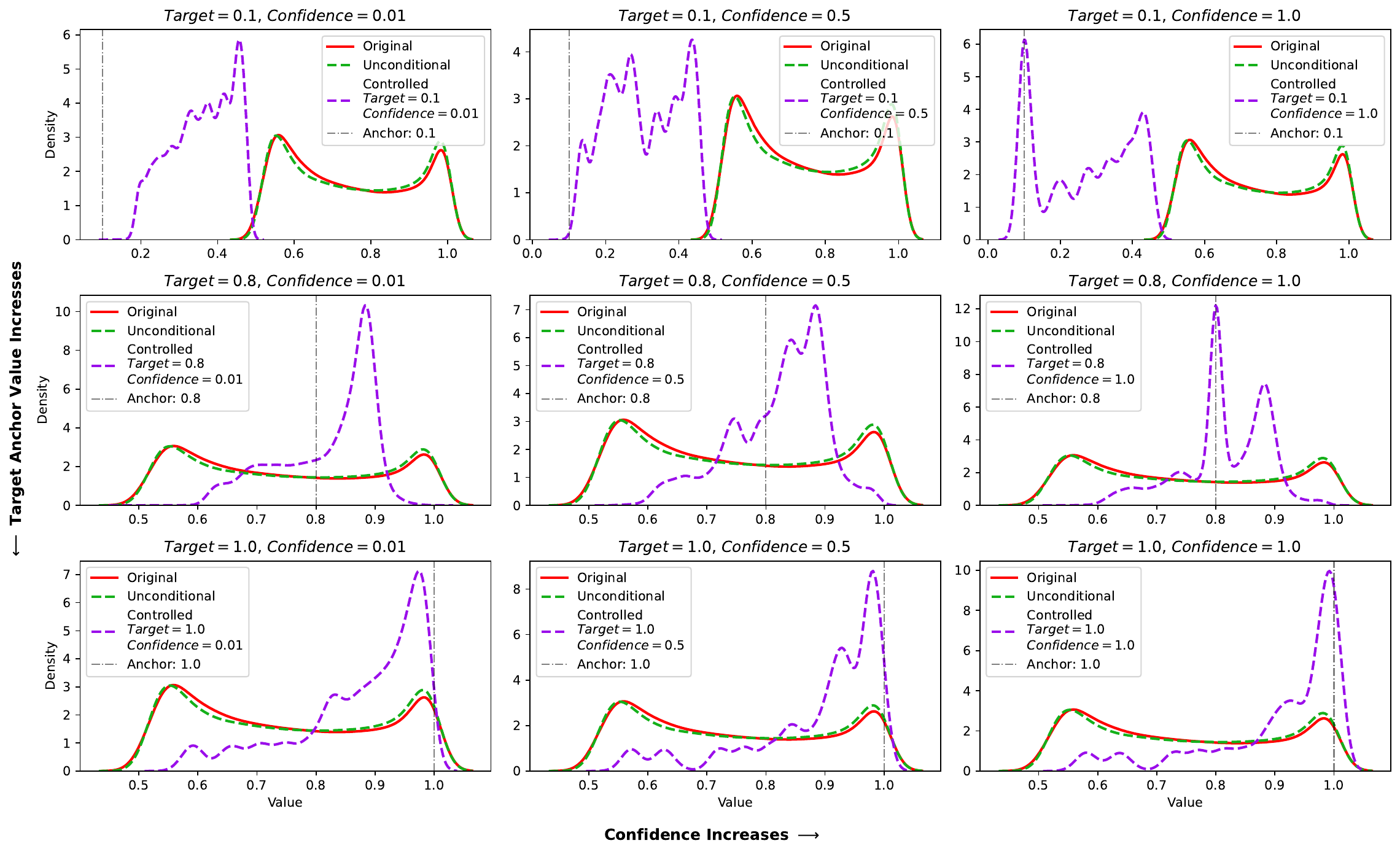}
    \caption{KDE analysis of synthetic sine wave dataset generation.}
\end{figure*}

\newpage
\newpage
\subsubsection{Float Mask Control with Extensions (Diffusion-TS)}
\begin{figure*}[!htbp]
\centering
    \includegraphics[width=0.9\linewidth]{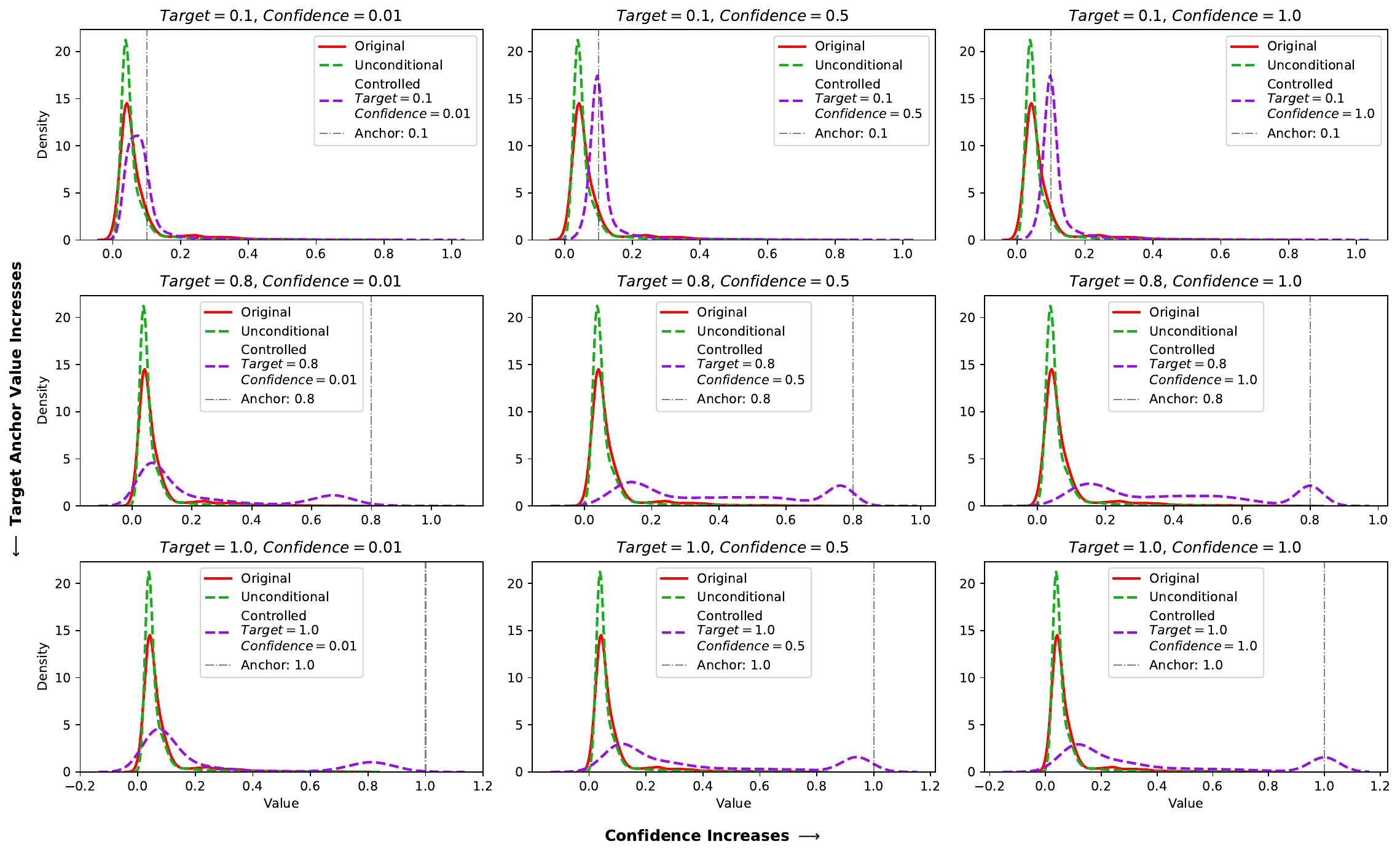}
    \caption{KDE analysis of ETTh dataset generation.}
\end{figure*}
\begin{figure*}[!htbp]
\centering
\includegraphics[width=0.9\linewidth]{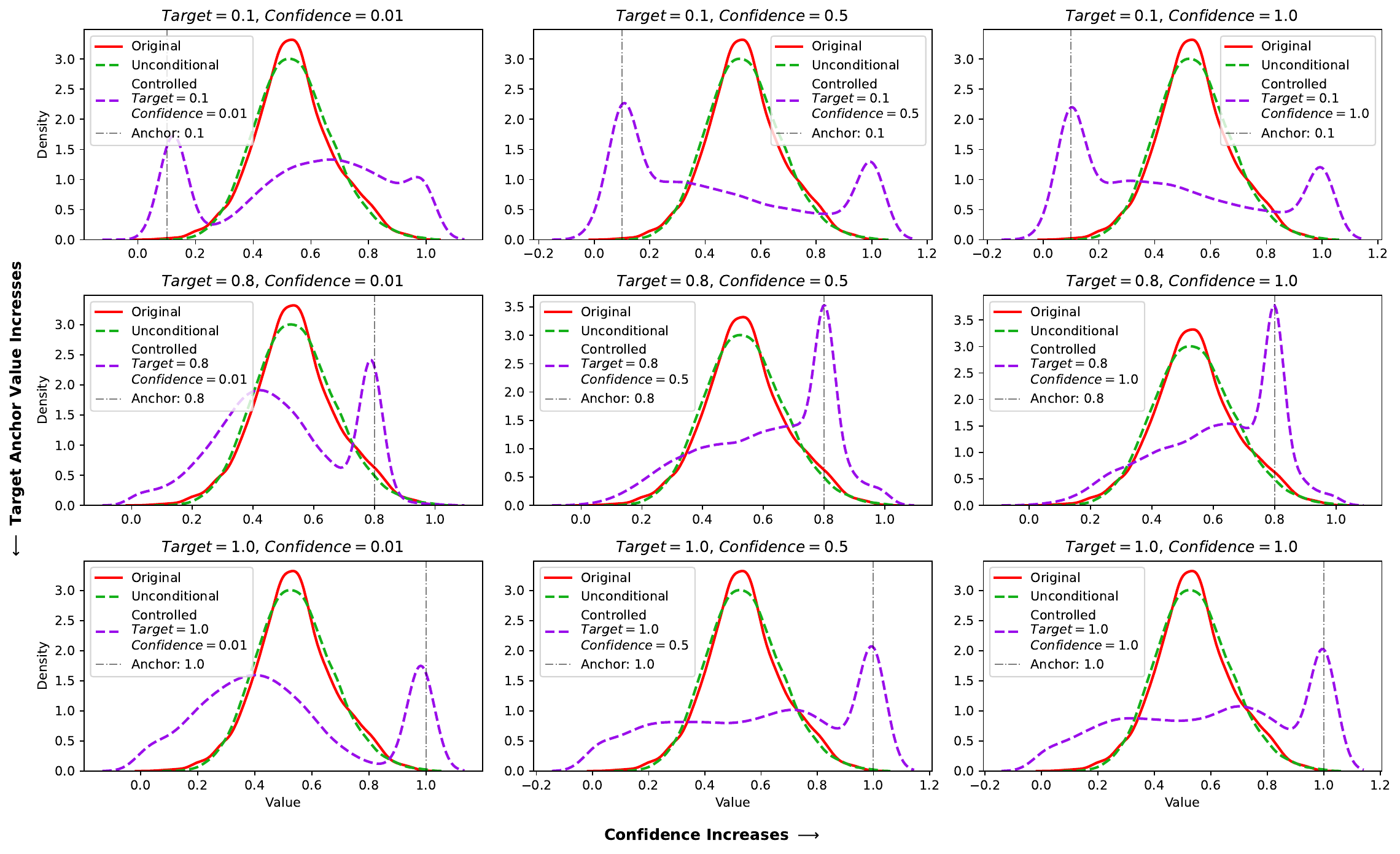}
    \caption{KDE analysis of fMRI dataset generation.}
\end{figure*}
\begin{figure*}[!htbp]
\centering
\includegraphics[width=0.9\linewidth]{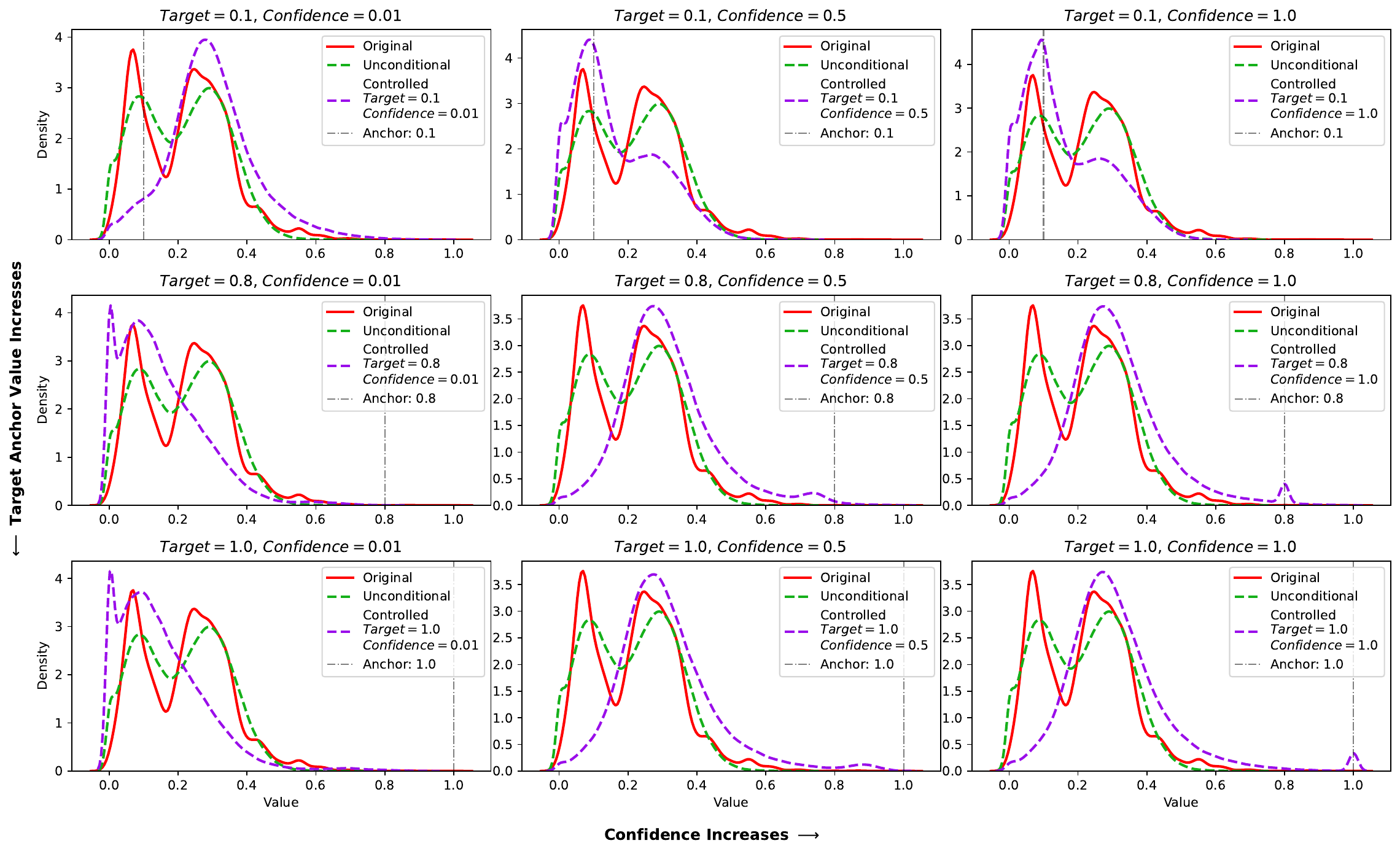}
    \caption{KDE analysis of Revenue dataset generation.}
\end{figure*}
\begin{figure*}[!htbp]
\centering
    \includegraphics[width=0.9\linewidth]{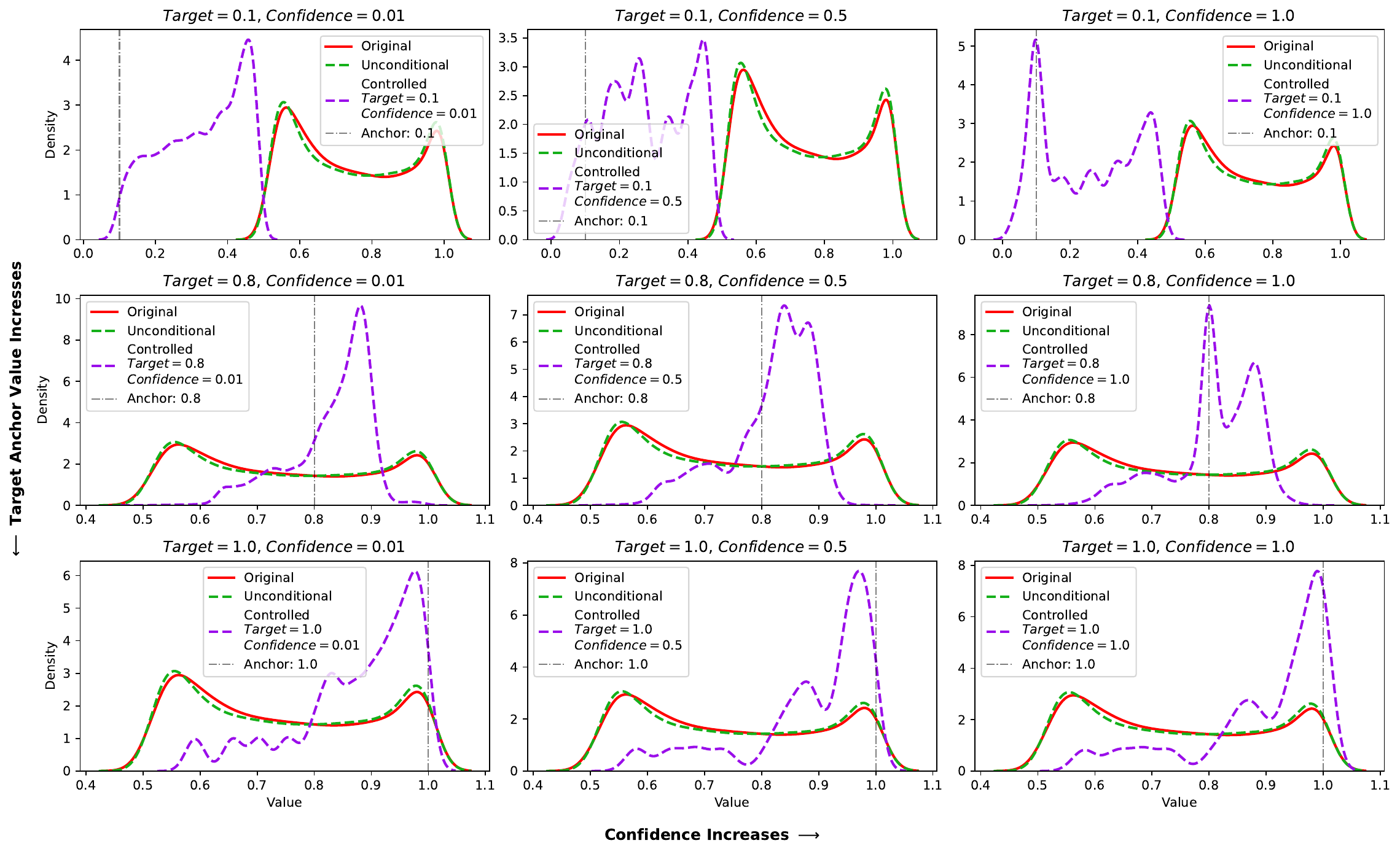}
    \caption{KDE analysis of synthetic sine wave dataset generation.}
\end{figure*}
    
\newpage
\subsubsection{Float Mask Control with Extensions (CSDI)}
\begin{figure*}[!htbp]
\centering
    \includegraphics[width=0.9\linewidth]{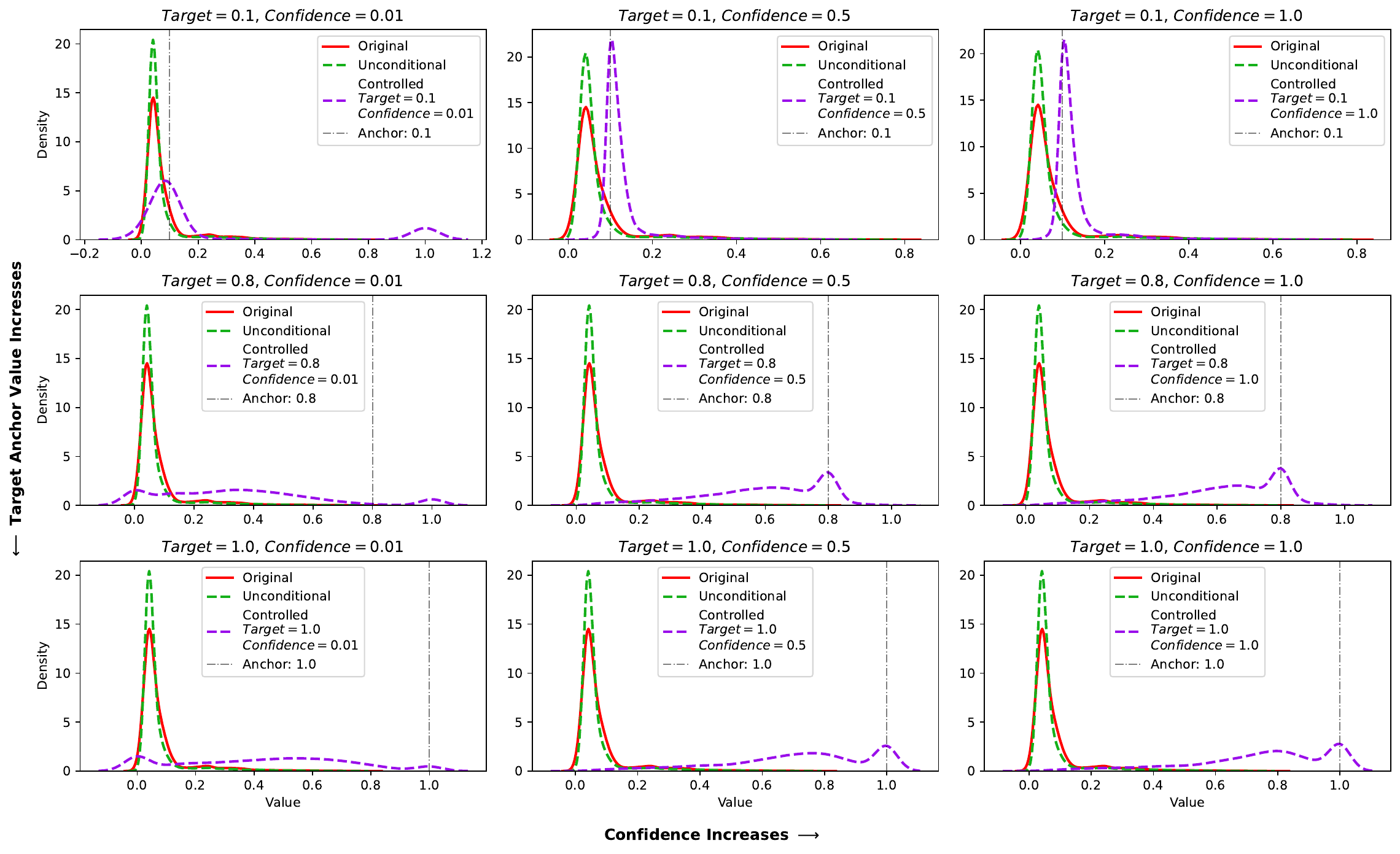}
    \caption{KDE analysis of ETTh dataset generation.}
\end{figure*}
\begin{figure*}[!htbp]
\centering
\includegraphics[width=0.9\linewidth]{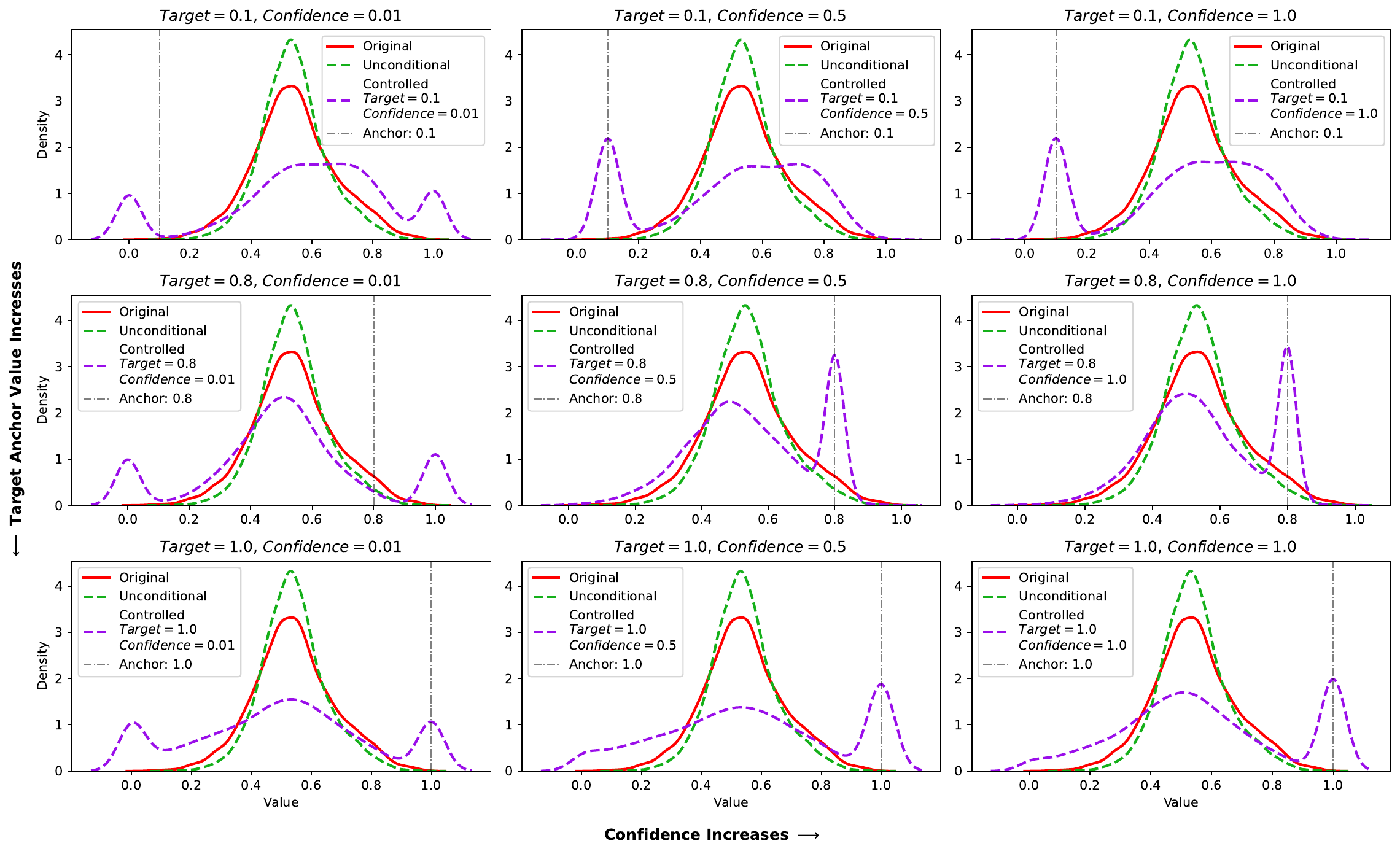}
    \caption{KDE analysis of fMRI dataset generation.}
\end{figure*}
\begin{figure*}[!htbp]
\centering
\includegraphics[width=0.9\linewidth]{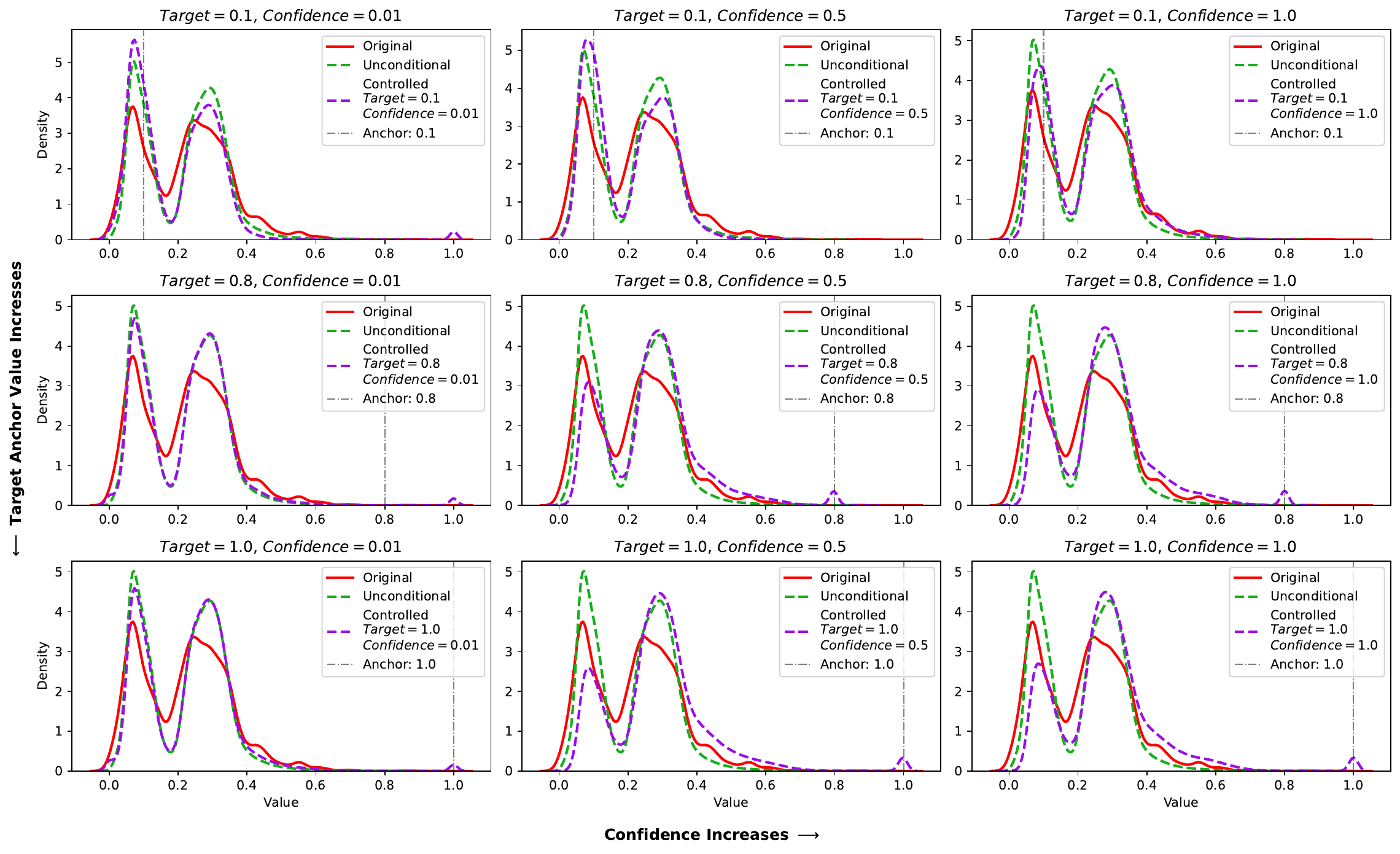}
    \caption{KDE analysis of Revenue dataset generation.}
\end{figure*}
\begin{figure*}[!htbp]
\centering
    \includegraphics[width=0.9\linewidth]{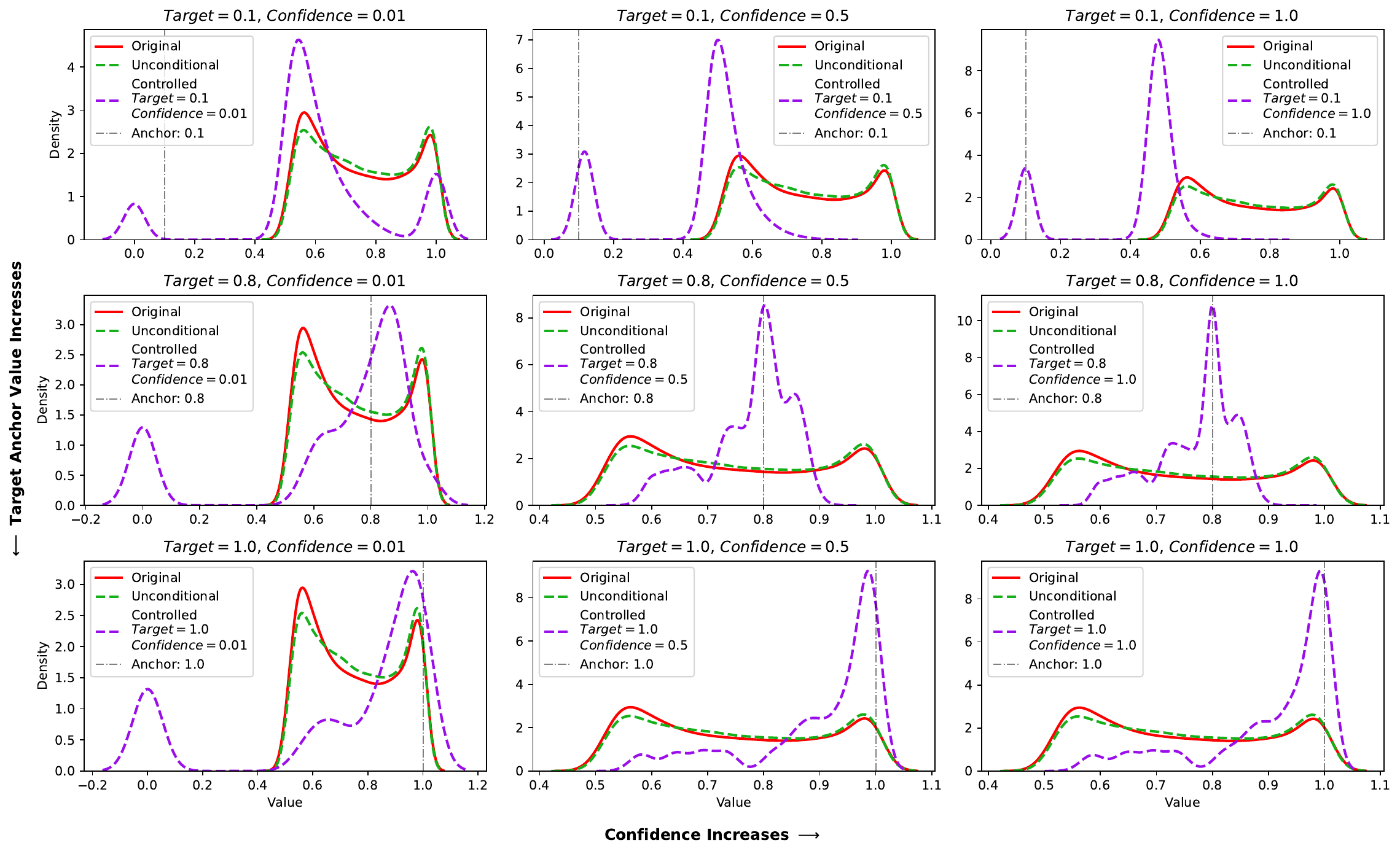}
    \caption{KDE analysis of synthetic sine wave dataset generation.}
\end{figure*}

\newpage
\subsection{Supplement Metrics}
\label{app:metrics}
The Discriminative, Predictive, Context-FID, and Correlational scores help quantify distribution shifts under point-wise control. Table \ref{tab:metrics-anchor-control} demonstrates that all metrics increase significantly after applying control, indicating that the control signals effectively influence the generated time series. But our method makes them more distinguishable from the original distribution, which is the improved direction for following research.

\begin{table}[H]
\label{tab:metrics-anchor-control}
\centering
\caption{Supplemental metrics for point-wise control performance across different datasets and target values. The results show discriminative,
predictive, context-FID, and correlational scores for each dataset and control configuration. Lower scores indicate better performance.}
\resizebox{0.8\columnwidth}{!}{%
\begin{tabular}{c|c|cccc}
\toprule
\multirow{2}{*}{Metrics} & \multirow{2}{*}{Control Signal} & \multicolumn{4}{c}{Dataset} \\
 &  & ETTh & Revenue & fMRI & Sine \\ \midrule
\multirow{4}{*}{\begin{tabular}[c]{@{}c@{}}Discriminative\\ Score\\ (Lower is\\ Better)\end{tabular}} & Unconditional & 0.103$\pm$0.042& 0.082$\pm$0.093& 0.141$\pm$0.037& 0.031$\pm$0.023\\
 & Confidence=0.01 & 0.497$\pm$0.005& 0.382$\pm$0.267& 0.500$\pm$0.000& 0.457$\pm$0.027\\
 & Confidence=0.5 & 0.496$\pm$0.006& 0.282$\pm$0.323& 0.499$\pm$0.001& 0.494$\pm$0.004\\
 & Confidence=1.0 & 0.498$\pm$0.001& 0.009$\pm$0.025& 0.498$\pm$0.000& 0.374$\pm$0.264\\ \hline
\multirow{4}{*}{\begin{tabular}[c]{@{}c@{}}Predictive\\ Score\\ (Lower is\\ Better)\end{tabular}} & Unconditional & 0.256$\pm$0.002& 0.065$\pm$0.026& 0.103$\pm$0.002& 0.094$\pm$0.000\\
 & Confidence=0.01 & 0.302$\pm$0.008& 0.181$\pm$0.002& 0.134$\pm$0.010& 0.120$\pm$0.007\\
 & Confidence=0.5 & 0.305$\pm$0.014& 0.179$\pm$0.005& 0.141$\pm$0.010& 0.118$\pm$0.012\\
 & Confidence=1.0 & 0.335$\pm$0.019& 0.175$\pm$0.009& 0.139$\pm$0.009& 0.116$\pm$0.008\\ \hline
\multirow{4}{*}{\begin{tabular}[c]{@{}c@{}}Context-FID\\ Score\\ (Lower is\\ Better)\end{tabular}} & Unconditional & 0.108$\pm$0.007& 1.230$\pm$0.284& 0.260$\pm$0.024& 0.034$\pm$0.005\\
 & Confidence=0.01 & 7.797$\pm$0.644& 4.654$\pm$1.389& 15.671$\pm$2.920& 4.169$\pm$1.139\\
 & Confidence=0.5 & 6.973$\pm$1.514& 5.921$\pm$0.711& 15.463$\pm$1.888& 10.842$\pm$2.103\\
 & Confidence=1.0 & 7.856$\pm$1.326& 7.083$\pm$0.512& 13.854$\pm$0.830& 8.906$\pm$1.168\\ \hline
\multirow{4}{*}{\begin{tabular}[c]{@{}c@{}}Correlational\\ Score\\ (Lower is\\ Better)\end{tabular}} & Unconditional & 2.313$\pm$0.743& 0.038$\pm$0.013& 2.672$\pm$0.091& 0.066$\pm$0.009\\
 & Confidence=0.01 & 9.321$\pm$0.764& 0.122$\pm$0.007& 16.699$\pm$0.453& 0.255$\pm$0.013\\
 & Confidence=0.5 & 8.445$\pm$0.675& 0.119$\pm$0.006& 15.488$\pm$0.174& 0.468$\pm$0.016\\
 & Confidence=1.0 & 9.640$\pm$0.835& 0.121$\pm$0.011& 17.259$\pm$0.434& 0.345$\pm$0.034\\ 
 \bottomrule
\end{tabular}%
}

\end{table}

\newpage
\subsection{Point-Wise Control Analysis}
Here, we demonstrate again the complete aggregated Mean Absolute Deviation (MAD) of point-wise control across all datasets, providing solid evidence for the effectiveness of point-wise control in maintaining anchor points.
\label{app:anchor_analysis_full}
\subsubsection{Pure Float Mask Control}
\vspace{-1em}
\begin{figure}[!ht]
\centering
\includegraphics[width=1.0\textwidth]{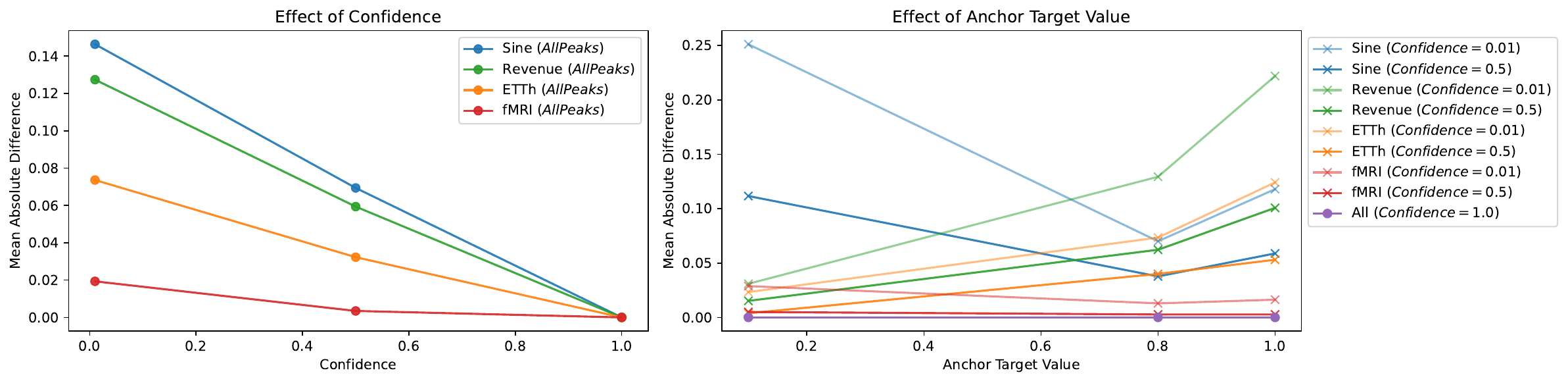}
\vspace{-1em}
\caption{Different combinations of confidence levels and target values across all datasets. (Diffusion-TS)}
\end{figure}
\subsubsection{Float Mask Control with Extensions}
\vspace{-1em}
\begin{figure}[!ht]
\centering
\includegraphics[width=1.0\textwidth]{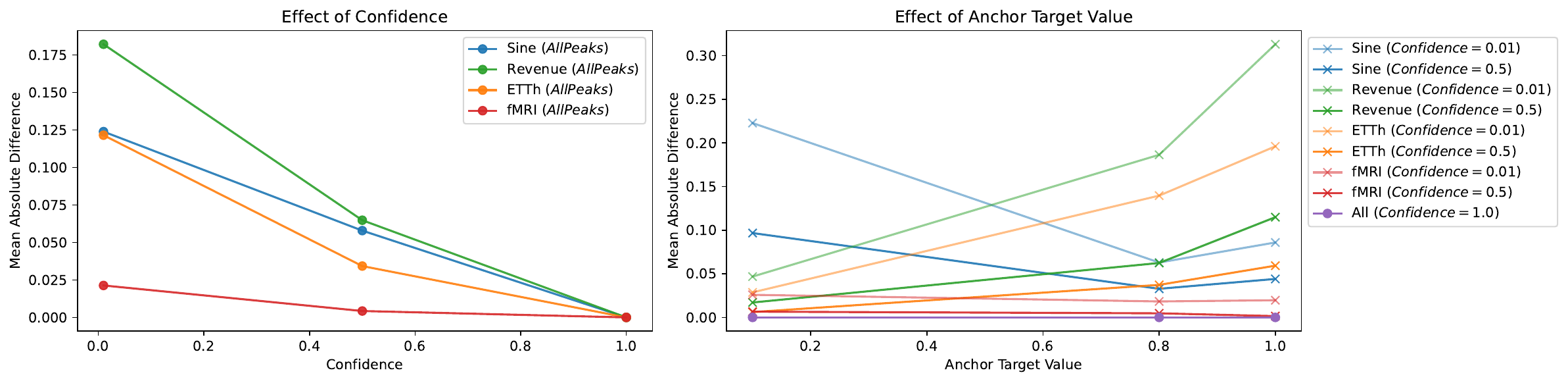}
\vspace{-1em}
\caption{Different combinations of confidence levels and target values cross all datasets. (Diffusion-TS)}
\end{figure}

\begin{figure}[!ht]
\centering
\includegraphics[width=1.0\textwidth]{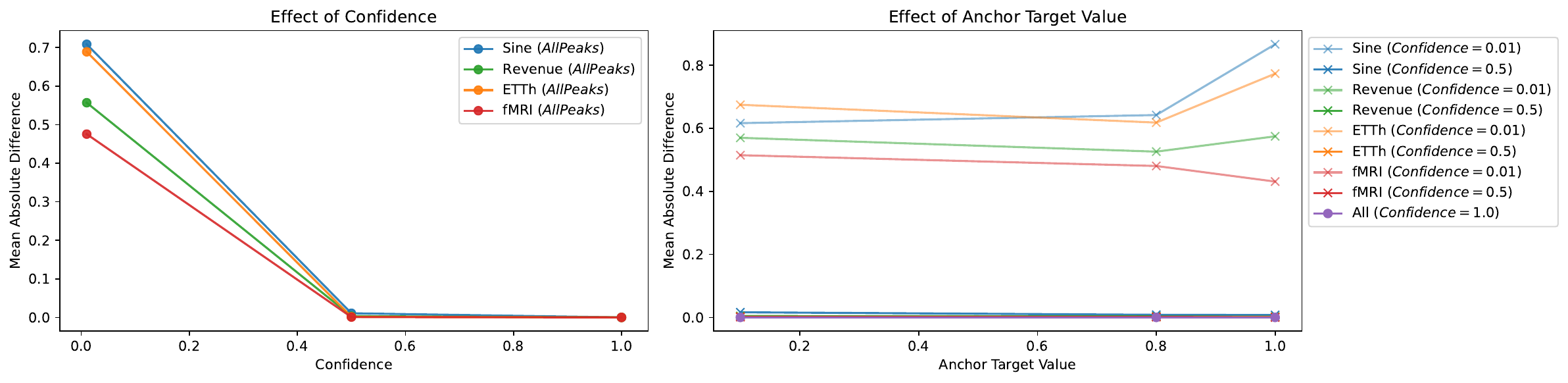}
\vspace{-1em}
\caption{Different combinations of confidence levels and target values across all datasets. (CDSI)}
\end{figure}
    
\section{Statistic Control}

This section provides supplementary materials for analyzing statistical control, focusing on both Sum Control and Segment-Wise Sum Control.

\subsection{Demonstration}

The following figures demonstrate the effectiveness of sum control across different datasets. As the target value increases (from left to right in each row), the model generates sequences that successfully adhere to the sum constraints while maintaining the dataset's inherent distributional properties. For segment-wise sum control, the results consistently show that the model tends to increase the overall sequence sum value, which aligns with the objective of preserving the original distribution learned from the dataset.

\label{app:sum_demo}

\subsubsection{Whole Time Series Summation Control (Diffusion-TS)}
\begin{figure*}[!htbp]
\centering
\includegraphics[width=\linewidth]{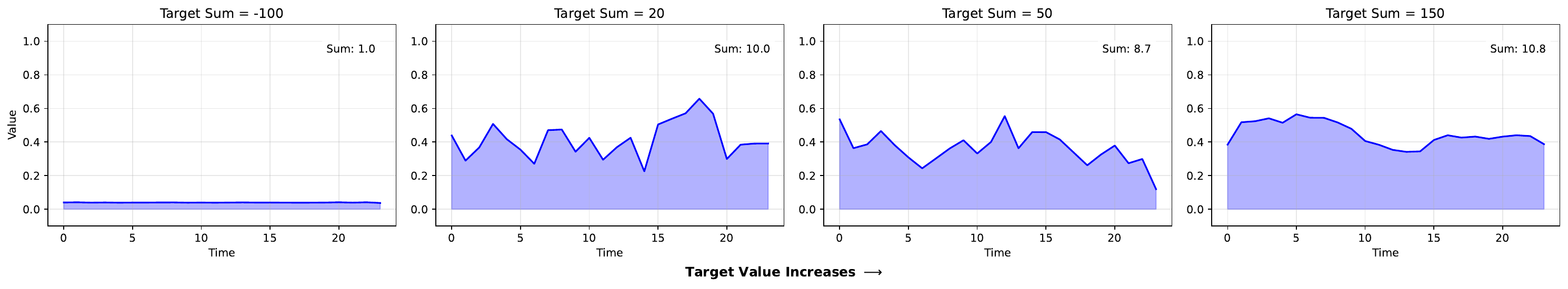}
\vspace{-0.3cm}
\caption{Demonstration of Sum Control on ETTh dataset.}
\end{figure*}
    \vspace{-1em}
\begin{figure*}[!htbp]
    \includegraphics[width=\linewidth]{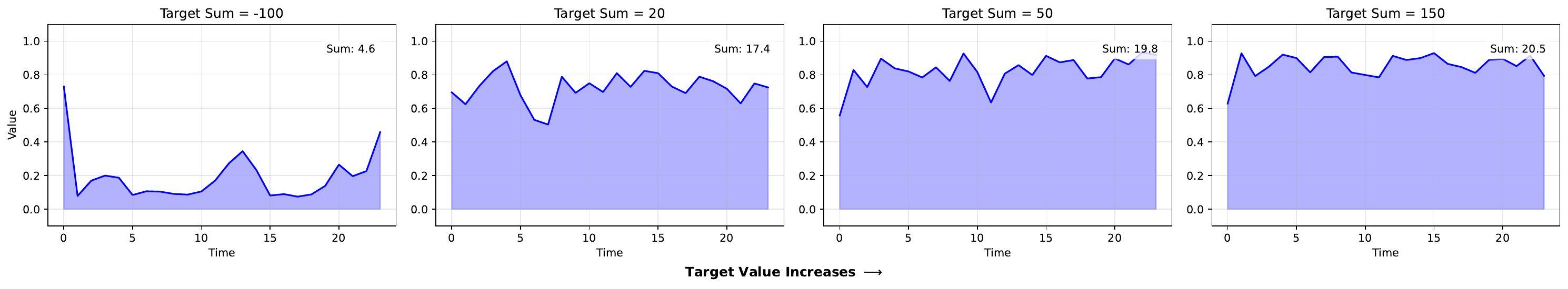}
    \vspace{-0.3cm}
    \caption{Demonstration of Sum Control on fMRI dataset.}
\end{figure*}
\vspace{-1em}

\begin{figure*}[!htbp]
    \includegraphics[width=\linewidth]{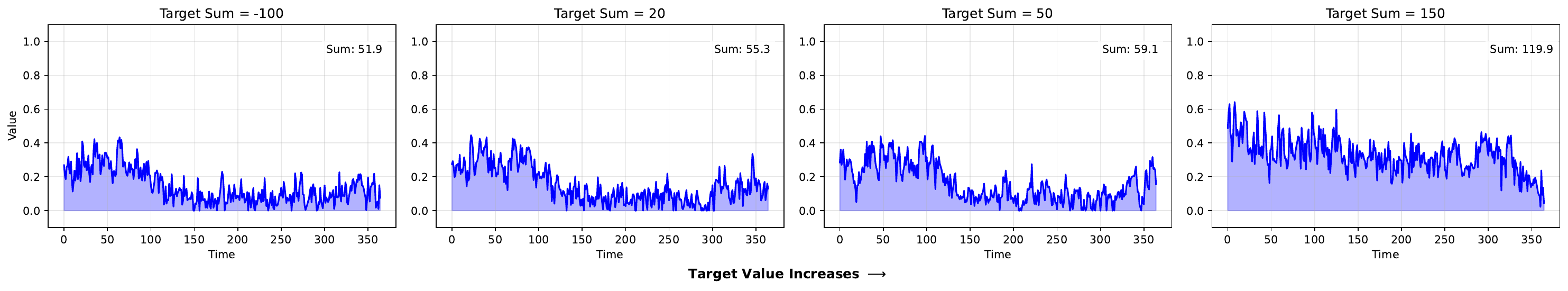}
    \vspace{-0.3cm}
    \caption{Demonstration of Sum Control on Revenue dataset.}
\end{figure*}
\vspace{-1em}

\begin{figure*}[!htbp]
\includegraphics[width=\linewidth]{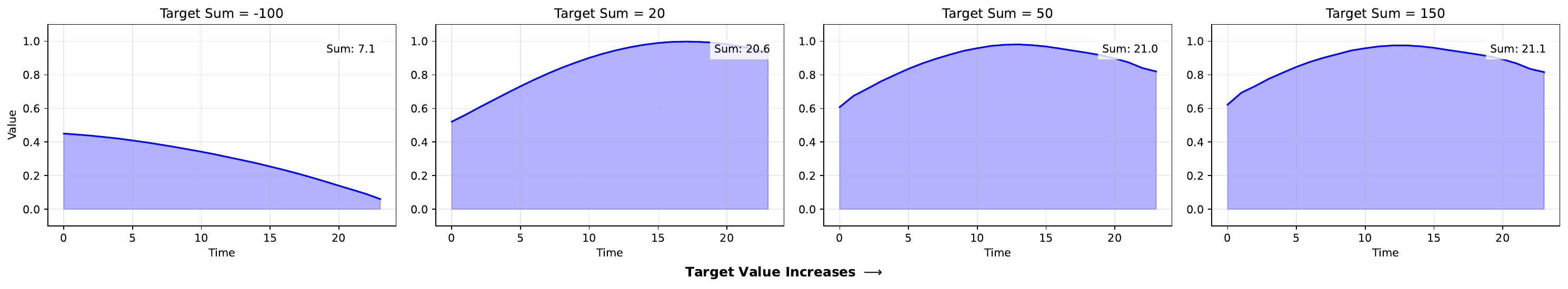}
\vspace{-0.3cm}
\caption{Demonstration of Sum Control on synthetic sine wave dataset.}
\end{figure*}

\newpage
\subsubsection{Whole Time Series Summation Control (CSDI)}
\begin{figure*}[!htbp]
\centering
\includegraphics[width=\linewidth]{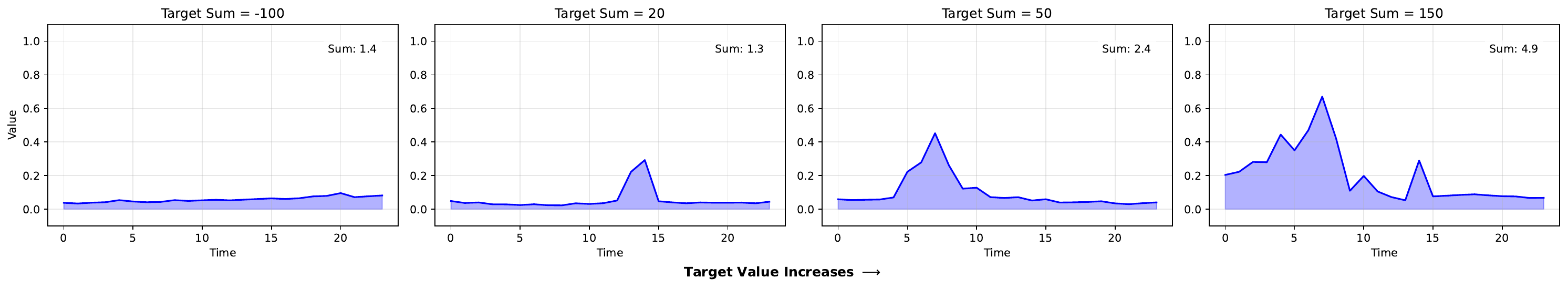}
\vspace{-0.3cm}
\caption{Demonstration of Sum Control on ETTh dataset.}
\end{figure*}
    \vspace{-1em}
\begin{figure*}[!htbp]
    \includegraphics[width=\linewidth]{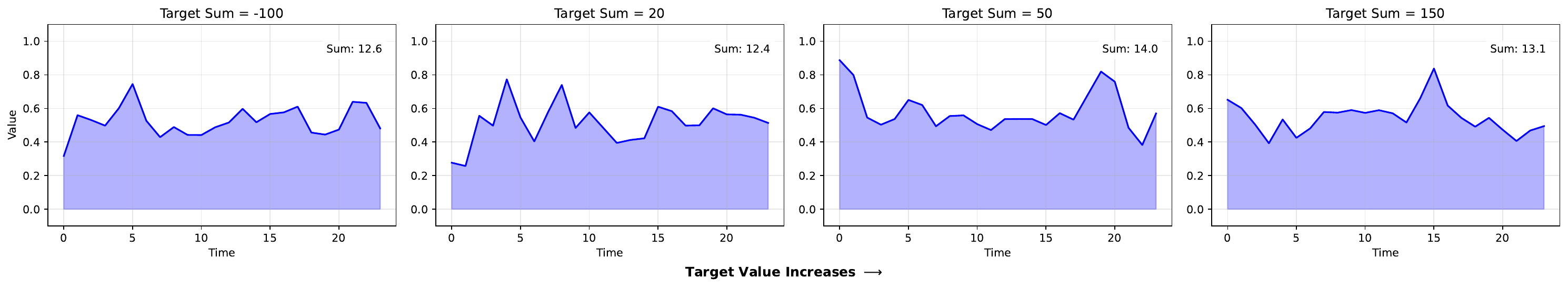}
    \vspace{-0.3cm}
    \caption{Demonstration of Sum Control on fMRI dataset.}
\end{figure*}
\vspace{-1em}

\begin{figure*}[!htbp]
    \includegraphics[width=\linewidth]{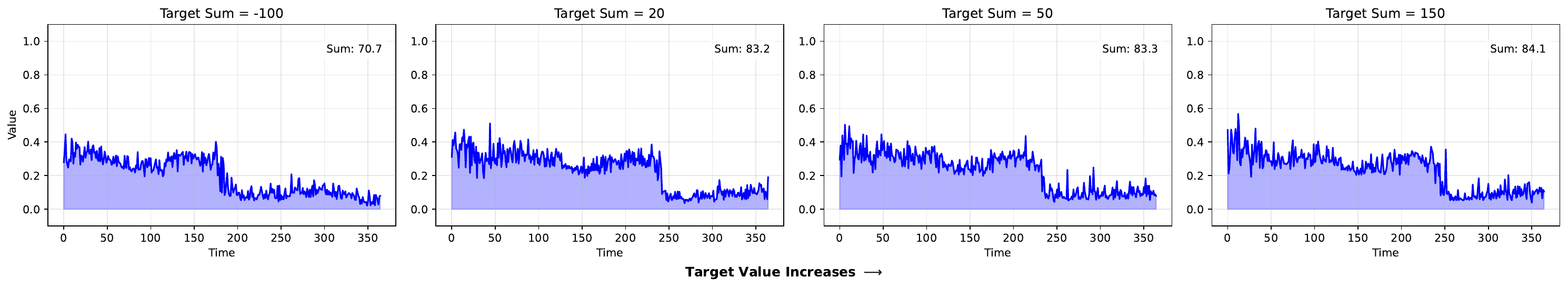}
    \vspace{-0.3cm}
    \caption{Demonstration of Sum Control on Revenue dataset.}
\end{figure*}
\vspace{-1em}

\begin{figure*}[!htbp]
\includegraphics[width=\linewidth]{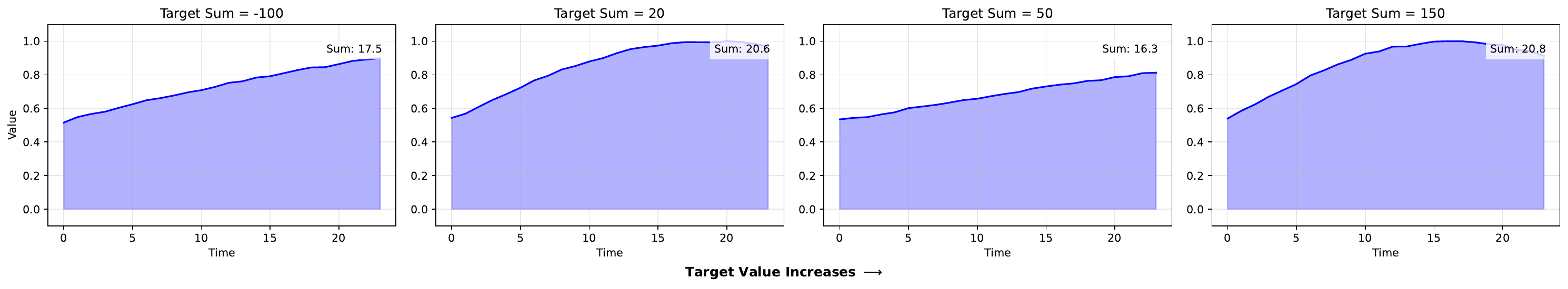}
\vspace{-0.3cm}
\caption{Demonstration of Sum Control on synthetic sine wave dataset.}
\end{figure*}

\newpage
\subsubsection{Segment Wise Sum Control (Diffusion-TS)}
\begin{figure*}[!htbp]
\centering
\includegraphics[width=\linewidth]{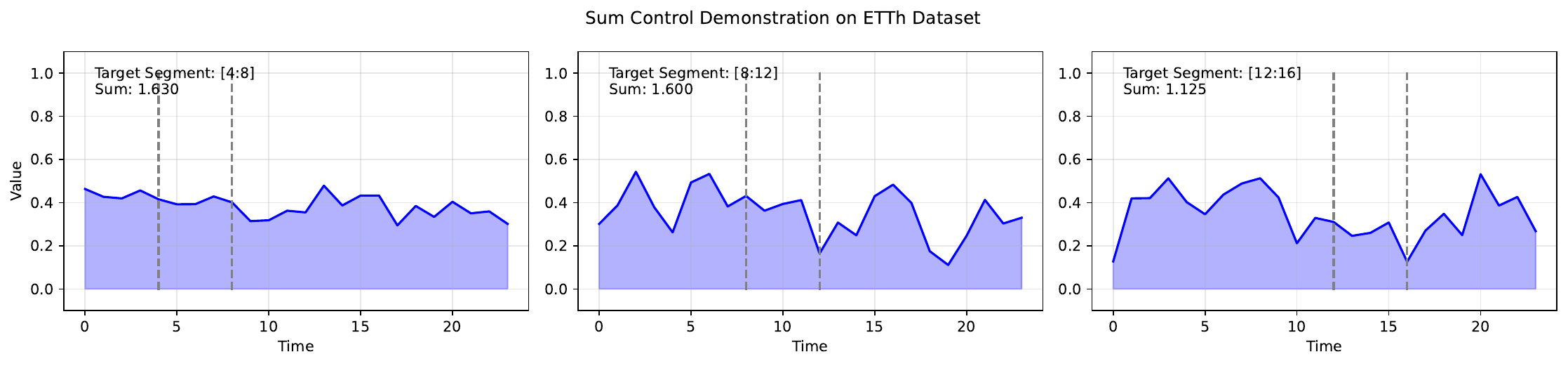}
\vspace{-0.3cm}
\caption{Demonstration of Segment-Wise Sum Control on ETTh dataset.}
\end{figure*}
    \vspace{-1em}
\begin{figure*}[!htbp]
    \includegraphics[width=\linewidth]{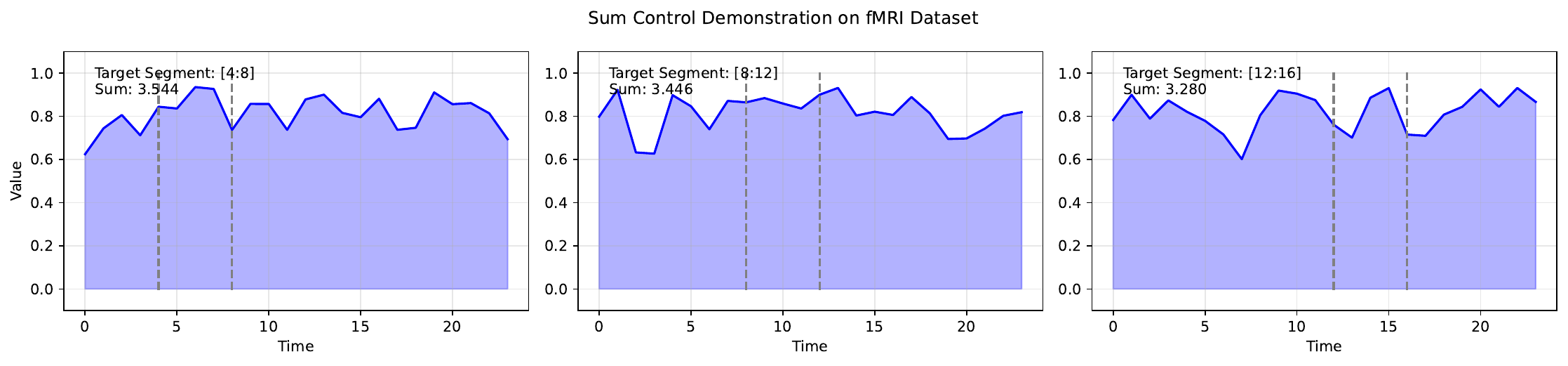}
    \vspace{-0.3cm}
    \caption{Demonstration of Segment-Wise Sum Control on fMRI dataset.}
\end{figure*}
\vspace{-1em}

\begin{figure*}[!htbp]
    \includegraphics[width=\linewidth]{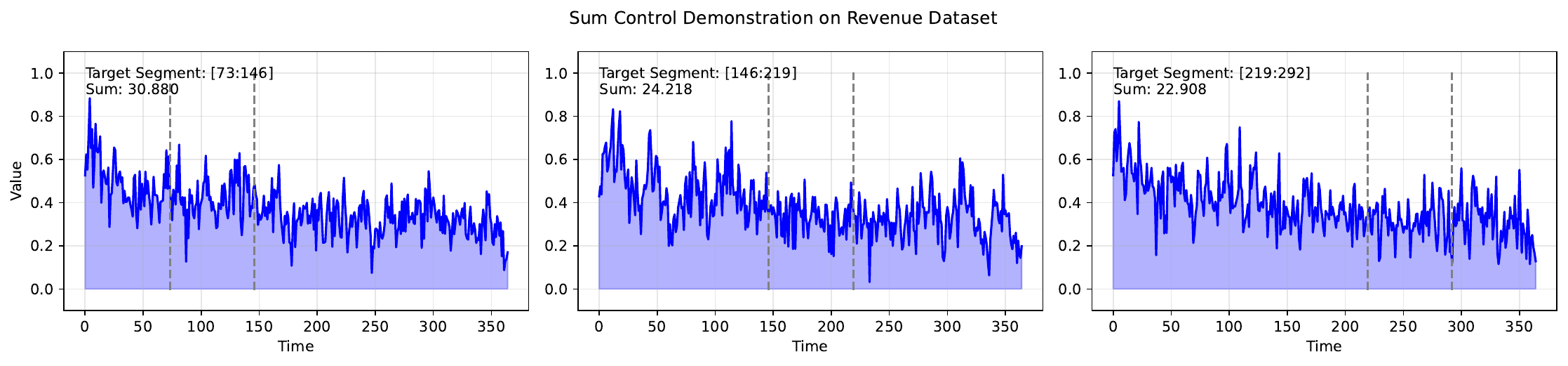}
    \vspace{-0.3cm}
    \caption{Demonstration of Segment-Wise Sum Control on Revenue dataset.}
\end{figure*}
\vspace{-1em}

\begin{figure*}[!htbp]
\includegraphics[width=\linewidth]{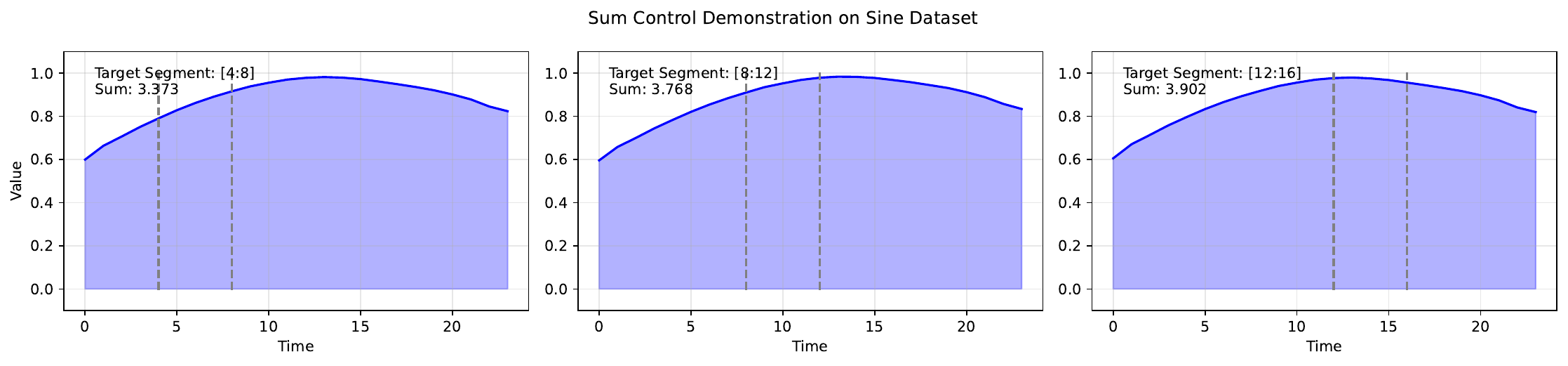}
\vspace{-0.3cm}
\caption{Demonstration of Segment-Wise Sum Control on synthetic sine wave dataset.}
\end{figure*}

\newpage
\subsubsection{Segment Wise Sum Control (CSDI)}
\begin{figure*}[!htbp]
\centering
\includegraphics[width=\linewidth]{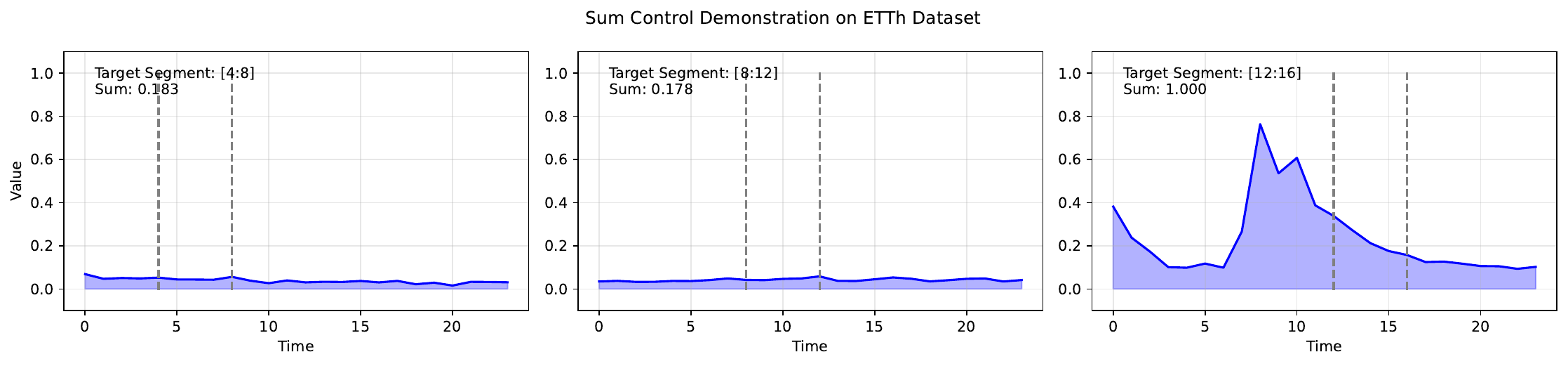}
\vspace{-0.3cm}
\caption{Demonstration of Segment-Wise Sum Control on ETTh dataset.}
\end{figure*}
    \vspace{-1em}
\begin{figure*}[!htbp]
    \includegraphics[width=\linewidth]{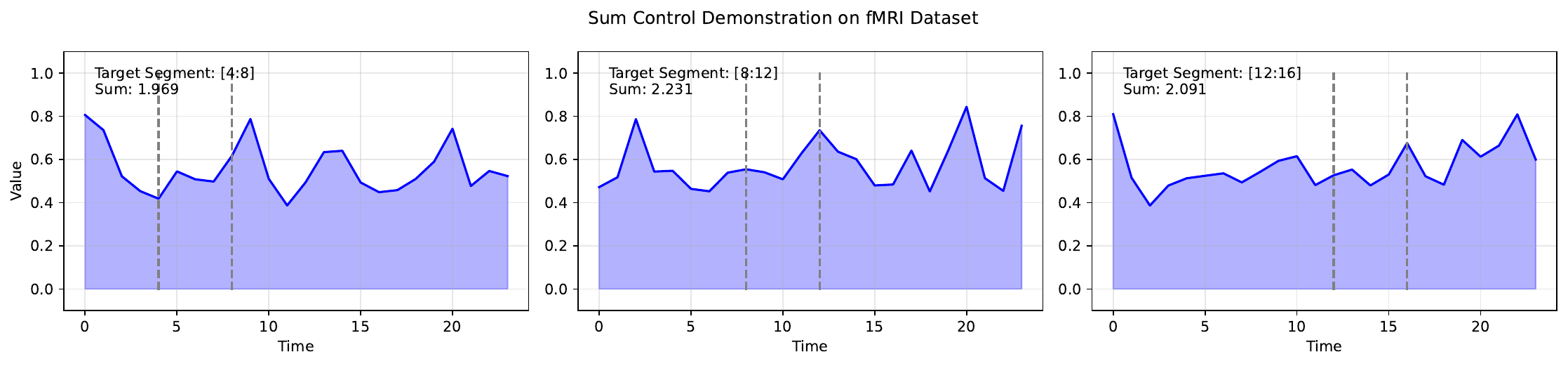}
    \vspace{-0.3cm}
    \caption{Demonstration of Segment-Wise Sum Control on fMRI dataset.}
\end{figure*}
\vspace{-1em}

\begin{figure*}[!htbp]
    \includegraphics[width=\linewidth]{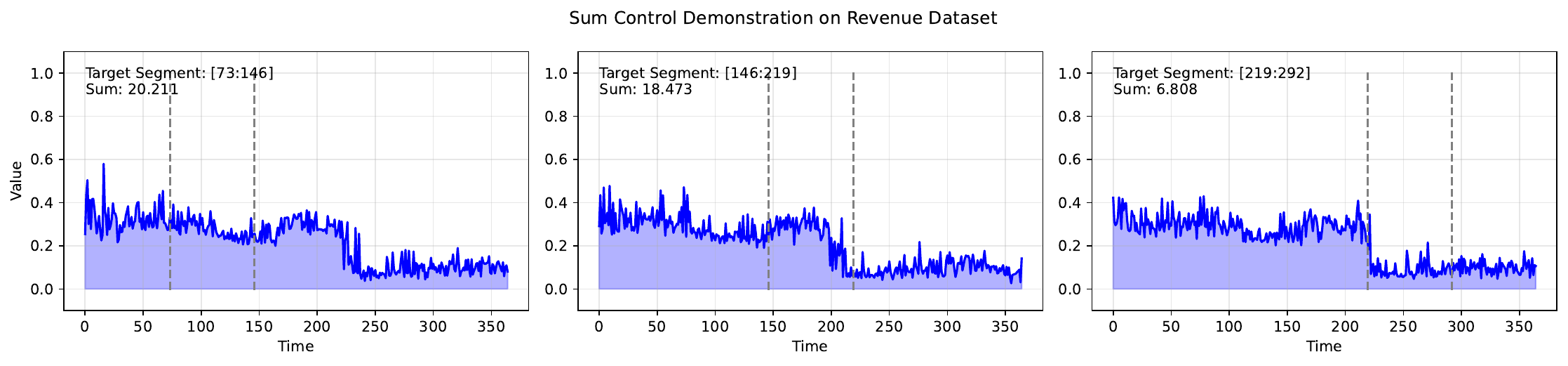}
    \vspace{-0.3cm}
    \caption{Demonstration of Segment-Wise Sum Control on Revenue dataset.}
\end{figure*}
\vspace{-1em}

\begin{figure*}[!htbp]
\includegraphics[width=\linewidth]{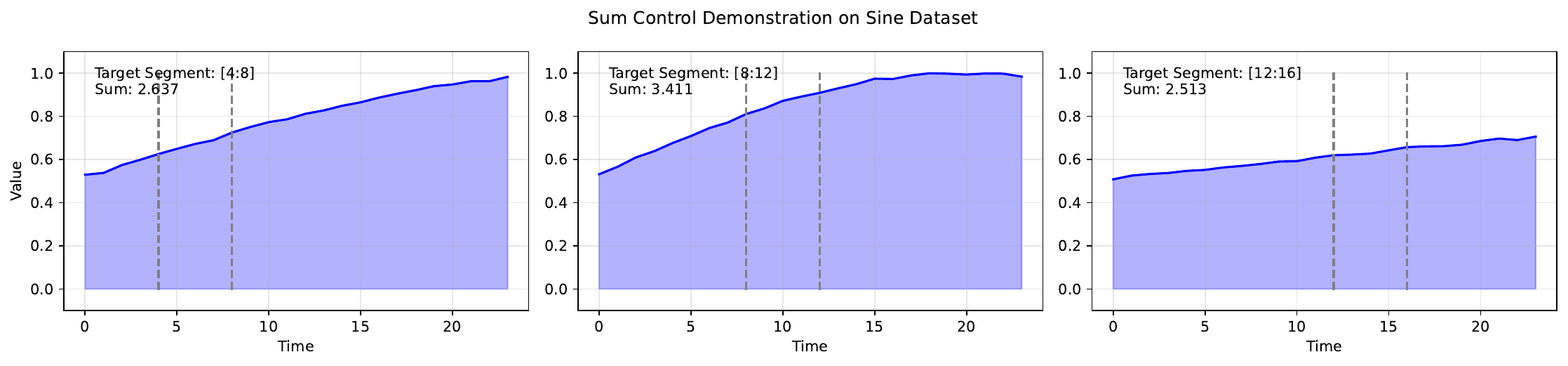}
\vspace{-0.3cm}
\caption{Demonstration of Segment-Wise Sum Control on synthetic sine wave dataset.}
\end{figure*}

\newpage
\subsection{Kernel Density Estimate of Sum Control}
Similar to the KDE analysis of point-wise control, we present KDE analysis for sum control across different datasets. The KDE peaks of controlled output (purple line) shift rightward compared to original and unconditional distributions, confirming that controlled sequences achieve higher sum values while preserving dataset-specific distributional characteristics. While this pattern does not persist consistently across different control weights and need to be further investigated.

\subsubsection{KDE of Total Sum Control (Diffusion-TS)}
\begin{figure*}[!htbp]
\label{fig:revenue_kde_sum}
\centering
\includegraphics[width=\textwidth]{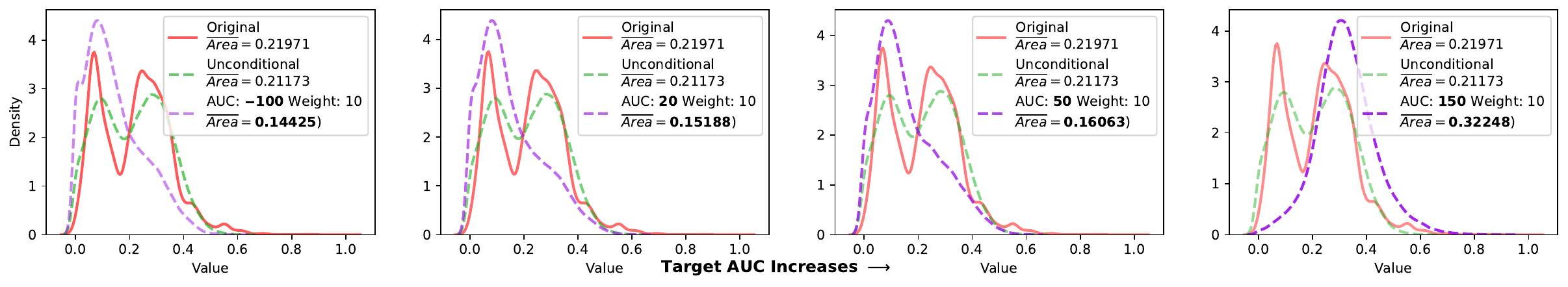}
\includegraphics[width=\textwidth]{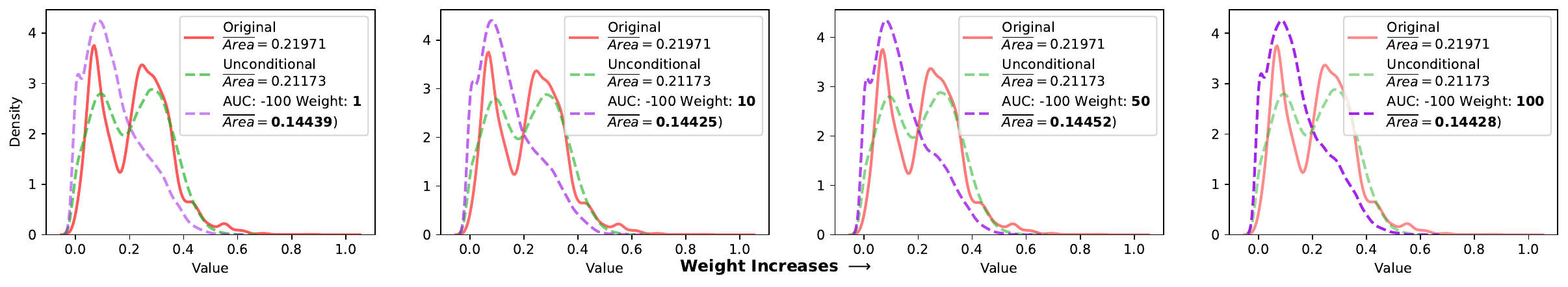}
\vspace{-0.3cm}
\caption{Kernel density estimation analysis of Revenue dataset under varying sum control targets. Top: Target analysis showing control effectiveness. Bottom: Weight analysis showing control effectiveness.}

\end{figure*}

\begin{figure*}[!htbp]
\label{fig:etth_kde_sum}
\centering
\includegraphics[width=0.95\textwidth]{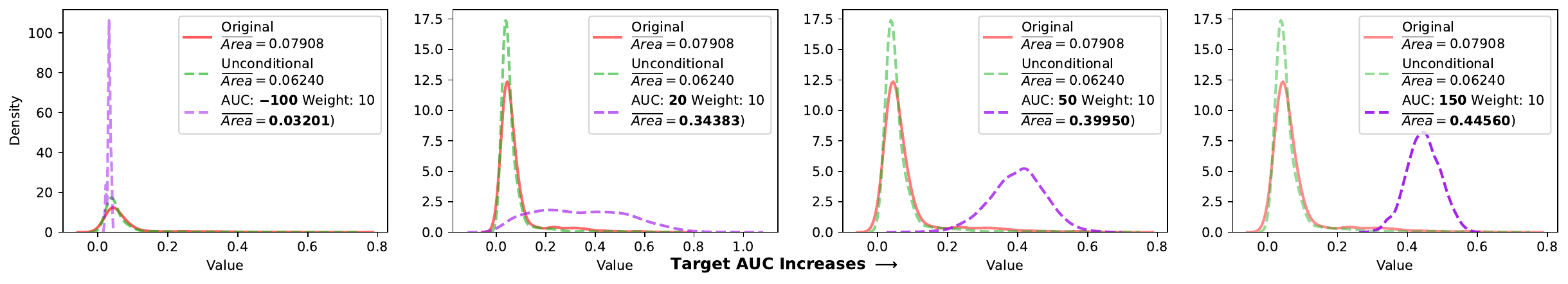}
\includegraphics[width=0.95\textwidth]{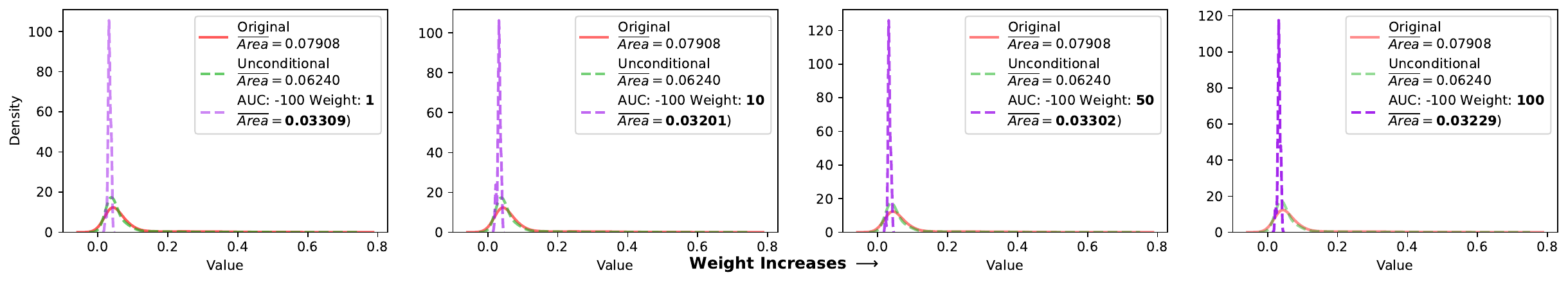}
\vspace{-0.3cm}
\caption{Kernel density estimation visualization for ETTh dataset. Top: Sum control analysis. Bottom: Weight analysis.}
\end{figure*}

\begin{figure*}[!htbp]
\label{fig:frmi_kde_sum}
\centering
\includegraphics[width=0.95\textwidth]{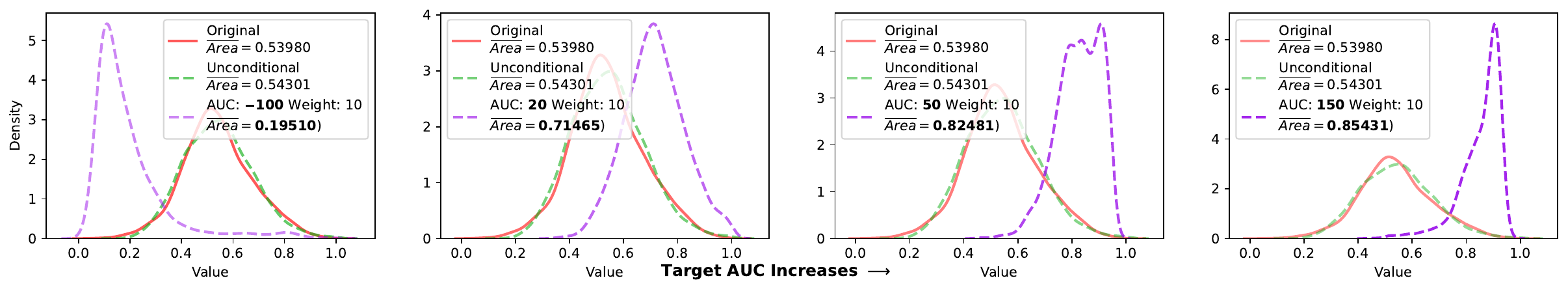}
\includegraphics[width=0.95\textwidth]{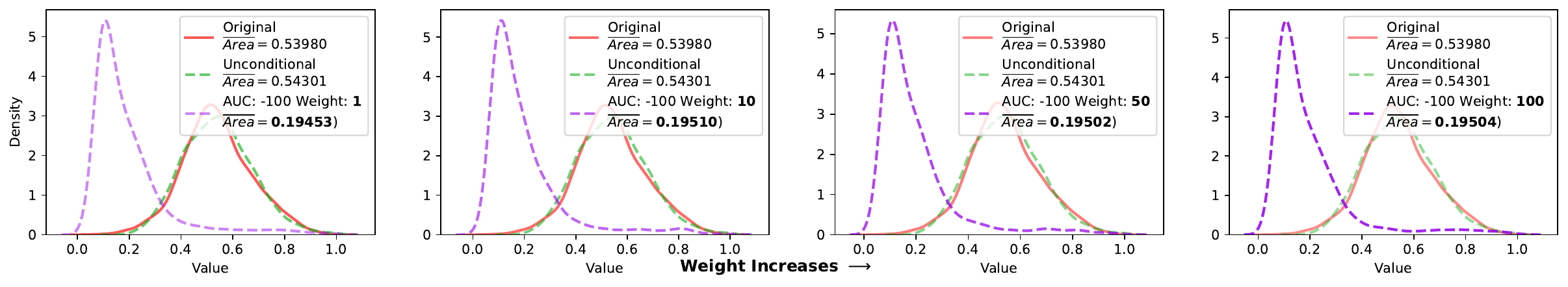}
\vspace{-0.3cm}
\caption{Kernel density estimation analysis of fMRI dataset. Top: Sum control analysis. Bottom: Weight analysis.}
\end{figure*}

\begin{figure*}[!htbp]
\label{fig:sine_kde_sum}
\centering
\includegraphics[width=0.95\textwidth]{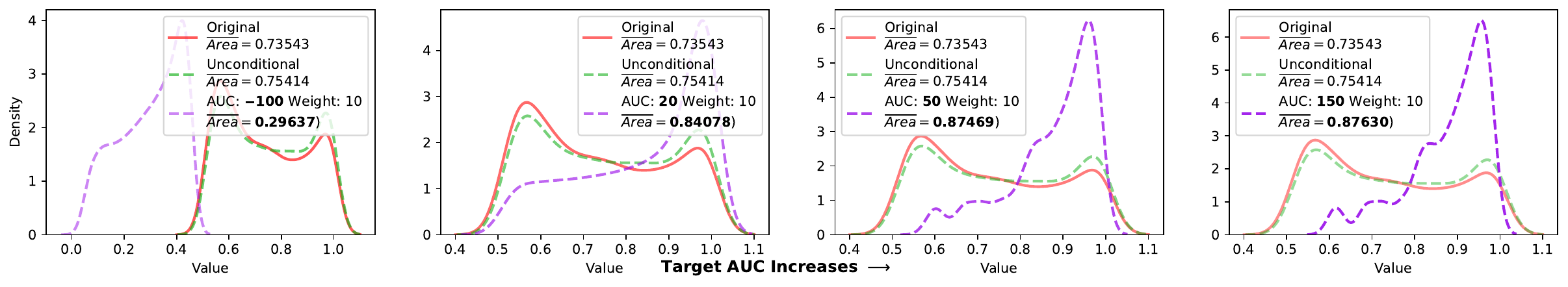}
\includegraphics[width=0.95\textwidth]{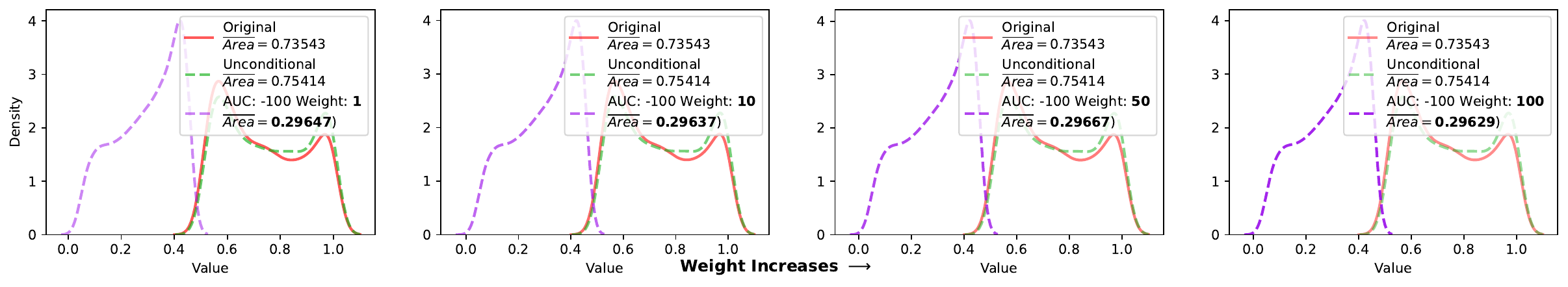}

\vspace{-0.3cm}
\caption{Kernel density estimation analysis of synthetic sine wave dataset. Top: Sum control analysis. Bottom: Weight analysis.}
\end{figure*}

\newpage
\subsubsection{KDE of Total Sum Control (CSDI)}

\begin{figure*}[!htbp]
\label{fig:revenue_kde_sum_csdi}
\centering
\includegraphics[width=0.95\textwidth]{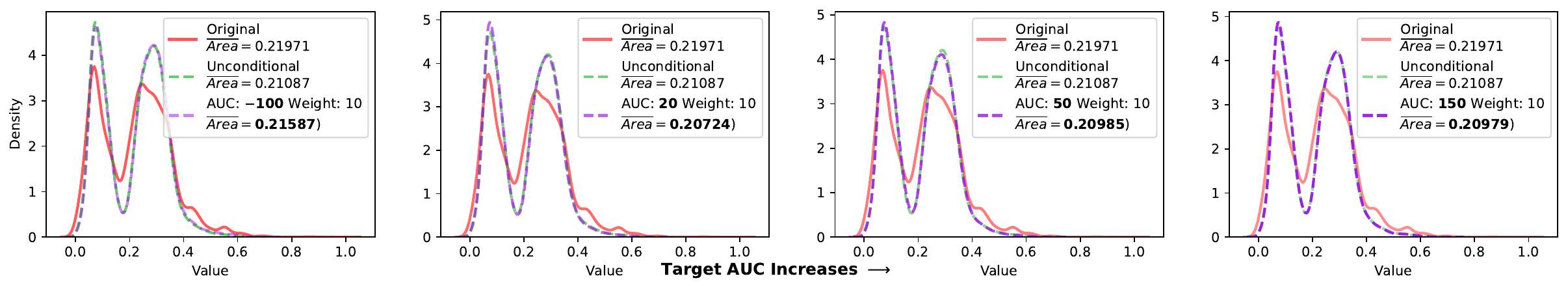}
\includegraphics[width=0.95\textwidth]{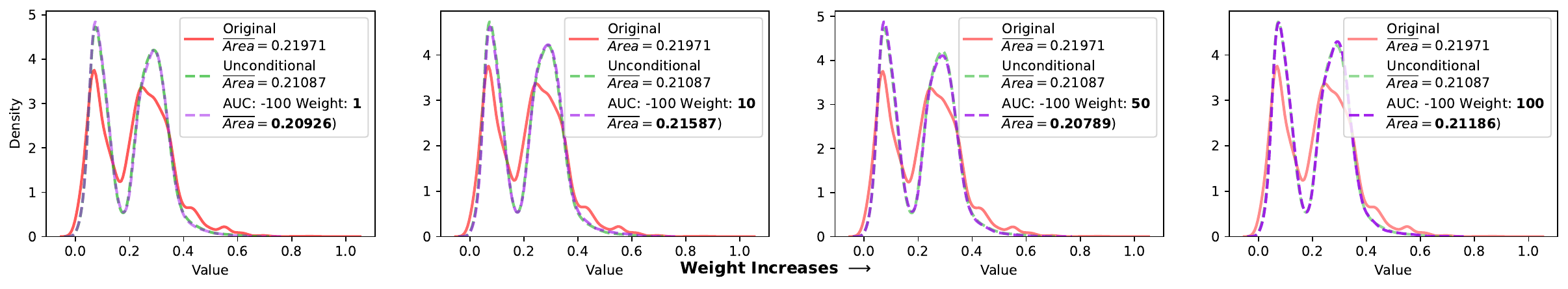}
\vspace{-0.3cm}
\caption{Kernel density estimation analysis of Revenue dataset under varying sum control targets. Top: Target analysis showing control effectiveness. Bottom: Weight analysis showing control effectiveness.}

\end{figure*}

\begin{figure*}[!htbp]
\label{fig:etth_kde_sum_csdi}
\centering
\includegraphics[width=0.95\textwidth]{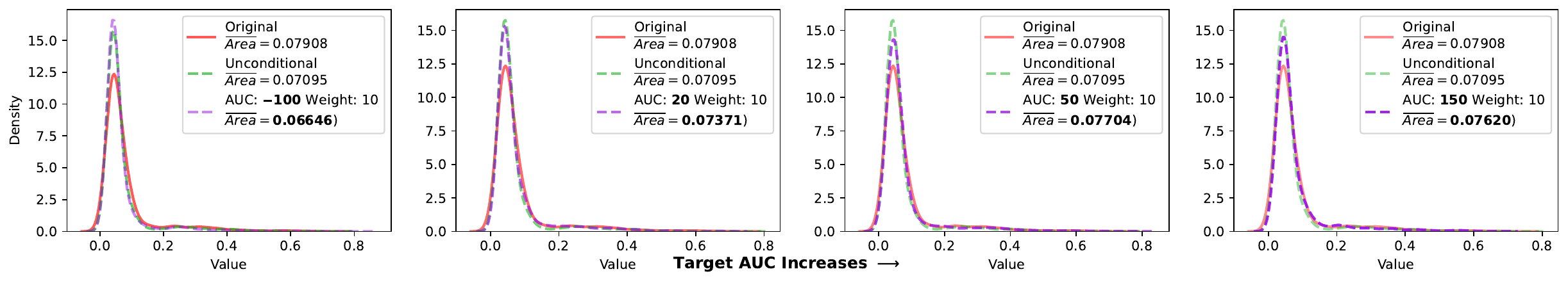}
\includegraphics[width=0.95\textwidth]{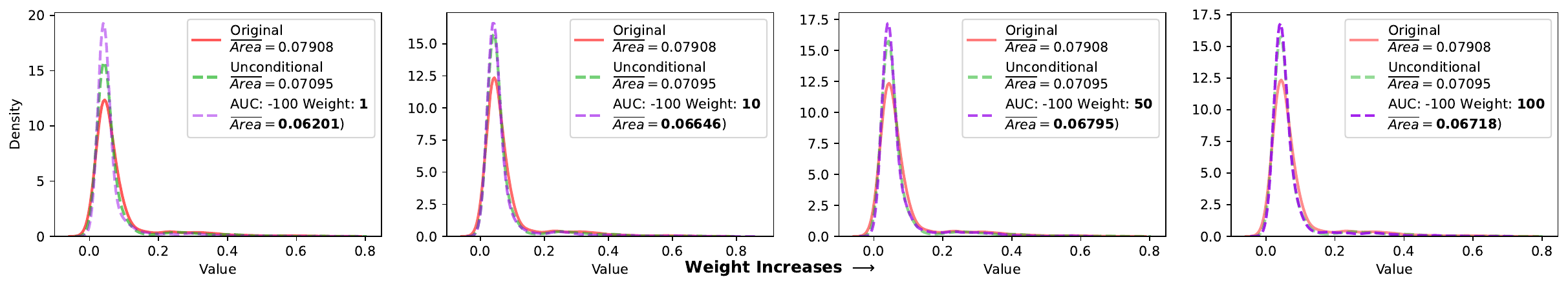}
\vspace{-0.3cm}
\caption{Kernel density estimation visualization for ETTh dataset. Top: Sum control analysis. Bottom: Weight analysis.}
\end{figure*}

\begin{figure*}[!htbp]
\label{fig:frmi_kde_sum_csdi}
\centering
\includegraphics[width=0.95\textwidth]{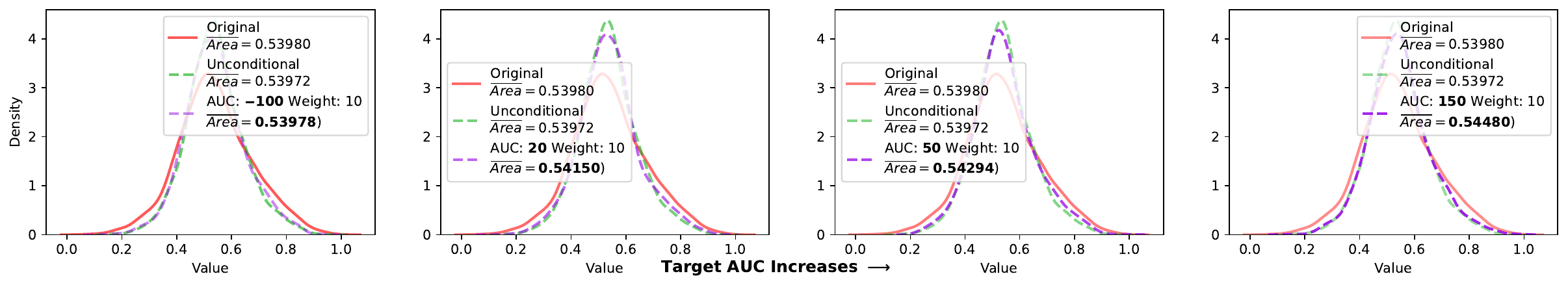}
\includegraphics[width=0.95\textwidth]{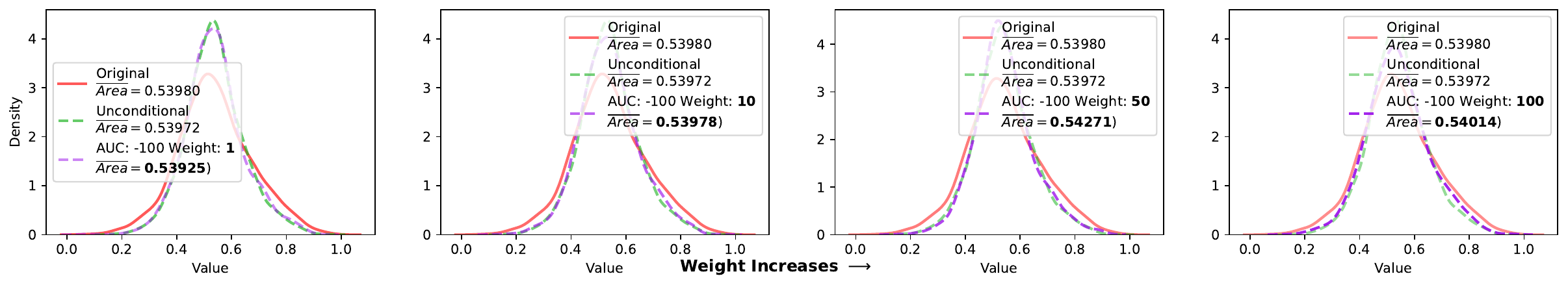}
\vspace{-0.3cm}
\caption{Kernel density estimation analysis of fMRI dataset. Top: Sum control analysis. Bottom: Weight analysis.}
\end{figure*}

\begin{figure*}[!htbp]
\label{fig:sine_kde_sum_csdi}
\centering
\includegraphics[width=0.95\textwidth]{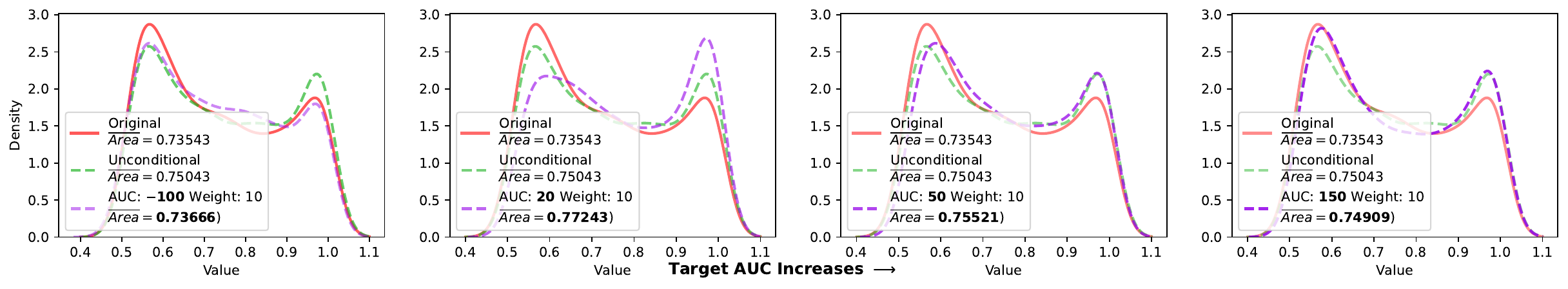}
\includegraphics[width=0.95\textwidth]{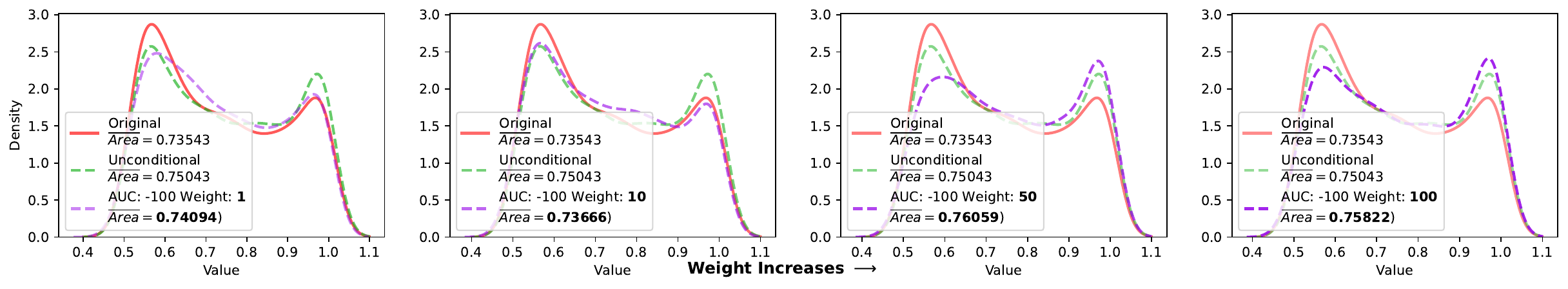}

\vspace{-0.3cm}
\caption{Kernel density estimation analysis of synthetic sine wave dataset. Top: Sum control analysis. Bottom: Weight analysis.}
\end{figure*}

\newpage
\subsection{Averaged Sum Change Over All Segments}
For the following aggregated sum change over all segments, we calculate the averaged value over segments for each dataset, with a target sum value of 150 for each segment.

\subsubsection{Value Change of Segment Sum Control (Diffusion-TS)}
\begin{figure*}[!htbp]
\label{fig:revenue_kde_sum_segement_diffusion_ts}
\centering
\includegraphics[width=\textwidth]{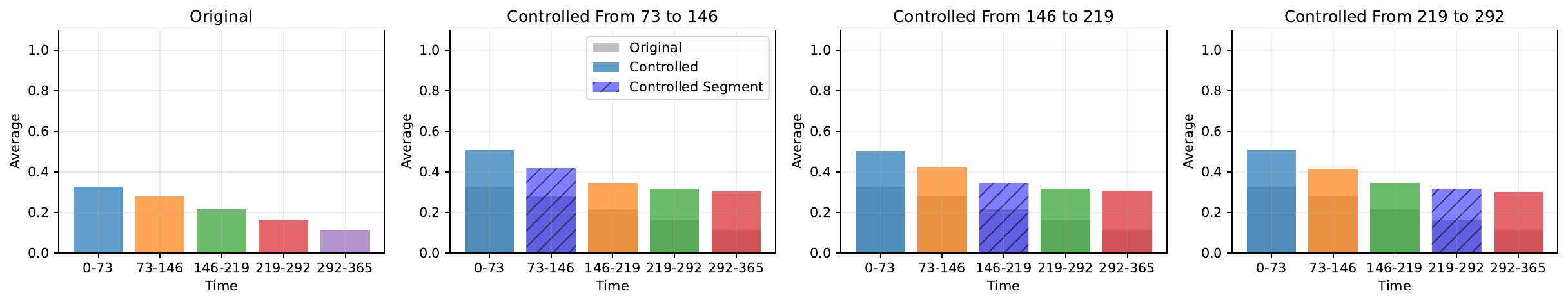}
\vspace{-0.3cm}
\caption{Segmented Summation Control on Revenue dataset.}
\end{figure*}

\begin{figure*}[!htbp]
\label{fig:etth_kde_sum_segment_bins_diffusion_ts}
\centering
\includegraphics[width=\textwidth]{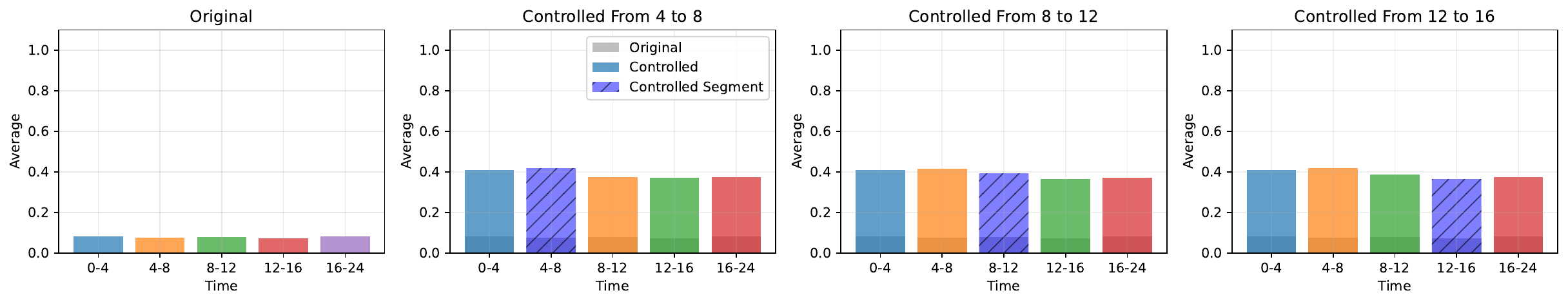}
\vspace{-0.3cm}
\caption{Segmented Summation Control on ETTh dataset.}
\end{figure*}

\begin{figure*}[!htbp]
\label{fig:frmi_kde_sum_segment_bins_diffusion_ts}
\centering
\includegraphics[width=\textwidth]{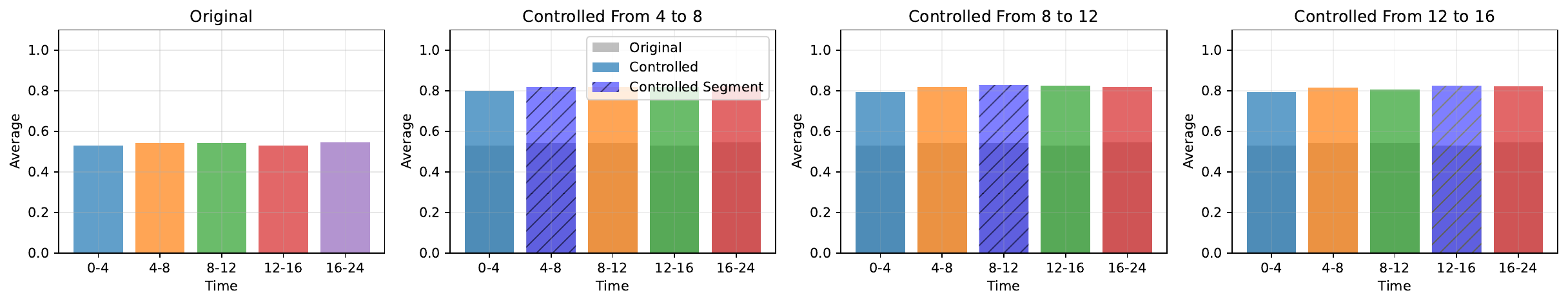}
\vspace{-0.3cm}
\caption{Segmented Summation Control on fMRI dataset.}
\end{figure*}

\begin{figure*}[!htbp]
\label{fig:sine_kde_sum_segment_bins_diffusion_ts}
\centering
\includegraphics[width=\textwidth]{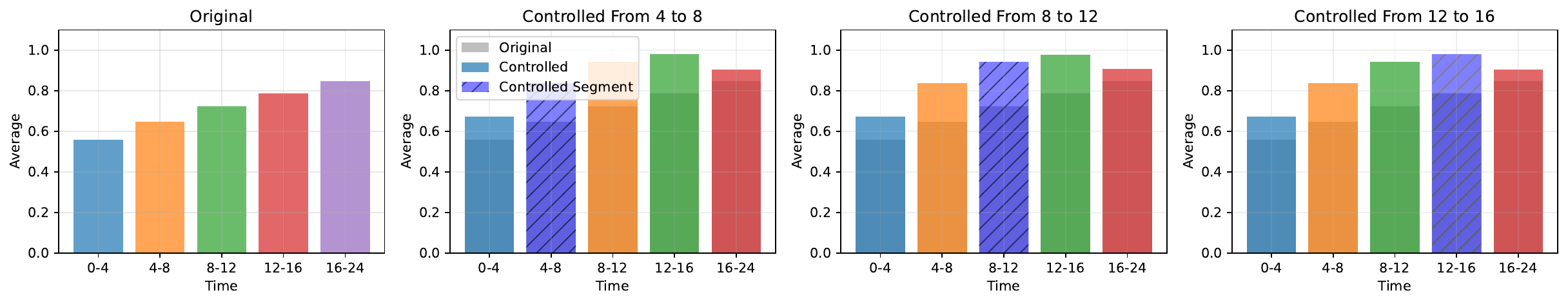}
\vspace{-0.3cm}
\caption{Segmented Summation Control on Sine dataset.}
\end{figure*}

\newpage
\subsubsection{Value Change of Segment Sum Control (CSDI)}

\begin{figure*}[!htbp]
\label{fig:revenue_kde_sum_segement_csdi}
\centering
\includegraphics[width=\textwidth]{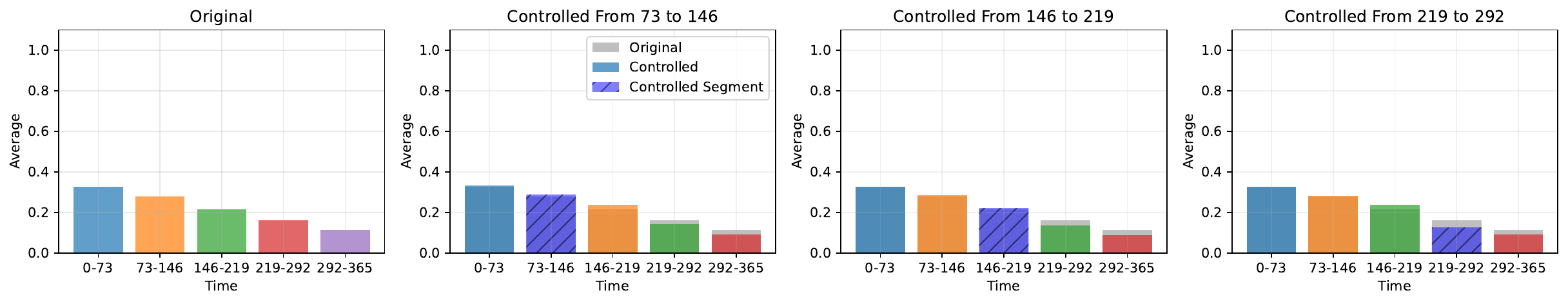}
\vspace{-0.3cm}
\caption{Segmented Summation Control on Revenue dataset.}
\end{figure*}

\begin{figure*}[!htbp]
\label{fig:etth_kde_sum_segment_bins_csdi}
\centering
\includegraphics[width=\textwidth]{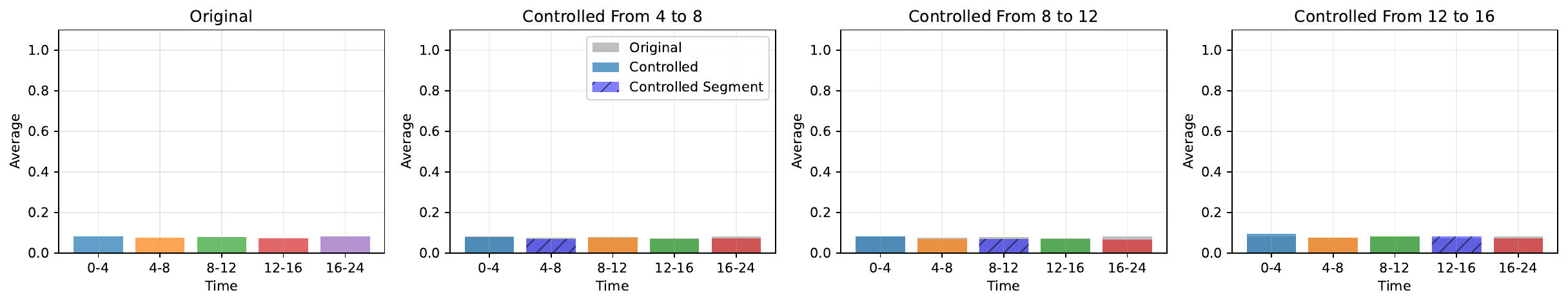}
\vspace{-0.3cm}
\caption{Segmented Summation Control on ETTh dataset.}
\end{figure*}

\begin{figure*}[!htbp]
\label{fig:frmi_kde_sum_segment_bins_csdi}
\centering
\includegraphics[width=\textwidth]{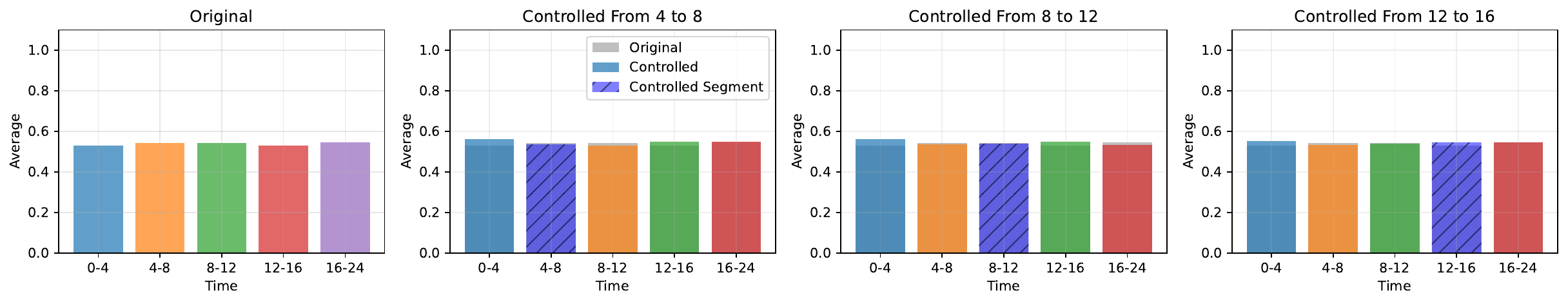}
\vspace{-0.3cm}
\caption{Segmented Summation Control on fMRI dataset.}
\end{figure*}

\begin{figure*}[!htbp]
\label{fig:sine_kde_sum_segment_bins_csdi}
\centering
\includegraphics[width=\textwidth]{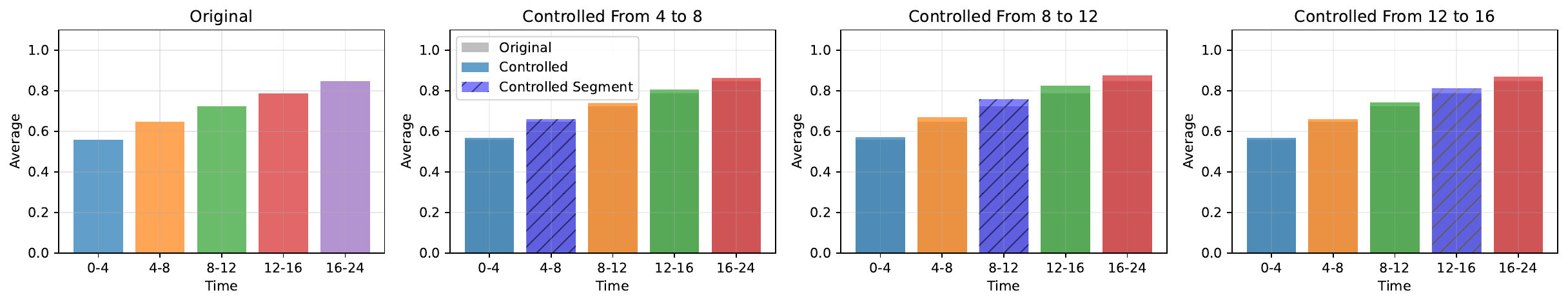}
\vspace{-0.3cm}
\caption{Segmented Summation Control on Sine dataset.}
\end{figure*}

\newpage
\subsection{Supplement Metrics}
The Discriminative, Predictive, Context-FID, and Correlational scores help quantify distribution shifts under sum control. Table \ref{tab:metrics-statistic-supplement} demonstrates that all metrics increase significantly after applying control, indicating that the control signals effectively influence the generated time series and make them more distinguishable from the original distribution.

\begin{table}[H]
\label{tab:metrics-statistic-supplement}
\centering
\caption{Supplemental metrics for sum control performance across different datasets and target values. The results show discriminative, predictive, context-FID, and correlational scores for each dataset and control configuration. Lower scores indicate better performance.}
\resizebox{0.8\columnwidth}{!}{
\begin{tabular}{c|c|cccc}
\toprule
\multirow{2}{*}{Metrics} & \multirow{2}{*}{Control Signal} & \multicolumn{4}{c}{Dataset} \\
 &  & ETTh & Revenue & fMRI & Sine \\ \hline
\multirow{5}{*}{\begin{tabular}[c]{@{}c@{}}Discriminative\\ Score\\ (Lower is\\ Better)\end{tabular}} & Unconditional & 0.103$\pm$0.042& 0.082$\pm$0.093& 0.141$\pm$0.037& 0.031$\pm$0.023\\
 & Sum Target = 150 & 0.499$\pm$0.002& 0.455$\pm$0.069& 0.500$\pm$0.000& 0.453$\pm$0.117\\
 & Sum Target = 50 & 0.499$\pm$0.002& 0.445$\pm$0.062& 0.500$\pm$0.000& 0.477$\pm$0.031\\
 & Sum Target = 20 & 0.488$\pm$0.005& 0.455$\pm$0.056& 0.500$\pm$0.000& 0.244$\pm$0.100\\
 & Sum Target = -100 & 0.476$\pm$0.015& 0.427$\pm$0.064& 0.500$\pm$0.000& 0.500$\pm$0.000\\ \hline
\multirow{5}{*}{\begin{tabular}[c]{@{}c@{}}Predictive\\ Score\\ (Lower is\\ Better)\end{tabular}} & Unconditional & 0.256$\pm$0.002& 0.065$\pm$0.026& 0.103$\pm$0.002& 0.094$\pm$0.000\\
 & Sum Target = 150 & 0.465$\pm$0.008& 0.146$\pm$0.066& 0.113$\pm$0.001& 0.099$\pm$0.009\\
 & Sum Target = 50 & 0.418$\pm$0.007& 0.107$\pm$0.006& 0.108$\pm$0.001& 0.108$\pm$0.040\\
 & Sum Target = 20 & 0.286$\pm$0.004& 0.104$\pm$0.007& 0.102$\pm$0.001& 0.094$\pm$0.000\\
 & Sum Target = -100 & 0.289$\pm$0.005& 0.106$\pm$0.009& 0.116$\pm$0.003& 0.111$\pm$0.008\\ \hline
\multirow{5}{*}{\begin{tabular}[c]{@{}c@{}}Context-FID\\ Score\\ (Lower is\\ Better)\end{tabular}} & Unconditional & 0.108$\pm$0.007& 1.230$\pm$0.284& 0.260$\pm$0.024& 0.034$\pm$0.005\\
 & Sum Target = 150 & 10.608$\pm$1.651& 2.648$\pm$0.927& 2.740$\pm$0.480& 4.346$\pm$0.818\\
 & Sum Target = 50 & 8.943$\pm$1.277& 2.408$\pm$0.245& 2.187$\pm$0.115& 2.827$\pm$0.389\\
 & Sum Target = 20 & 3.129$\pm$0.473& 3.533$\pm$0.849& 0.614$\pm$0.111& 0.884$\pm$0.281\\
 & Sum Target = -100 & 5.049$\pm$0.537& 5.056$\pm$0.459& 3.122$\pm$0.389& 19.805$\pm$1.737\\ \hline
\multirow{5}{*}{\begin{tabular}[c]{@{}c@{}}Correlational\\ Score\\ (Lower is\\ Better)\end{tabular}} & Unconditional & 2.313$\pm$0.743& 0.038$\pm$0.013& 2.672$\pm$0.091& 0.066$\pm$0.009\\
 & Sum Target = 150 & 15.328$\pm$0.511& 0.098$\pm$0.010& 9.387$\pm$0.149& 0.297$\pm$0.012\\
 & Sum Target = 50 & 10.872$\pm$0.812& 0.079$\pm$0.015& 7.809$\pm$0.094& 0.210$\pm$0.029\\
 & Sum Target = 20 & 5.219$\pm$0.507& 0.082$\pm$0.011& 4.247$\pm$0.136& 0.111$\pm$0.016\\
 & Sum Target = -100 & 8.345$\pm$0.324& 0.074$\pm$0.002& 9.508$\pm$0.169& 0.837$\pm$0.008\\
 \bottomrule
\end{tabular}%
}
\end{table}
\subsection{Additional Sum Control Analysis}
The following figures provide comprehensive analysis of sum control performance across datasets (Unnormalized). The plots demonstrate achieved sum values compared to target values, with Original and Unconditional (Uncon) baselines as references. Analysis of control weight's impact shows minimal influence on achieved sum values across different datasets and target configurations.
\begin{figure*}[!htbp]
\label{fig:sum_trends}
\centering
\includegraphics[width=\textwidth]{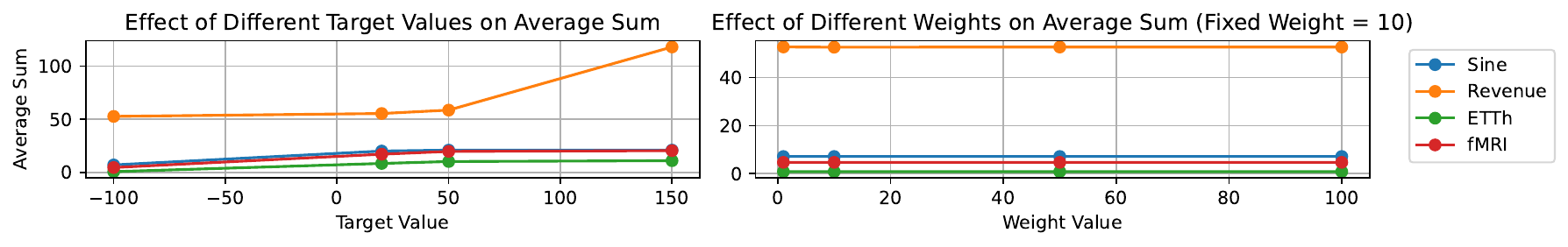}
\vspace{-0.3cm}
\caption{Comparison of achieved sum values versus target sum values across different datasets. The plots demonstrate the effectiveness of sum control guidance in reaching desired targets.}
\end{figure*}

\section{Distribution Analysis}

The FID scores reveal a fundamental trade-off between controllability and distribution preservation. While Diffusion-TS demonstrates stronger modification capabilities (FID increasing from 0.416 to 16.812 with Anchor control on fMRI), it comes at the cost of significant distribution shifts. In contrast, CSDI shows more resistance to modification but better preserves the original distribution (FID changes from 1.188 to only 1.016 under similar conditions).

This raises an important open question: How can we achieve precise temporal control while maintaining distribution fidelity? Future research should investigate mechanisms to balance these competing objectives, potentially through adaptive control strength or hybrid architectures that combine the stability of CSDI with the flexibility of Diffusion-TS.

\newpage
\label{app:distruibution}
\begin{table*}[h]
\centering
\caption{The complete distribution of discriminative, predictive, correlational, and FID scores for our method across different datasets and control configurations. For all metrics, lower scores indicate better performance.}
\resizebox{0.9\textwidth}{!}{
\begin{tabular}{l|l|l|cccc}
\toprule
Model & Metric & Control & ETTh & Revenue & fMRI & Sines \\ \midrule
\multirow{12}{*}{Our - CSDI} & \multirow{3}{*}{Discriminative Score} & Unconditional & 0.361$\pm$0.007 & 0.245$\pm$0.164& 0.306$\pm$0.021 & 0.017$\pm$0.007 \\
    &  & Point-Wise Control & 0.470$\pm$0.003 & 0.313$\pm$0.046& 0.482$\pm$0.004 & 0.430$\pm$0.038 \\
    &  & Statistics Control & 0.373$\pm$0.007 & 0.272$\pm$0.055& 0.377$\pm$0.019 & 0.034$\pm$0.007 \\ \cmidrule{2-7} 
    & \multirow{3}{*}{Predictive Score} & Unconditional & 0.261$\pm$0.003 & 0.054$\pm$0.012& 0.106$\pm$0.000 & 0.090$\pm$0.000 \\
    &  & Point-Wise Control & 0.263$\pm$0.001 & 0.070$\pm$0.003& 0.106$\pm$0.000 & 0.091$\pm$0.000 \\
    &  & Statistics Control & 0.261$\pm$0.001 & 0.060$\pm$0.005& 0.106$\pm$0.001 & 0.091$\pm$0.000 \\ \cmidrule{2-7} 
    & \multirow{3}{*}{Correlational Score} & Unconditional & 8.428$\pm$0.000 & 0.034$\pm$0.000& 2.212$\pm$0.000 & 0.062$\pm$0.000 \\
    &  & Point-Wise Control & 8.641$\pm$0.000 & 0.024$\pm$0.000& 2.619$\pm$0.000 & 0.149$\pm$0.000 \\
    &  & Statistics Control & 8.531$\pm$0.000 & 0.034$\pm$0.000& 3.944$\pm$0.000 & 0.065$\pm$0.000 \\ \cmidrule{2-7} 
    & \multirow{3}{*}{FID Score} & Unconditional & 1.643$\pm$0.171 & 1.129$\pm$0.122& 1.188$\pm$0.054 & 0.034$\pm$0.006 \\
    &  & Point-Wise Control & 2.720$\pm$0.133 & 2.233$\pm$0.088& 1.016$\pm$0.019 & 3.170$\pm$0.319 \\
    &  & Statistics Control & 1.564$\pm$0.050 & 1.097$\pm$0.040& 1.175$\pm$0.018 & 0.043$\pm$0.002 \\ \midrule
\multirow{12}{*}{Our - Diffusion-TS} & \multirow{3}{*}{Discriminative Score} & Unconditional & 0.034$\pm$0.026 & 0.209$\pm$0.185& 0.089$\pm$0.033 & 0.019$\pm$0.008 \\
    &  & Point-Wise Control & 0.437$\pm$0.004 & 0.393$\pm$0.030& 0.495$\pm$0.001 & 0.460$\pm$0.011 \\
    &  & Statistics Control & 0.477$\pm$0.003 & 0.426$\pm$0.032& 0.498$\pm$0.001 & 0.451$\pm$0.029 \\ \cmidrule{2-7} 
    & \multirow{3}{*}{Predictive Score} & Unconditional & 0.260$\pm$0.002 & 0.070$\pm$0.015& 0.110$\pm$0.001 & 0.090$\pm$0.000 \\
    &  & Point-Wise Control & 0.314$\pm$0.003 & 0.128$\pm$0.011& 0.136$\pm$0.002 & 0.153$\pm$0.006 \\
    &  & Statistics Control & 0.310$\pm$0.004 & 0.114$\pm$0.005& 0.117$\pm$0.001 & 0.110$\pm$0.003 \\ \cmidrule{2-7} 
    & \multirow{3}{*}{Correlational Score} & Unconditional & 1.728$\pm$0.000 & 0.033$\pm$0.000& 1.673$\pm$0.000 & 0.037$\pm$0.000 \\
    &  & Point-Wise Control & 5.647$\pm$0.000 & 0.107$\pm$0.000& 16.791$\pm$0.000 & 0.405$\pm$0.000 \\
    &  & Statistics Control & 9.190$\pm$0.000 & 0.083$\pm$0.000& 8.361$\pm$0.000 & 0.606$\pm$0.000 \\ \cmidrule{2-7} 
    & \multirow{3}{*}{FID Score} & Unconditional & 0.177$\pm$0.015 & 1.221$\pm$0.040& 0.416$\pm$0.011 & 0.021$\pm$0.003 \\
    &  & Point-Wise Control & 5.819$\pm$0.303 & 3.479$\pm$0.157& 16.812$\pm$0.485 & 4.523$\pm$0.456 \\
    &  & Statistics Control & 5.335$\pm$0.221 & 3.822$\pm$0.210& 2.652$\pm$0.092 & 5.954$\pm$0.438 \\ \bottomrule
\end{tabular}%
}

\end{table*}

\newpage
\newpage
\section{Combined Control on Revenue Dataset}
\label{app:combine_demo}

To demonstrate our method's capability to handle multiple control signals simultaneously, we present a comprehensive example using the Revenue dataset. The model successfully generates sequences that respect both anchor points and sum constraints, highlighting the flexibility and effectiveness of our approach. These generated sequences maintain the dataset's inherent distributional characteristics while precisely adhering to multiple control signals. For additional examples of combined control mechanisms, see Figure ~\ref{fig:editor} which showcases our Time Series Editor interface in action.

\begin{figure}[!hbpt]
    \label{fig:combined-control}
    \centering
    \includegraphics[width=0.9\linewidth]{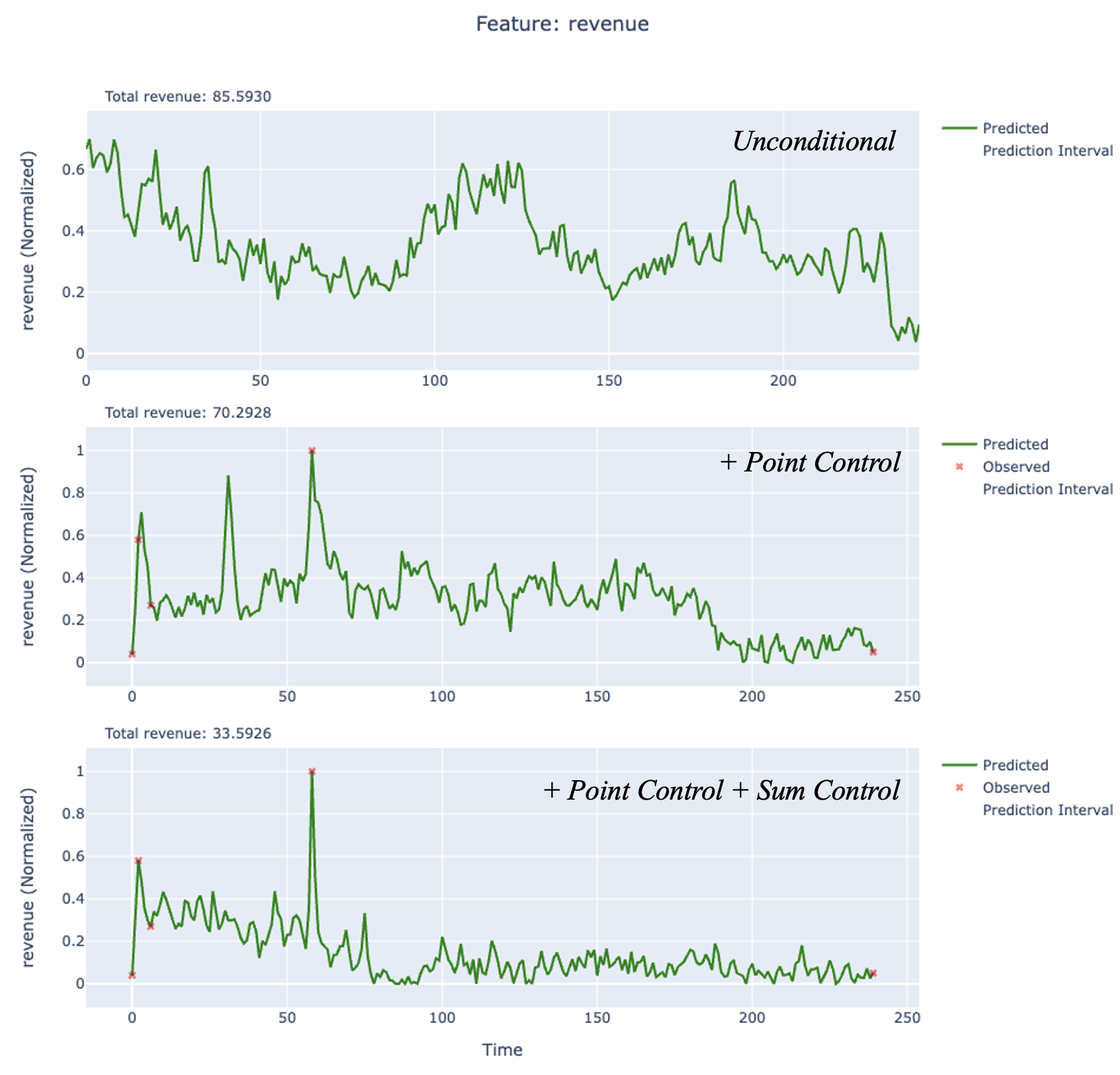}
    \caption{Demonstration of combined anchor and sum control on the Revenue dataset, showing the interaction across point-wise constraints and overall sum requirements.}
\end{figure}

\newpage

\section{Time Series Editor}

\begin{figure}[H]
    \label{fig:editor}
    \centering
    \includegraphics[width=0.82\linewidth]{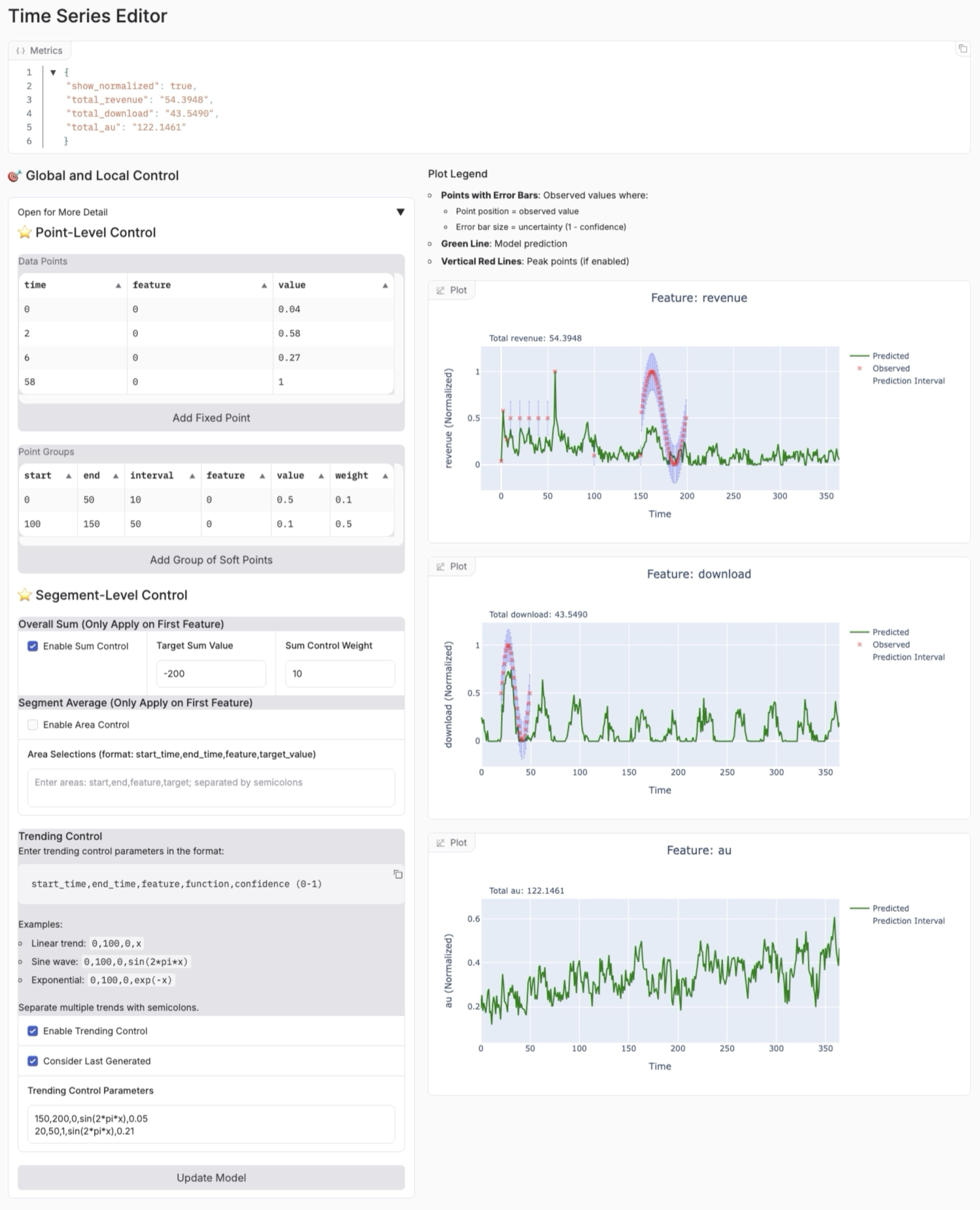}
    \caption{The Screen Shot of Time Series Editor User Interface}
\end{figure}
\vspace{-1em}
The Time Series Editor is designed to solve the Time Series Editing problem. It can add fixed points, soft anchors, trending control, segment-level sum, and average control. For the soft anchors, you can add anchor points based on start, end, and interval setup. The Trending Control is based on the provided function expressions with independent variable $x$ for better precise control. We are currently developing the SketchPad mode for better user interaction, aiming to provide an All-in-One application for efficient time series editing without training.

In Figure~\ref{fig:editor}, the green line represents the model prediction, and the red dots represent the observed/provided anchor points. The error bar of each anchor point is the ($1-\mathrm{confidence}$) to demonstrate the uncertainty of the observed points. In Figure~\ref{fig:editor}, we use the Revenue Dataset with three features: Revenue, Download, and Daily Active Users.

\end{document}